\documentclass[a4paper]{article}[11pts]

\usepackage[english]{babel}
\usepackage[utf8x]{inputenc}
\usepackage[T1]{fontenc}

\normalsize

\usepackage[a4paper,top=1in,bottom=1in,left=1in,right=1in,marginparwidth=1in]{geometry}
\usepackage{setspace}

\usepackage{appendix}
\usepackage{color}
\definecolor{strcolor}{rgb}{0.6, 0.2, 0.6}
\definecolor{commentcolor}{rgb}{0.3125, 0.5, 0.3125}
\definecolor{keycol}{rgb}{0, 0, 1}

\usepackage{amssymb}
\usepackage{amsmath}
\usepackage{amsthm}
\usepackage{amsfonts, mathtools}
\usepackage{mathrsfs}
\usepackage{bbm}
\usepackage{tablefootnote}
\usepackage{booktabs}
\usepackage{multirow}

\DeclareMathOperator*{\argmax}{arg\,max}

\newtheorem{proposition}{Proposition}
\newtheorem{lemma}{Lemma}

\newtheorem{assumption}{Assumption}
\newtheorem{theorem}{Theorem}

\newtheorem{definition}{Definition}

\usepackage{listings}
\usepackage{url}
\usepackage{hyperref}
\hypersetup{
    hidelinks,
    colorlinks=true,
    linkcolor=red,
    citecolor=magenta
} % delete the box around the link

\newcommand {\bea}{\begin{eqnarray}}
	\newcommand {\eea}{\end{eqnarray}}

\newtheorem{algorithm}{Algorithm}
\newtheorem{remark}{Remark}

\def\blot{\quad \mbox{$\vcenter{ \vbox{ \hrule height.4pt
				\hbox{\vrule width.4pt height.9ex \kern.9ex \vrule width.4pt}
				\hrule height.4pt}}$}}

\usepackage{natbib}
 \bibpunct[, ]{(}{)}{,}{a}{}{,}%
\usepackage{xspace}

\usepackage{subfig}
\usepackage{adjustbox}
\usepackage{algorithm}
\usepackage{algpseudocode}
\usepackage{tikz}
\usepackage{pgfplots}
\pgfplotsset{compat=1.5}
\usepackage{multirow}
\usepgfplotslibrary{fillbetween, groupplots}
\usepackage{makecell}

\makeatletter
\newenvironment{breakablealgorithm}
  {% \begin{breakablealgorithm}
   \begin{center}
     \refstepcounter{algorithm}% New algorithm
     \hrule height.8pt depth0pt \kern2pt% \@fs@pre for \@fs@ruled
     \renewcommand{\caption}[2][\relax]{% Make a new \caption
       {\raggedright\textbf{\fname@algorithm~\thealgorithm} ##2\par}%
       \ifx\relax##1\relax % #1 is \relax
         \addcontentsline{loa}{algorithm}{\protect\numberline{\thealgorithm}##2}%
       \else % #1 is not \relax
         \addcontentsline{loa}{algorithm}{\protect\numberline{\thealgorithm}##1}%
       \fi
       \kern2pt\hrule\kern2pt
     }
  }{% \end{breakablealgorithm}
     \kern2pt\hrule\relax% \@fs@post for \@fs@ruled
   \end{center}
  }
\makeatother

\pgfplotsset{
  every nth point/.style={
    x filter/.code={
      \pgfmathtruncatemacro{\keep}{mod(\coordindex,#1)}
      \ifnum\keep=0
      \else
        
      \fi
    }
  }
}

\begin{document}
	%%%%%%%%%%%%%%%%

	\title{Self-Improving Neural-Guided Pruning: A Graph Neural Network Framework for Scalable Mixed Bundle Pricing}

	\author{Liangyu Ding \and Chenghan Wu \and Guokai Li \and Zizhuo Wang} %
    \date{\small
        School of Data Science, The Chinese University of Hong Kong, Shenzhen, Guangdong, China\\
        \{liangyuding, chenghanwu, guokaili\}@link.cuhk.edu.cn, wangzizhuo@cuhk.edu.cn}
     \maketitle

     \onehalfspacing

    \begin{abstract}
Mixed bundle pricing is a classic revenue management problem arising in industries such as e-commerce, tourism, and video games. It refers to designing product combinations (i.e., \textit{bundles}) and determining their prices to maximize expected profit. Exact mixed bundling models capture this structure but become computationally intractable because the number of candidate bundles grows exponentially with the number of products. We develop a graph neural network (GNN)-guided pruning-then-optimization framework for bundle pricing with (non-)additive valuations. The method represents each instance as a compact segment-product graph, predicts segment-product inclusion probabilities, and accordingly prunes the exponential bundle space into a small candidate family; the final prices and bundle offerings are obtained by solving the mixed bundling formulation over the retained bundles, possibly refined by a GNN-guided local search. Because exact labels are available only at small scales, we further propose an \textit{iterative self-improvement} procedure: the current GNN policies generate high-quality solutions on large-scale instances, which serve as near-optimal labels for training a stronger model at larger scales. Theoretically, we show that under mild conditions the proposed edge-output GNN class is expressive enough to
represent the optimal product-assignment mapping, justifying the edge-level learning target. Numerical experiments show that the fastest proposed policy delivers 13--21\% higher profit than bundle-size pricing on instances with up to 100 products at about 2\% of its runtime. The framework shows \textit{when} and \textit{why} learning creates value in bundle pricing: prediction is used not to replace optimization but to identify where optimization effort should be concentrated, preserving the pricing rigor of exact mixed bundling while breaking its scalability barrier. As product catalogs grow beyond the reach of exact labels, the self-improvement procedure lets the deployed model generate its own retraining supervision, keeping the GNN maintainable at scale.

\end{abstract}
\section{Introduction}
\begin{figure}[h]
     \centering
     \includegraphics[width=0.8\linewidth]{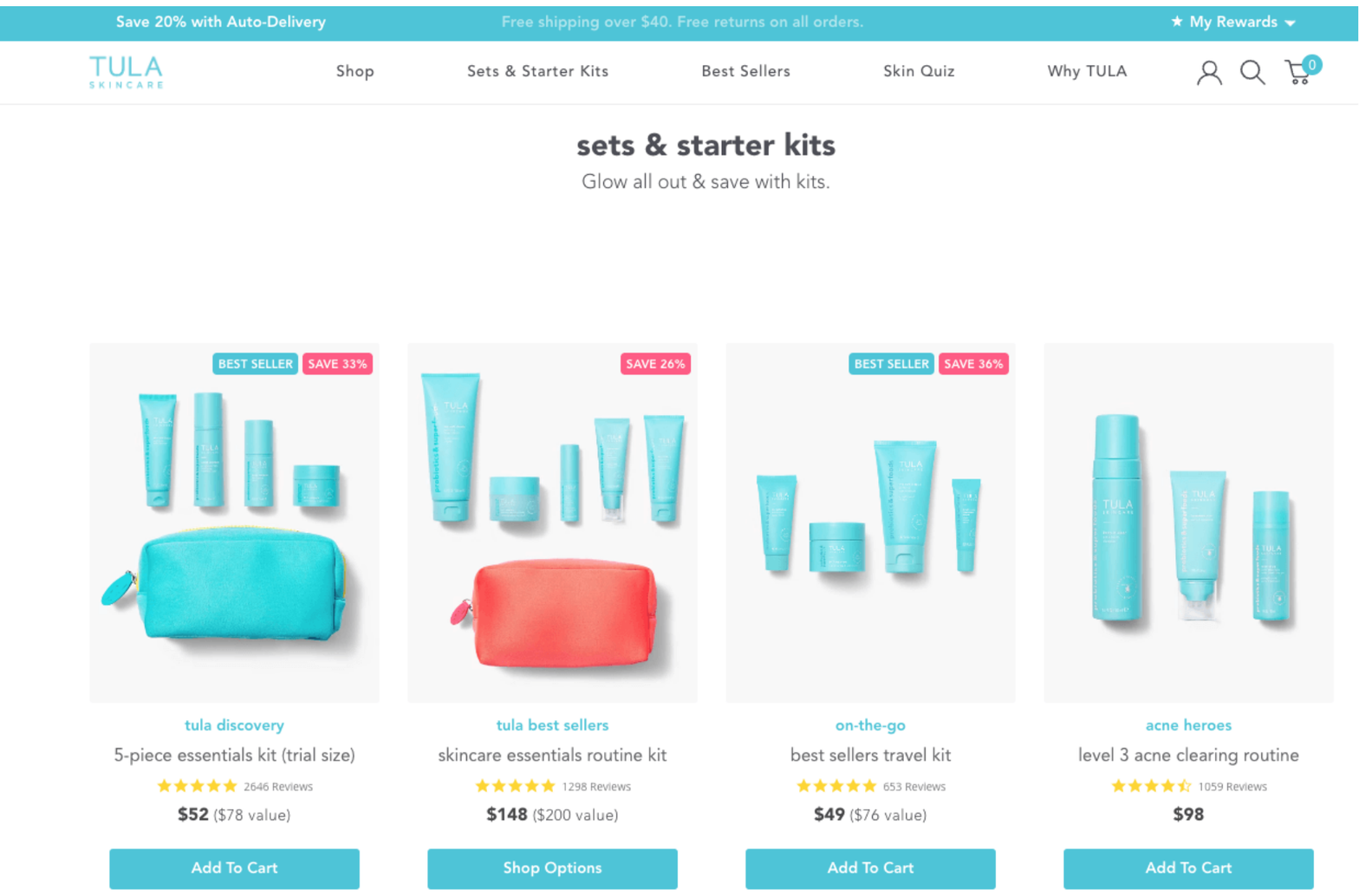}
     \caption{An illustration of a mixed bundling strategy in e-commerce (Tula Skincare). The retailer offers various ``sets and starter kits'' while explicitly displaying the discounted bundle price alongside the original component value to influence customer choice.}
     \label{fig:tula}
 \end{figure}

Bundle pricing is a widely adopted strategy across industries such as e-commerce, digital subscriptions, and retail. It refers to the practice where a firm provides combinations (i.e., ``\textit{bundles}'') of products or services at discounted prices, supplementing the traditional component pricing (CP) strategy where products are only sold separately. For instance, brands like Tula Skincare actively employ a mixed bundling strategy by offering individual items alongside a range of curated sets and starter kit bundles.\footnote{https://www.ordergroove.com/blog/product-bundling/} As illustrated in Figure~\ref{fig:tula}, these bundles explicitly highlight both the percentage and absolute dollar savings to clearly communicate value and incentivize larger purchases. In the e-commerce sector, industry reports suggest that KIND Snacks reported a 24\% increase in average order value (AOV) after introducing ``build-your-own'' bundles that empower customers to actively select their own personalized product combinations, while Peet's Coffee drove a 27\% increase in new subscribers within a single year through targeted subscription bundles. To implement this strategy effectively, a firm needs to solve a complex optimization problem: determining which subsets of products to offer and setting their corresponding prices to maximize total profit, under the constraint that heterogeneous customers self-select the options that maximize their own surplus.

The difficulty of the bundle pricing problem stems from two intertwined challenges: the combinatorial explosion of the search space and the complex customer dynamics. In particular, with $n$ products, the number of possible bundles grows exponentially ($2^n$), and customers strategically choose the bundle that maximizes their individual surplus. This self-selection behavior creates highly coupled constraints across all offered bundles, as a price change in one bundle can drastically alter the demand for others. Classical formulations, such as the mixed bundling (MB) model by \citet{hanson1990optimal}, rigorously capture these customer surplus constraints but become computationally intractable as $n$ increases (e.g., $n>15$). To address these computational bottlenecks, researchers have proposed various approximation strategies aimed at improving scalability. A notable example is bundle size pricing \citep{chu2011bundle}, which simplifies the problem by enforcing identical prices for all bundles of the same size. However, because such methods rely on strong simplifying assumptions---such as completely ignoring product heterogeneity---both their runtime efficiency and final profit performance remain fundamentally limited in large-scale settings.

In recent years, neural networks have demonstrated remarkable success in ``Learning to Optimize'' (L2O) for combinatorial problems, such as mixed-integer linear programming (MILP) solving via neural branching \citep{gasse2019exact} or neural diving \citep{nair2020solving}. Motivated by these advances, one might consider leveraging these neural-network-based MILP solvers to accelerate the solution process. However, they share a fundamental dependency on the explicit graph representation of the MILP's variable-constraint matrix. In the bundle pricing context, the number of potential bundles $2^n$ leads to a prohibitive memory bottleneck. For any large product size $n$, the resulting bipartite graph becomes too massive to be stored or processed by a GNN, rendering these neural heuristics computationally infeasible at the very stage of model initialization.

To overcome these challenges, we investigate how to design a scalable GNN framework that learns to identify high-quality bundle structures directly from data, without constructing the explicit exponential graph. By doing so, our framework integrates learning with optimization: a GNN predicts where the optimal solution concentrates, and the exact mixed bundling formulation is then solved only over that predicted region. Specifically, we aim to address the following research questions:
\begin{enumerate}
    \item[(1)] \textit{\textbf{Representation:} How can we leverage graph neural networks to extract features from the bundle pricing problem without explicitly constructing the exponentially large variable-constraint graph?}
    \item[(2)] \textit{\textbf{Search Space Reduction:} Given the neural network predictions, how can we design inference policies that effectively prune the intractable $2^n$ search space into a manageable subset?}
    \item[(3)] \textit{\textbf{Scalability:} How does the performance of the neural-network-based approach scale with problem size?}
\end{enumerate}

To address the first research question, we propose a GNN-based framework that learns to identify the latent structure of the bundle pricing problem and hence provides high-quality solutions. Leveraging the structural properties of this problem, we construct a bipartite graph representing these segment-product interactions, which scales linearly with the number of segments and products. We then train a GNN on this compact representation to extract latent structural features characterizing segment-product affinities. These features are decoded to predict the inclusion-probability matrix, representing the probability that each customer segment will purchase each product in the optimal solution. 
Theoretically, we show that, under mild distinguishability conditions, the
edge-output GNN class can represent the optimal product-assignment mapping and make the binary cross-entropy loss arbitrarily small.

To address the second research question, we design pruning-based inference policies that translate the GNN output into a restricted bundle family. Fixed cutoff pruning (FCP) forms one candidate bundle per segment by retaining products whose predicted probabilities exceed a cutoff. Progressive cutoff pruning (PCP) is more conservative: for each segment, it ranks retained products by predicted probability and constructs a sequence of nested prefix bundles. In both cases, the seller then solves the mixed bundling formulation of \citet{hanson1990optimal} over the retained candidate family for the joint optimization of bundle offerings and pricing. To further refine the solution quality, we incorporate a GNN-guided local search method that strategically adds or drops products from the assigned bundles.

To address the third research question, we evaluate the framework across problem sizes. Specifically, the GNN's weight-sharing capability enables us to train the model on tractable small-scale instances (where optimal labels are efficiently computable) and directly apply it to intractable large-scale ones. Because size generalization from small exact labels alone is limited, we further introduce an iterative self-improvement procedure: a GNN trained on small optimal labels is first used to generate high-quality near-optimal labels on larger instances, and these near-optimal labels are then used to train a larger-scale model. This procedure expands the training regime without requiring exact large-scale mixed bundling labels. Numerical experiments are provided to illustrate its effectiveness. 

Our contributions are threefold.
First, we introduce a compact learning representation for mixed bundling. Instead of predicting over the exponential bundle space, we learn the segment-product projection of an optimal mixed bundling solution on a segment-product graph. Second, we develop a GNN-guided pruning-then-optimization framework, together with a GNN-guided local-search procedure to further refine the candidate family, and an iterative self-improvement framework to further enhance the GNN performance on large-scale instances.
Third, we provide theoretical and computational evidence for the proposed framework. Theoretically, we establish an expressiveness result for edge-output GNNs under mild conditions. Computationally, we show that the proposed policies achieve near-optimal profits on small instances and outperform scalable bundle-pricing benchmarks on larger instances. 

\section{Literature Review}

This work is broadly related to three streams of research: bundle pricing, neural-network-based optimization, and bundle recommendation.

\textbf{Bundle pricing.} The study of bundle pricing has a long history in economics and operations
research communities. Early work in the two-product setting established that
bundling can be more profitable than separate sales, including pure bundling
(PB), component pricing (CP), and mixed bundling (MB)
\citep{stigler1963united,adams1976commodity,schmalensee1984gaussian}. Later
research extends these insights to large-scale settings. A stream of work
analyzes simplified mechanisms such as PB and bundle-size pricing (BSP)
\citep{bakos1999bundling,hitt2005bundling,chu2011bundle,abdallah2019benefit}.
In particular, PB can approximate the revenue of MB when the number of products
grows large under zero costs and independent additive valuations
\citep{bakos1999bundling}, while later work studies large-scale bundling with
positive production costs \citep{abdallah2019benefit}. Empirical and
theoretical studies show that BSP can outperform CP and PB in certain
conditions and can achieve asymptotic optimality when product costs are
homogeneous, though it may deteriorate under cost heterogeneity
\citep{hitt2005bundling,chu2011bundle, abdallah2021large, li2021convex}. More recently, novel mechanisms have been proposed to address different
limitations of classical bundling schemes. Pure bundling with disposal for cost
(PBDC) mitigates the adverse effect of high production costs
\citep{ma2021reaping}; component pricing with bundle-size discounts
(CPBSD) subsumes CP, PB, BSP, PBDC, and two-part tariffs as special cases and
inherits constant-factor guarantees under additive independent valuations
\citep{chen2025component}; and \citet{sun2025partition} study the design and pricing of a single bundle
offered together with individually sold remaining products, combining
structural analysis with optimization-based algorithms. Complementary to mechanism design, a computational
line of work models bundle pricing directly as a mixed-integer program. The
seminal MB formulation explicitly models customer self-selection and price
subadditivity constraints \citep{hanson1990optimal}, though its scalability is
limited by the exponential growth of candidate bundles. Our work partially follows this
computational tradition: we retain the mixed bundling framework while using
graph-based learning to construct a restricted candidate bundle family before
invoking the MILP solver, thereby combining the structural rigor of MB with
practical scalability.

\textbf{Neural-network-based optimization.} Recently, the use of machine learning to accelerate the solution of optimization problems has attracted increasing attention. For MILPs, a seminal work by \citet{gasse2019exact} encodes an instance as a constraint--variable bipartite graph and trains a GCN to imitate strong branching decisions. Subsequent studies learn primal heuristics, branching, diving, cut selection, separator configuration, and local-search decisions within B\&B or branch-and-cut \citep{ding2020accelerating,nair2020solving,gupta2022lookback,paulus2022learning,li2023separators}. 
A related predict-then-optimize line uses neural predictions to restrict or guide the subsequent search: \citet{han2023gnn} predict marginal variable probabilities and search around the predicted solution, \citet{liu2025apollo} alternate prediction and correction to fix reliable variables, and \citet{chen2026pmvb} use probabilistic multi-variable branching to partition MIP feasible regions. 
On the theory side, some researchers study the capability of GNNs (see
\citealt{zhang2024expressive} for a comprehensive review). For example, \citet{chen2023lp}
show that GNNs can represent key properties of linear programs, including
feasibility, boundedness, and optimal solutions. For MILPs,
\citet{chen2023milp} characterize the expressive limitations of standard
message-passing GNNs and identify conditions under which these limitations can
be overcome. Our theoretical result for GNNs is different in an important way:
rather than representing and predicting the solution of the full
mixed bundling MILP over the exponential bundle-level decision space, we learn
a polynomial-size product-assignment projection of the optimal solution, with
one edge-level prediction for each segment-product pair.

Machine learning has also been directly integrated into the core structure of
operations research problems. Attention-based and reinforcement-learning models
have been developed for routing problems such as TSP and VRP
\citep{kool2019attentionlearnsolverouting}. For example, \citet{li2025smalllargegraphconvolutional} leverage the generalizability of GCNs to solve large-scale NP-hard constrained assortment optimization problems. \citet{guo2025solving} integrate first-order methods with a self-supervised neural network framework to solve constrained assortment optimization problems. \citet{yang2026neural} propose a choice-model-agnostic neural
assortment optimizer that combines continuous extension, particle-based search,
and rounding to solve large-scale assortment optimization problems. In a similar vein,
we apply GNNs to the bundle pricing problem. Instead of constructing the
exponential bundle-level formulation, our GNN learns segment--product
interaction patterns on a compact segment-product graph and predicts
product-assignment probabilities, which are then converted into a restricted
candidate bundle family. This preserves the pricing and
self-selection framework of \citet{hanson1990optimal} while using GNN predictions to accelerate large-scale
instances. 

Recent works on neural combinatorial optimization have explored self-improvement methods that reduce reliance on externally generated expert labels. For example, \citet{luo2024selfimproved} improve model-generated routing solutions through local reconstruction and use the improved solutions as pseudo-labels. \citet{pirnay2024selfimprovement} sample multiple complete solutions from the current policy and imitate the best one. Both approaches operate entirely within the learned policy such that the label quality is bounded by what the network itself can construct. In contrast, our self-improvement scheme is optimization-in-the-loop: the GNN supplies only pruning guidance, and each retraining label is obtained by solving the mixed bundling formulation exactly over the pruned candidate family. The generated label thus inherits the optimality of the restricted MILP, which is particularly valuable in bundle pricing due to the exponential bundle space.

\textbf{Bundle recommendation.} Our work is also related to bundle recommendation; we refer readers to 
\citet{chang2021bundle}, \citet{zhang2024customizing}, \citet{sun2026survey} and references therein for comprehensive reviews and representative methods. Recommendation research typically predicts user preference scores or personalized rankings for bundles from historical user--item, user--bundle, and bundle--item interactions, and is usually evaluated by ranking metrics such as Recall and NDCG. This differs from bundle pricing in two essential ways. First, recommendation methods usually treat price as exogenous or absent and therefore do not optimize a market-wide price menu. Second, they do not impose self-selection constraints across customer segments. Introducing pricing makes the problem substantially harder: in a full mixed bundling formulation, the seller may need a price variable for each of the exponential number of possible bundles, and each price can change all customers' surplus-maximizing choices.

\section{Problem Definition}
\label{sec:problem}

We study a bundle pricing problem where a seller offers $n$ products
to $m$ customer segments. 
For any positive integer \(f\), let \([f]:=\{1,\dots,f\}\) and let \([f]_+:=\{0,\dots,f\}\). 
The product set is \([n]\), the customer-segment set is \([m]\), and
\(
\mathfrak F:=2^{[n]}
\)
denotes the full family of product bundles, including the empty bundle. We use $\mathcal K:=[2^n-1]_+$
as the corresponding bundle index set. We fix a bijection
\(\mathcal A:\mathcal K\to\mathfrak F,\)
where \(\mathcal{A}(b)\subseteq[n]\) denotes the product set represented by bundle index \(b\).
A candidate bundle family is denoted by
\(
\mathfrak B\subseteq\mathfrak F,
\)
and its associated index set is
\(
\mathcal I_{\mathfrak B}
:=
\{b\in\mathcal K: \mathcal{A}(b)\in\mathfrak B\}.
\)
The full mixed bundling problem corresponds to \(\mathfrak B=\mathfrak F\), or equivalently
\(\mathcal I_{\mathfrak B}=\mathcal K\). 

Each customer segment \(k \in [m]\) represents a cohort of homogeneous consumers, characterized by a proportion \(\alpha_k\) and a distinct valuation vector
\(
\mathbf v_k=\{R_{kb}\mid b\in\mathcal K\},
\)
where \(R_{kb}\) denotes segment \(k\)'s maximum willingness-to-pay for the product set \(\mathcal{A}(b)\). The index \(0\) corresponds to the empty bundle, representing the outside option, and we normalize $R_{k0}=0$ for $k\in[m]$.
For a candidate bundle family \(\mathfrak B\), the seller chooses the prices
\(
\{p_b\mid b\in\mathcal I_{\mathfrak B}\}
\)
to maximize the total profit, which can be defined as
\(
\sum_{k=1}^{m}\sum_{b\in\mathcal I_{\mathfrak B}}
(p_b-c_{kb})\cdot \alpha_k \theta_{kb}.
\)
Here, \(c_{kb}\) is the serving cost of selling bundle index \(b\) to segment \(k\), and
\(\theta_{kb}\in\{0,1\}\) is a binary decision variable denoting whether segment \(k\) selects bundle index \(b\).

After observing all displayed bundles in \(\mathfrak B\), each buyer selects exactly one bundle index \(b\in\mathcal I_{\mathfrak B}\) that maximizes its individual surplus,
defined by
\(
\Delta_{kb}=R_{kb}-p_b,
\)
provided that the surplus is non-negative. We follow the assumption of free disposal of unwanted products within a bundle \citep{hanson1990optimal}. Under this setting, customers can discard unwanted items at no cost. 
Consequently, if a superset bundle \(b_2\in\mathcal I_{\mathfrak B}\) were
priced lower than its subset bundle \(b_1\in\mathcal I_{\mathfrak B}\), i.e.,
\(
\mathcal A(b_1)\subseteq \mathcal A(b_2)
\) and \(
p_{b_2}<p_{b_1},
\)
then rational customers desiring the product set \(\mathcal A(b_1)\) could simply
purchase bundle \(b_2\), discard the items in \(\mathcal A(b_2)\setminus \mathcal A(b_1)\), and
effectively obtain \(\mathcal A(b_1)\) at a lower price. Thus, to preclude such
arbitrage opportunities, bundle prices should be non-decreasing with respect
to set inclusion.

\begin{assumption}[Price Monotonicity]
\label{ass:price_monotonicity}
For any two bundle indices
\(b_1,b_2\in\mathcal I_{\mathfrak B}\), if
\(\mathcal A(b_1)\subseteq \mathcal A(b_2)\), then
\(
p_{b_1}\le p_{b_2}.
\)
\end{assumption}

Under these assumptions, for a candidate bundle family \(\mathfrak B\), the seller's profit optimization problem can be formulated as follows:
\begin{align}
\max \quad &
\sum_{k=1}^m \sum_{b \in \mathcal I_{\mathfrak B}}
(p_b - c_{kb}) \cdot \alpha_k \, \theta_{kb}  \\
\text{s.t.} \quad &
\theta_{kb} \leq
\mathbbm{1}\!\left(
b \in \arg\max_{b' \in \mathcal I_{\mathfrak B}}
\{R_{kb'} - p_{b'}\}
\ \wedge\
R_{kb} - p_b \ge 0
\right),
\quad \forall k,\ b \in \mathcal I_{\mathfrak B} \\[6pt]
&
\sum_{b \in \mathcal I_{\mathfrak B}} \theta_{kb} = 1,
\quad \forall k  \\[6pt]
&
p_b \ge 0,
\quad \forall b \in \mathcal I_{\mathfrak B}.
\end{align}
Here, $\mathbbm{1}(\cdot)$ is the indicator function that enforces buyers' surplus maximization choice. 

In the following discussion, we set \(\mathcal A(0)=\emptyset\).
For any bundle index \(b\in\mathcal K\), we define the
valuation of segment \(k\) as
\(
R_{kb}
=
f\!\left(\sum_{j\in \mathcal{A}(b)}u_{kj}\right),
\)
where \(u_{kj}\) is the utility of product \(j\) for segment \(k\), and
\(f(\cdot)\) is an increasing concave function with \(f(0)=0\). 
For any bundle index
\(b\in\mathcal K\setminus \{0\}\), we assume that the serving cost is
\(
c_{kb}= \sum_{j\in \mathcal{A}(b)} c_j^{\mathit u}+c_k^s,
\)
where \(c_k^s\) is the cost associated with serving customer segment \(k\)
(e.g., the shipping cost) and \(c_j^{\mathit u}\) is the unit cost of
product \(j\). For the outside option, \(c_{k0}=0\) for all \(k \in [m]\). The concavity of $f(\cdot)$ captures the economic consensus of diminishing marginal utility, and the cost structure captures the principle of economies of scale: since the fixed cost is spread across all items in the bundle, the average cost per item decreases as more products are added. 

To solve this problem, \citet{hanson1990optimal} formulate a mixed-integer linear program (MILP) with an exponential number of variables. For the sake of completeness, we describe this formulation in the following.

\subsection{Mixed Bundling Formulation \citep{hanson1990optimal}}
\label{app:hanson}
Consider a candidate bundle family
\(
\mathfrak B\subseteq\mathfrak F
\)
with associated index set
\(
\mathcal I_{\mathfrak B}
=
\{b\in\mathcal K: \mathcal{A}(b)\in\mathfrak B\}.
\)
Define
\(
R_{\max}:=\max_{k\in[m],\, b\in\mathcal K} R_{kb}.
\)

We define the following variables: \(\theta_{kb}\) is a binary indicator for whether segment \(k\) chooses bundle index \(b\); \(p_b\) is the price of bundle index \(b\); \(P_{kb}\) is the effective price paid by segment \(k\); \(s_k\) is the surplus of segment \(k\); \(Z_{kb}\) is the profit from assigning bundle index \(b\) to segment \(k\); and \(\mathcal{A}(b)\) is the set of components which define bundle index \(b\). 
We denote \(\boldsymbol{\Theta}:=[\theta_{kb}]\), \(\mathbf p:=(p_b)\), \(\mathbf P^{\mathrm{pay}}:=[P_{kb}]\), \(\mathbf s:=(s_k)\), and \(\mathbf Z:=[Z_{kb}]\). 
The mixed bundle pricing problem over \(\mathfrak B\) can be written as follows:
\begin{align}
\small
\Xi(\mathfrak B):=\max \quad & \sum_{k=1}^m \sum_{b \in \mathcal I_{\mathfrak B}} \alpha_k \cdot Z_{kb} \\
\text{s.t.} \quad 
 & \sum_{b \in \mathcal I_{\mathfrak B}} \theta_{kb} = 1, && \forall k \\
 & s_k \ge R_{kb} - p_b, && \forall k,b \in \mathcal I_{\mathfrak B}   \\
& p_b - R_{\max}(1 - \theta_{kb}) \leq P_{kb}, && \forall k, b \in \mathcal I_{\mathfrak B} \\
& P_{kb} \leq p_b, && \forall k, b \in \mathcal I_{\mathfrak B} \\
% & P_{kb} \geq p_b - R_{\max}(1 - \theta_{kb}), && \forall k,b\\
& s_k = \sum_{b \in \mathcal I_{\mathfrak B}} (R_{kb}\theta_{kb} - P_{kb}), && \forall k \\
& R_{kb}\theta_{kb} - P_{kb} \ge 0, && \forall k, b \in \mathcal I_{\mathfrak B} \\
& s_k \geq \sum_{b \in \mathcal I_{\mathfrak B}} (R_{kb}\theta_{k'b} - P_{k'b}), && \forall k,k' \\
& Z_{kb} = P_{kb} - c_{kb}\theta_{kb}, && \forall k, b \in \mathcal I_{\mathfrak B} \\
& p_b \le \sum_{c\in\mathcal C} p_c,
&& \forall b\in\mathcal I_{\mathfrak B},\
\forall \mathcal C\subseteq \mathcal I_{\mathfrak B}
\text{ with }
\mathcal{A}(b)\subseteq \bigcup_{c\in\mathcal C} \mathcal A(c)  \label{eq:hm_cover_subadditivity}
\\
& p_b, P_{kb},s_k \geq 0, R_{k0}\theta_{k0} = P_{k0}, \theta_{kb} \in \{0,1\} 
\end{align}

The full mixed bundling formulation is obtained as the special case
\(
\Xi(\mathfrak F),
\mathfrak F=2^{[n]},
\)
where all product subsets are offered as candidate bundles. 
A restricted candidate family should be interpreted as a new pricing problem
over the retained menu, not simply as the set of bundles purchased in a
full-menu optimum. Appendix~\ref{app:restricted_menu_two_stage} provides an example illustrating this distinction.

In \(\Xi(\mathfrak B)\), price subadditivity is imposed in the cover form
\eqref{eq:hm_cover_subadditivity}: a retained collection
\(\mathcal C\subseteq\mathcal I_{\mathfrak B}\) covers a target bundle
\(b\in\mathcal I_{\mathfrak B}\) if
\(\mathcal A(b)\subseteq\bigcup_{c\in\mathcal C}\mathcal A(c)\), and the model
requires \(p_b\le\sum_{c\in\mathcal C}p_c\). This form is used for restricted
candidate families because they are generally not closed under partitions; see
Appendix~\ref{app:subadd} for its connection to full-universe partition
subadditivity.

We note that the main computational challenge of the above formulation lies in the exponential number of variables corresponding to all possible bundles. In the following, we demonstrate how the GNN framework can be leveraged to prune the search space and help solve large-scale bundle pricing problems efficiently.

\section{GNN-Based Strategies}

In this section, we design GNN-based strategies to solve the optimal bundle pricing problem. Specifically, we employ a GNN architecture characterized by an edge-centric update mechanism. Unlike standard graph convolutional networks (GCNs) that primarily focus on learning node embeddings over static edge attributes, our framework executes a dual update process: it iteratively refines both node and edge embeddings. Our model predicts a segment--product inclusion probability matrix $\mathbf{P} \in \mathbb{R}^{m\times n}$, where $\mathbf{P}_{kj}$ is the predicted probability that customer segment $k$ selects product $j$, derived directly from the final learned edge embeddings. These probabilities serve as a compact representation of heterogeneous preferences and provide the basis for pruning the exponentially large bundle space. We describe our strategies in detail in the following sections.

\subsection{Motivation for GNN} 
We motivate our choice of architecture by analyzing the structural limitations of standard Multilayer Perceptrons (MLPs) in the context of bundle pricing, and explaining how GNNs effectively address these challenges:

\paragraph{Limitations of MLPs.} Standard feed-forward networks face two fundamental bottlenecks when applied to combinatorial problems of this nature:
\begin{enumerate} 
\item \textbf{Inability to Generalize Across Problem Sizes}: MLPs are constrained by fixed input and output dimensions. Consequently, instances with different numbers of products or customer segments typically require training separate models. While zero-padding can accommodate smaller instances, this approach is inefficient and fails to generalize to instances larger than the training configuration. In contrast, GNNs operate on graph topologies of arbitrary size using shared weights, allowing for seamless scalability. 
\item \textbf{Lack of Permutation Equivariance}: MLPs treat inputs as flattened feature vectors and lack the inductive bias to capture the relational structure among agents. In our context, reordering product or customer indices should not alter the optimal product assignment (i.e., the target mapping is permutation equivariant). However, because MLPs assign weights based on specific input positions, permuting the input order arbitrarily changes the output. GNNs naturally respect this symmetry by applying the same local operations across all nodes and edges, ensuring consistent predictions regardless of indexing. \end{enumerate}

\paragraph{Advantages of GNNs.} In contrast, GNNs operate directly on graph-based representations, where nodes correspond to entities (products and customer segments) and edges encode their interactions (utilities).
\begin{enumerate} 
\item \textbf{Scalability through Weight Sharing:} A key feature of GNNs is the weight-sharing mechanism, where the same learnable weight matrices are applied across all nodes and edges. This decouples the number of parameters from the graph size, allowing a model trained on small-scale instances to generalize effectively to large-scale instances without retraining.
\item \textbf{Context-Aware Representations:} Through message passing, GNNs explicitly model the bipartite interactions. Each node iteratively aggregates information from its neighbors, building a representation that reflects both its local attributes (e.g., costs) and its structural context (e.g., demand from connected segments). This enables the model to capture the latent patterns of optimal bundles in a manner that is both scalable and permutation invariant.
\end{enumerate}

\subsection{Graph Representation}
\label{subsec:GraphRepresentation}
\begin{figure}[h]
\centering
\includegraphics[width=0.4\linewidth]{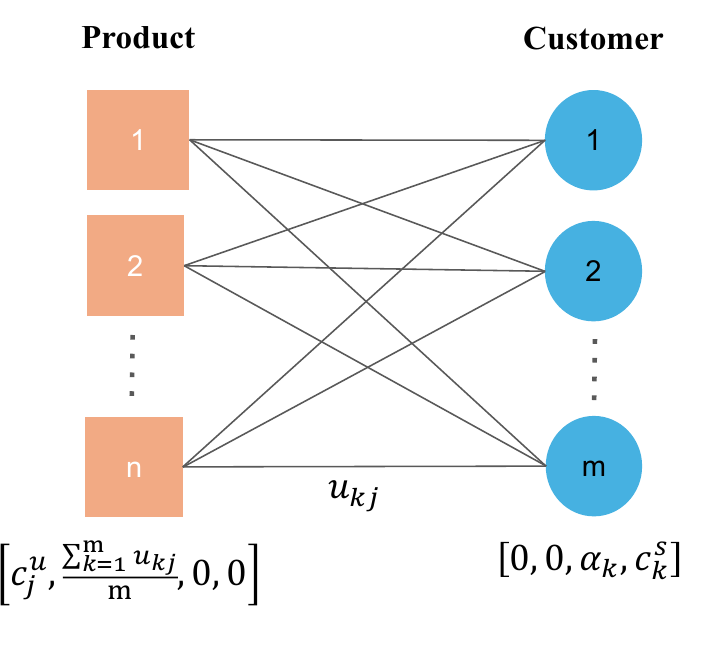}
\caption{Graph representation of the bundle pricing problem with customer and product nodes.}
\label{fig:GR}
\end{figure}

Figure~\ref{fig:GR} illustrates the graph representation of the optimal bundle pricing problem, which serves as the input to our Graph Neural Network (GNN). The graph consists of two types of nodes: product nodes ($\square$) and customer segment nodes ($\bigcirc$), as shown in Figure~\ref{fig:GR}.

Let $\mathcal{V}$ and $\mathcal{E}$ denote the sets of nodes and edges, respectively. We define the raw features for the problem instance as follows:
\begin{itemize}
    \item \textbf{Product Features:} Let $\mathbf{y}_j^P \in \mathbb{R}^4$ denote the initial feature vector for product \(j\in[n]\). It is constructed as $\mathbf{y}_j^P=\bigl[c_j^{\mathit{u}},\;\frac{1}{m}\sum_{k=1}^m u_{kj},\;0,\;0\bigr]$.
    \item \textbf{Customer Features:} Let $\mathbf{y}_k^S \in \mathbb{R}^4$ denote the initial feature vector for customer segment \(k\in[m]\). It is constructed as $\mathbf{y}_k^S=\bigl[0,\;0,\;\alpha_k,\;c_k^s\bigr]$.
    \item \textbf{Edge Features:} Each edge connecting product $j$ and customer segment $k$ is characterized by the scalar utility value $u_{kj}$.
\end{itemize}

\subsection{GNN Structure}
\label{subsec:GNN_structure}
We implement our model following the \textit{GNN framework} proposed by
\cite{battaglia2018relational}, utilizing the Generalized Graph Convolution
\citep{li2020deepergcn} for effective message aggregation. Specifically, we
construct an edge-output graph neural network consisting of \(\omega\) graph
network blocks followed by a shared scalar edge head. Each graph network block
follows the same message-passing template: node embeddings are first updated by
aggregating information from neighboring nodes and incident edges, and edge
embeddings are then updated using the newly updated endpoint node embeddings.
After \(\omega\) graph network blocks, the scalar edge head maps each final
product--segment edge embedding to a logit, which is subsequently transformed
into a segment-product inclusion probability by a sigmoid function.

\paragraph{General Layer Notation.}
Let $G^{(l)} = (\mathbf{V}^{(l)}, \mathbf{Z}^{(l)})$ denote the graph embeddings at layer $l$. 
The node feature matrix $\mathbf{V}^{(l)} \in \mathbb{R}^{(n+m) \times d_v^{(l)}}$ is defined as the vertical concatenation of the node feature vectors. The matrix is constructed as:
\[
    \mathbf{V}^{(l)} = 
    \begin{bmatrix}
    \mathbf{v}_1^{(l)} \\
    \vdots \\
    \mathbf{v}_{n+m}^{(l)}
    \end{bmatrix}.
\]
Here, $\mathbf{v}_w^{(l)} \in \mathbb{R}^{1 \times d_v^{(l)}}$ represents the embedding vector of the $w$-th node, and $d_v^{(l)}$ denotes the embedding dimensionality at layer $l$.

Furthermore, $\mathbf{V}^{(l)}$ can be explicitly partitioned into product and customer components. We denote $\mathbf{Y}^{P,(l)} \in \mathbb{R}^{n \times d_v^{(l)}}$ as the embedding matrix for the $n$ product nodes and $\mathbf{Y}^{S,(l)} \in \mathbb{R}^{m \times d_v^{(l)}}$ as the embedding matrix for the $m$ customer nodes. Thus, 
\(
    \mathbf{V}^{(l)} = 
    \begin{bmatrix}
    \mathbf{Y}^{P,(l)} \\
    \mathbf{Y}^{S,(l)}
    \end{bmatrix}.
\)

Similarly, the edge embedding tensor
\(
\mathbf{Z}^{(l)}\in\mathbb R^{n\times m\times d_e^{(l)}}
\)
collects the embeddings for all product--segment pairs. The relationship between the tensor \(\mathbf Z^{(l)}\) and an individual edge embedding vector \(\mathbf e_{jk}^{(l)}\in\mathbb R^{d_e^{(l)}}\) is
\(\mathbf{Z}^{(l)}[j,k,:]=\mathbf e_{jk}^{(l)}\), for \( j\in[n],\ k\in[m].\)
Here \(d_e^{(l)}\) denotes the edge embedding dimension at layer \(l\).
The input edge feature is one-dimensional \((d_e^{(0)}=1)\), while all graph network blocks output hidden edge embeddings of
dimension \(d_e^{(l)}=d_{\mathrm{mid}}\).
Therefore, after the \(\omega\)-th graph network block, the final hidden edge
embedding tensor is
\(
\mathbf Z^{(\omega)}
\in
\mathbb R^{n\times m\times d_{\mathrm{mid}}}.
\)
A shared scalar edge head, introduced
below, maps each vector \(\mathbf e_{jk}^{(\omega)}\in\mathbb R^{d_{\mathrm{mid}}}\)
to a scalar logit.

\paragraph{Initialization ($l=0$).}
The input features for the GNN are initialized directly from the raw feature vectors defined in Section~\ref{subsec:GraphRepresentation}. 
For node features, we map the product and customer features to the global node index $w$:
\[
    \mathbf{v}_w^{(0)} = 
    \begin{cases} 
        \mathbf{y}_w^P & \text{if } 1 \le w \le n \quad (\text{Product Node}) \\
        \mathbf{y}_{w-n}^S & \text{if } n < w \le n+m \quad (\text{Customer Node})
    \end{cases}
\]
For edge features, we initialize the tensor with the utility values 
\(\mathbf{e}_{jk}^{(0)} = [u_{kj}]\).

\begin{figure}[h]
\centering
\includegraphics[width=1\linewidth]{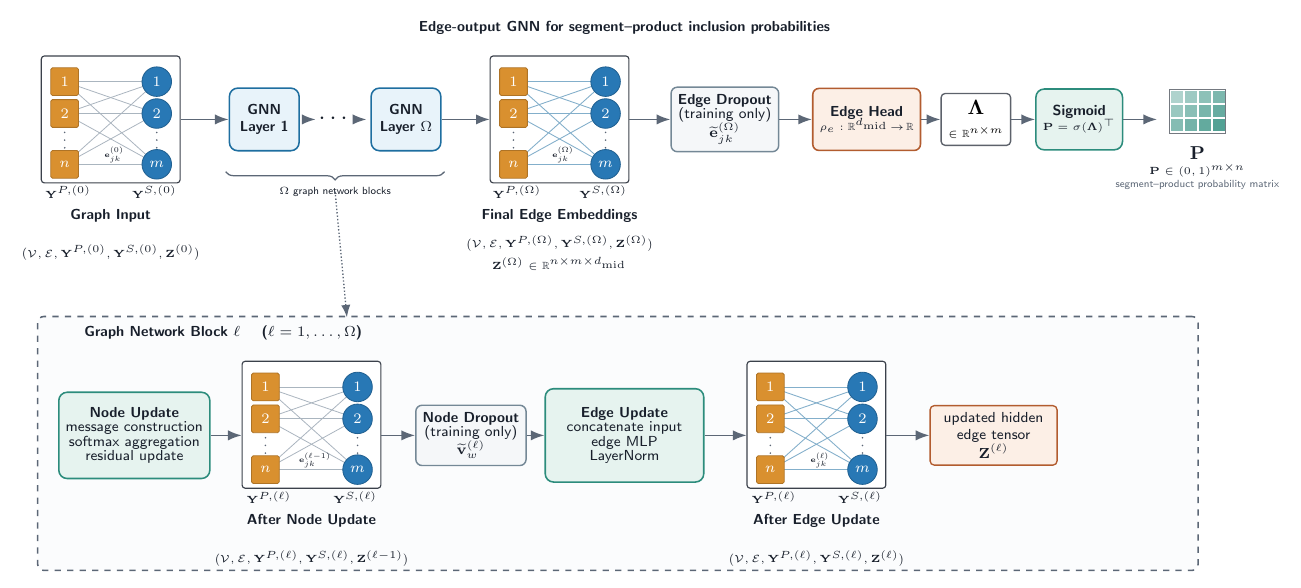}
\caption{Illustration of the Graph Neural Network.}
\label{fig:GNN}
\end{figure}

Each GNN layer executes two update functions sequentially: a node update followed by an edge update. We provide a schematic illustration of the GNN architecture in Figure~\ref{fig:GNN}. Below, we describe the computations performed in each layer.

\textbf{Node Update ($\phi^v$).} In the first stage of layer $l$, we update the embeddings of all nodes to capture the structural information from their neighbors. 
We employ the Generalized Graph Convolution of \citet{li2020deepergcn} as the update function. For a target node $w$ and its neighbor set $\mathcal{N}(w)$, the update consists of five operations:

\paragraph{1. Message Construction.}
For a directed message from neighbor \(z\) to
target node \(w\), we first project the source-node embedding, the target-node
embedding, and the edge embedding into a common hidden dimension
\(d_{\mathrm{mid}}\):
\(
\bar{\mathbf v}_{z\to w}^{(l-1)}
=
\mathbf v_z^{(l-1)}\mathbf W_{\mathrm{s}}^{(l)},
\bar{\mathbf v}_{w}^{(l-1)}
=
\mathbf v_w^{(l-1)}\mathbf W_{\mathrm{t}}^{(l)},
\bar{\mathbf e}_{zw}^{(l-1)}
=
\mathbf e_{zw}^{(l-1)}\mathbf W_{\mathrm{e}}^{(l)}.
\)
Here
\(
\bar{\mathbf v}_{z\to w}^{(l-1)},\ 
\bar{\mathbf v}_{w}^{(l-1)},\ 
\bar{\mathbf e}_{zw}^{(l-1)}
\in\mathbb R^{1\times d_{\mathrm{mid}}}.
\)
The matrices
\(
\mathbf W_{\mathrm{s}}^{(l)},
\mathbf W_{\mathrm{t}}^{(l)},
\mathbf W_{\mathrm{e}}^{(l)}
\)
are learnable parameters shared across all nodes and edges in layer \(l\).
If an input dimension already matches \(d_{\mathrm{mid}}\), the corresponding
projection can be interpreted as the identity map.

The message from \(z\) to \(w\) is constructed as
\[
\mathbf m_{z\to w}^{(l)}
=
\operatorname{ReLU}
\left(
\bar{\mathbf v}_{z\to w}^{(l-1)}
+
\bar{\mathbf e}_{zw}^{(l-1)}
\right)
+
\epsilon_{\mathrm{mp}}\mathbf 1,
\qquad \forall z\in\mathcal N(w).
\]
Here, \(\operatorname{ReLU}(\mathbf x)=\max\{\mathbf x,0\}\) is applied element-wise when
\(\mathbf x\) is a vector. The constant \(\epsilon_{\mathrm{mp}}>0\) is a fixed
numerical-stability constant and is not learned.

\paragraph{2. Softmax Aggregation.}
We aggregate incoming messages using a Softmax mechanism, which assigns
data-dependent weights to different neighbors:
\[
\hat{\mathbf m}_w^{(l)}
=
\sum_{z\in\mathcal N(w)}
\left(
\frac{
\exp\!\left(\beta_{\mathrm{mp}}\mathbf m_{z\to w}^{(l)}\right)
}{
\sum_{\gamma\in\mathcal N(w)}
\exp\!\left(\beta_{\mathrm{mp}}\mathbf m_{\gamma\to w}^{(l)}\right)
}
\right)
\odot
\mathbf m_{z\to w}^{(l)}.
\]
Here, \(\exp(\cdot)\), division, and multiplication are applied
coordinate-wise, and \(\odot\) denotes element-wise multiplication. The scalar
\(\beta_{\mathrm{mp}}>0\) is the inverse-temperature hyperparameter of the Softmax
aggregator.

\paragraph{3. Residual Update.}
The aggregated message is combined with the target node's own projected
embedding through a residual connection and passed through an MLP:
\(
\mathbf h_w^{(l)}
=
\operatorname{MLP}_{v}^{(l)}
\left(
\bar{\mathbf v}_{w}^{(l-1)}
+
\hat{\mathbf m}_w^{(l)}
\right),
\quad
\mathbf h_w^{(l)}\in\mathbb R^{1\times d_{\mathrm{mid}}}.
\)
An MLP is a feed-forward neural network obtained by composing affine
transformations and nonlinear activation functions. The same
\(\operatorname{MLP}_{v}^{(l)}\) is shared across all nodes in layer \(l\).

\paragraph{4. Activation.} 
We then apply a ReLU activation to the node-update
output
\(
\mathbf v_w^{(l)}
=
\operatorname{ReLU}
\left(
\mathbf h_w^{(l)}
\right).
\)

\paragraph{5. Node Dropout.}
To reduce overfitting during training, we apply dropout to the updated node
embeddings before they are used in the edge update. Let \(p_D\in[0,1)\) denote
the dropout probability. For each node \(w\) and feature coordinate \(r\), let
\(v_{wr}^{(l)}\) be the \(r\)-th entry of the updated node embedding
\(\mathbf v_w^{(l)}\). We define the post-dropout feature coordinate
\(\widetilde v_{wr}^{(l)}\) by
\[
\widetilde v_{wr}^{(l)}
=
\begin{cases}
\dfrac{\xi_{wr}^{(l)}}{1-p_D}v_{wr}^{(l)},
&
\text{during training},\\[3mm]
v_{wr}^{(l)},
&
\text{during validation and testing},
\end{cases}
\qquad
\xi_{wr}^{(l)}\sim \operatorname{Bernoulli}(1-p_D).
\]

\textbf{Edge Update (\(\phi^e\)).}
In the second stage of layer \(l\), we explicitly update the embedding of each
product--segment edge. Unlike standard GNNs where edge features may remain
static, our framework refines edge features based on the newly updated endpoint
node embeddings and the edge's own history.

For the edge connecting product \(j\) and customer segment \(k\), whose
customer node has global index \(n+k\), we construct the concatenated input
vector
\(
\mathbf z_{jk}^{(l)}
=
\widetilde{\mathbf v}_{j}^{(l)}
\Vert
\widetilde{\mathbf v}_{n+k}^{(l)}
\Vert
\mathbf e_{jk}^{(l-1)},
\)
where \(\Vert\) denotes concatenation, and
\(\widetilde{\mathbf v}_{j}^{(l)}\) and
\(\widetilde{\mathbf v}_{n+k}^{(l)}\) are the post-dropout endpoint node
embeddings. 
This vector is processed by a two-layer edge MLP followed by layer
normalization:
\[
\mathbf e_{jk}^{(l)}
=
\operatorname{LayerNorm}_{e}^{(l)}
\left(
\operatorname{ReLU}
\left(
\mathbf z_{jk}^{(l)}\mathbf W_{e,1}^{(l)}
+
\mathbf b_{e,1}^{(l)}
\right)
\mathbf W_{e,2}^{(l)}
+
\mathbf b_{e,2}^{(l)}
\right),
\qquad
\mathbf e_{jk}^{(l)}\in\mathbb R^{1\times d_{\mathrm{mid}}}.
\]
Here,
\(
\mathbf W_{e,1}^{(l)}
\in
\mathbb R^{(2d_{\mathrm{mid}}+d_e^{(l-1)})\times d_{\mathrm{mid}}},
\mathbf W_{e,2}^{(l)}
\in
\mathbb R^{d_{\mathrm{mid}}\times d_{\mathrm{mid}}},
\mathbf b_{e,1}^{(l)},\mathbf b_{e,2}^{(l)}
\in
\mathbb R^{1\times d_{\mathrm{mid}}}
\)
are learnable parameters of the edge update network. The parameters of
\(\operatorname{LayerNorm}_{e}^{(l)}\), defined below, are also learnable. The
same edge update network is shared across all product--segment edges in layer
\(l\).

Layer normalization, denoted by \(\operatorname{LayerNorm}_{e}^{(l)}\),
normalizes the coordinates of a feature vector and then applies learnable scale
and shift parameters. Specifically, for a vector
\(\mathbf{x}\in\mathbb R^{1\times d}\),
\[
\operatorname{LayerNorm}_{e}^{(l)}(\mathbf{x})
=
\boldsymbol{\gamma}_{e}^{(l)}
\odot
\frac{\mathbf{x}-\mu(\mathbf{x})\mathbf{1}}
{\sqrt{\sigma^2(\mathbf{x})+\varepsilon_{\mathrm{LN}}}}
+
\boldsymbol{\delta}_{e}^{(l)}.
\]
Here, \(\mu(\mathbf{x})\) and \(\sigma^2(\mathbf{x})\) are the mean and
variance of the coordinates of \(\mathbf{x}\),
\(\varepsilon_{\mathrm{LN}}>0\) is a fixed numerical-stability constant, and
\(\boldsymbol{\gamma}_{e}^{(l)},\boldsymbol{\delta}_{e}^{(l)}
\in\mathbb R^{1\times d}\) are learnable parameters.

\textbf{Output.}
After \(\omega\) graph network blocks, each product--segment edge has a final
hidden embedding
\(
\mathbf e_{jk}^{(\omega)}\in\mathbb R^{1\times d_{\mathrm{mid}}},
j\in[n],\ k\in[m].
\)
Equivalently,
\(
\mathbf Z^{(\omega)}
\in
\mathbb R^{n\times m\times d_{\mathrm{mid}}}
\)
collects all final hidden edge embeddings.

We apply dropout to the terminal edge
embeddings during training. For each edge coordinate \(r\in[d_{\mathrm{mid}}]\),
let \(e_{jk,r}^{(\omega)}\) be the \(r\)-th entry of
\(\mathbf e_{jk}^{(\omega)}\). Define
\[
\widetilde e_{jk,r}^{(\omega)}
=
\begin{cases}
\dfrac{\zeta_{jk,r}^{(\omega)}}{1-p_D}e_{jk,r}^{(\omega)},
&
\text{during training},\\[3mm]
e_{jk,r}^{(\omega)},
&
\text{during validation and testing},
\end{cases}
\qquad
\zeta_{jk,r}^{(\omega)}
\sim
\operatorname{Bernoulli}(1-p_D).
\]
Let \(\widetilde{\mathbf e}_{jk}^{(\omega)}\) denote the resulting
post-dropout terminal edge embedding. Dropout introduces no learnable
parameters; \(p_D\) is a hyperparameter.

A shared scalar edge head
\(
\rho_e:\mathbb R^{1\times d_{\mathrm{mid}}}\to\mathbb R
\)
then maps each terminal edge embedding to a scalar logit:
\[
\Lambda_{jk}
=
\rho_e\!\left(\widetilde{\mathbf e}_{jk}^{(\omega)}\right)
=
\widetilde{\mathbf e}_{jk}^{(\omega)}\mathbf w_o+b_o,
\qquad
j\in[n],\ k\in[m],
\]
where
\(
\mathbf w_o\in\mathbb R^{d_{\mathrm{mid}}\times 1},
b_o\in\mathbb R
\)
are learnable parameters shared across all product--segment edges. Collecting
these logits gives the product-by-segment logit matrix
\(
\boldsymbol\Lambda
=
[\Lambda_{jk}]_{j\in[n],\,k\in[m]}
\in
\mathbb R^{n\times m}.
\)

We apply the sigmoid function
entry-wise and transpose the result to obtain the segment-product probability
matrix
\[
\mathbf P
=
\sigma(\boldsymbol\Lambda)^\top
\in
(0,1)^{m\times n},
\]
where \(\sigma(x)=1/(1+\exp(-x))\). Thus,
\(
\mathbf P_{kj}
=
\sigma(\Lambda_{jk})
\)
is the predicted probability that product \(j\) is included in the bundle
purchased by segment \(k\).

\subsection{Training Process}
\label{subsec:GNN_training}
To train the GNN, we first collect historical problem instances, and then use the Mixed Bundling policy proposed by \citet{hanson1990optimal} to obtain the corresponding optimal bundle assignments $\theta_{kb}^{\star}$'s. We then
project these bundle assignments to segment-product binary labels where \(q_{kj}^{\star}=1\) if product \(j\) is included in the optimal bundle assigned to
segment \(k\), and \(q_{kj}^{\star}=0\) otherwise, i.e., $q_{kj}^{\star}:=\sum_{\substack{b: j\in \mathcal{A}(b)}} \theta_{kb}^{\star}$.
Specifically, our training procedure adopts a supervised learning paradigm, where the GNN is trained to predict selection probabilities for each segment-product pair, serving as a probabilistic predictor of the underlying binary product-selection decisions. Importantly, the model does not learn bundle prices directly; instead, it learns to approximate the optimal bundle assignment structure implied by the Mixed Bundling solution, which is leveraged to conduct high-quality search space pruning before solving the MILP, thereby accelerating the solution process by reducing computational complexity while maintaining near-optimal solution quality.

Thus, each instance $\ell$ is characterized by $(\mathbf{c}_{(\ell)}^u, \mathbf{c}_{(\ell)}^s, \mathbf{U}_{(\ell)}, \boldsymbol{\alpha}_{(\ell)}, f(\cdot), \mathbf{Q}^{\star}_{(\ell)})$, where $\mathbf{U}_{(\ell)} = [u_{kj,(\ell)}]_{m \times n}$ and $\mathbf{Q}^{\star}_{(\ell)} = [q^{\star}_{kj,(\ell)}]_{m \times n}$. The instances were split into a training set (80\%) and a validation set (20\%).
We then convert each problem instance into the corresponding graph representation and construct the input tensors $\mathbf{Y}^P$ and $\mathbf{Y}^S$. We formulate the edge prediction task as a binary classification problem. During each training epoch, we first pass the graphs in the training set to the Graph Neural Network to obtain the predicted product selection probability $\mathbf{P}_{kj}$'s. We compute the weighted binary cross-entropy loss $\mathcal{L}$. Let $S_T$ denote the number of instances in the training set. To address the inherent class imbalance (as optimal bundles typically contain only a small subset of available products), we introduce a positive weight parameter $w_{pos} = (1 - \kappa) / \kappa$, where $\kappa = \frac{\sum_{\ell=1}^{S_T}\sum_{k=1}^{m_{(\ell)}}\sum_{j=1}^{n_{(\ell)}} q_{kj,(\ell)}^{\star}}{\sum_{\ell=1}^{S_T} m_{(\ell)}n_{(\ell)}}$ is the proportion of positive edges across the entire training set. The loss function is defined as:
\[
\mathcal L
=
-\frac{1}{
\sum_{\ell=1}^{S_T}m_{(\ell)}n_{(\ell)}
}
\sum_{\ell=1}^{S_T}
\sum_{k=1}^{m_{(\ell)}}
\sum_{j=1}^{n_{(\ell)}}
\left[
w_{\mathrm{pos}}q_{kj,(\ell)}^{\star}
\log \mathbf P_{kj,(\ell)}
+
(1-q_{kj,(\ell)}^{\star})
\log(1-\mathbf P_{kj,(\ell)})
\right].
\]
The AdamW optimizer is used to update the model weights to minimize the training loss. The resulting model is then applied to the inference policies described in Section~\ref{subsec:inference_policy}.

\subsection{Pruning-based Inference Policies}
\label{subsec:inference_policy}
In this subsection, we leverage the probability matrix $\mathbf{P}$ predicted by the GNN model to guide two pruning strategies that construct a compact yet high-quality bundle space. The resulting reduced problem is then solved using the MILP formulation of \citet{hanson1990optimal}, but restricted to the pruned bundle set rather than the full exponential space. 

\paragraph{Fixed Cutoff Pruning (FCP).}
Our first policy generates one candidate bundle for each segment. For each
segment \(k\in[m]\), we construct the bundle by including product \(j\in[n]\)
whenever the predicted inclusion probability exceeds a fixed cutoff.
That is,
\[
B_k^{\mathrm{FCP}}
=
\{j\in[n]\mid \mathbf P_{kj}\ge 0.5\},
\qquad k\in[m],
\]
and the overall candidate bundle family is
\(
\mathfrak B^{\mathrm{FCP}}
=
\{B_k^{\mathrm{FCP}}:k\in[m]\}\cup\{\emptyset\}
\subseteq\mathfrak F.
\)
The threshold 0.5 is theoretically justified in
Section~\ref{sec:theory_main}; the sensitivity analysis further shows that it
provides a robust profit--runtime trade-off.
Here, the overall candidate family is shared across all segments, meaning
that any segment is free to select any \(B_k^{\mathrm{FCP}}\), not only its
own. This reduces the feasible space from \(2^n\) to \(O(m)\), making the
subsequent MILP much more tractable. We then solve
\(\Xi(\mathfrak B^{\mathrm{FCP}})\).

In the case where all probabilities of a segment fall below the cutoff, we retain the single product with the highest probability to avoid producing an empty bundle. The detailed procedure is given in Algorithm~\ref{alg:fcp} in Appendix~\ref{sec:pseudocode}.

\paragraph{Progressive Cutoff Pruning (PCP).}
The PCP policy is a more conservative pruning rule. For each segment \(k\),
we first retain products whose predicted probabilities exceed the cutoff
\(
U_k=\{j\in[n]:\mathbf P_{kj}\ge 0.5\},
\)
and sort them in descending order of \(\mathbf P_{kj}\):
\(
U_k=(j_{k,(1)},j_{k,(2)},\dots,j_{k,(|U_k|)}).
\)
We then construct a sequence of prefix bundles
\[
B_{k,i}^{\mathrm{PCP}}
=
\{j_{k,(1)},j_{k,(2)},\dots,j_{k,(i)}\},
\qquad i=1,\dots,|U_k|.
\]
The resulting candidate family is
\(
\mathfrak B^{\mathrm{PCP}}
=
\bigcup_{k=1}^m
\{B_{k,i}^{\mathrm{PCP}}:i=1,\dots,|U_k|\}
\cup\{\emptyset\}
\subseteq \mathfrak F.
\)
We then solve \(\Xi(\mathfrak B^{\mathrm{PCP}})\) over the associated index set
\(\mathcal I_{\mathfrak B^{\mathrm{PCP}}}\).

For PCP, the retained prefix bundles lead to a richer candidate family and
therefore more cover-form price-subadditivity constraints than FCP. We handle
these constraints using a cutting-plane implementation, with details provided
in Appendix~\ref{sec:cover_subadditivity_implementation}.

\begin{figure}[h]
    \centering
    \includegraphics[width=0.8\linewidth]{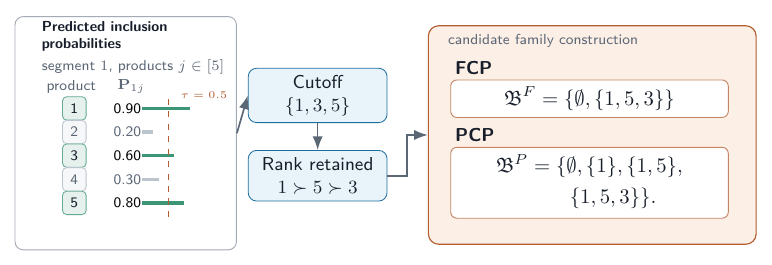}
    \caption{Candidate bundle constructions under FCP and PCP policies}
    \label{fig:candidate_bundle}
\end{figure}

Figure~\ref{fig:candidate_bundle} illustrates the construction of candidate
bundles under the FCP and PCP policies for a scenario with \(m_{\mathrm{test}}=1\)
and \(n_{\mathrm{test}}=5\). First, products 2 and 4 are excluded from
consideration because their predicted probabilities fall below the cutoff
\(\tau=0.5\). Under the FCP policy, the remaining high-probability products form
a single candidate bundle,
\(
B_1^{\mathrm{FCP}}=\{1,5,3\}.
\)
Thus, the resulting candidate bundle family is
\(
\mathfrak B^{\mathrm{FCP}}
=
\{\emptyset,\{1,5,3\}\}.
\)
In contrast, under the PCP policy, given the probability order
\(
\mathbf{P}_{11}>\mathbf{P}_{15}>\mathbf{P}_{13},
\)
the products are added sequentially to form a nested chain. Consequently, the
segment-specific PCP candidate family is
\(
\mathfrak B_1^{\mathrm{PCP}}
=
\bigl\{\{1\},\{1,5\},\{1,5,3\}\bigr\},
\)
and the global candidate bundle family is
\(
\mathfrak B^{\mathrm{PCP}}
=
\{\emptyset\}\cup\mathfrak B_1^{\mathrm{PCP}}.
\)

\paragraph{Pruning Effect on Model Size.}
The full mixed bundling formulation optimizes over the entire bundle family
\(\mathfrak F=2^{[n]}\), leading to \(O(m\,2^n)\) assignment-related variables
and self-selection constraints, together with exponentially many
price-subadditivity constraints. FCP provides the strongest compression by
retaining at most one candidate bundle per segment, so
\(|\mathfrak B^{\mathrm{FCP}}|\le m+1\). PCP is more conservative: it retains a
nested prefix chain for each segment, so
\(|\mathfrak B^{\mathrm{PCP}}|\le mn+1\), and uses cutting planes to avoid
pre-enumerating all cover-subadditivity constraints. Thus, both policies
replace the exponential bundle universe with polynomial-size restricted
families; FCP prioritizes efficiency, while PCP retains more pricing
flexibility.

\begin{table}[h]
\centering
\caption{Problem size comparison of MB, FCP, PCP, and BSP\, strategies}
\label{tab:problem_size}
\begin{tabular}{lccc}
\toprule
Policy & Effective Space & Variables & Total Constraints \\
\midrule
MB
& \(2^n\)
& \(O(m\,2^n)\)
& \(O(m\,2^n+m^2+3^n)\)
\\
FCP
& \(O(m)\)
& \(O(m^2)\)
& \(O(m^2+m\,2^m)\)
\\
PCP
& \(O(mn)\)
& \(O(m^2\,n)\)
& \(O(m^2\,n^2+mn\,(n+1)^m)\)
\\
BSP
& \(O(n)\)
& \(O(mn)\)
& \(O(mn+m^2+n^2)\)
\\
\bottomrule
\end{tabular}
\end{table}

\vspace{-0.1in}

\subsection{GNN-Guided Local Search Policy}
In order to further improve our solution, we propose an inference policy based on GNN-guided local search, which we denote FCPLS. The basic idea is to iteratively modify the bundle assignment by either adding an unselected product or dropping a selected product, and accept modifications if the profit increases.

To maximize the efficiency of the local search, we develop a \textit{global top-$K$ local search strategy} guided by the probability matrix $\mathbf{P}$ predicted by the GNN model. Instead of evaluating neighbors for all segments, we focus only on the most promising modifications globally and prioritize candidates based on a confidence score.
Specifically, for any segment $k$ and product $j$, we define the confidence score for a potential modification as:
\[
    \text{Score}_{kj} = 
    \begin{cases} 
        \mathbf{P}_{kj} & \text{if product } j \text{ is currently unselected (Candidate for Addition)} \\
        1 - \mathbf{P}_{kj} & \text{if product } j \text{ is currently selected (Candidate for Removal)}
    \end{cases}
\]
Intuitively, a higher $\mathbf{P}_{kj}$ indicates a strong suggestion to include the product, while a lower $\mathbf{P}_{kj}$ implies a strong suggestion to exclude it.

In each iteration, we select the top \(K\) candidate additions and the top
\(K\) candidate removals according to the GNN confidence scores. The selected
moves are pooled and evaluated in descending order of confidence; the first
improving move is accepted, and the search restarts from the updated
assignment. We use the sublinear budget \(K=\lceil\sqrt m\rceil\), with
supporting sensitivity analysis reported in Appendix~\ref{app:k_justification}.

The local search process terminates when no improvement is found among the top candidates in a full cycle, or when the predefined iteration limit is reached. Furthermore, we rely on the fixed-assignment LP to efficiently evaluate each neighbor, as it provides a valid lower bound on the restricted MILP optimum and significantly reduces computational cost. We demonstrate a high degree of consistency between LP and MILP objective improvements during the local search process, justifying the use of LP as a reliable proxy in Appendix~\ref{app:lp_consistency_analysis}. 

\subsection{Iterative Self-Improvement}
\label{subsec:iter_self_improve}

The preceding policies exploit a key advantage of GNNs: a model trained on
small instances can be applied to larger instances because its parameters are
shared across nodes and edges. This provides useful scalability and enables
small-to-large transfer. Nevertheless, this scalability can be limited. When the
product dimension increases substantially, the structure of the optimal bundle
family becomes richer, and a model trained only on small exact labels may be
less well calibrated for larger product spaces. As a result, its predicted
segment-product probabilities may become less accurate, leading to lower-quality
candidate families and weaker inference-policy performance.

A natural solution would be to train directly on large-scale optimal labels.
However, it is computationally infeasible to generate one exact label by solving
the full mixed bundling formulation \(\Xi(\mathfrak F)\), whose bundle space
grows exponentially in \(n\). We therefore propose an \textit{iterative
self-improvement strategy} that improves scalability stage by stage.
In particular, starting from a base GNN trained on small exact labels, we use its predictions to run
PCP on larger unlabeled instances. PCP policy is fast and
provide high-quality solutions for retraining. The retrained GNN can then be used
to generate solutions for even larger instances, so the procedure can in principle
be repeated over multiple stages. In this
sense, the method is a
bootstrapping scheme in which the current learned policy generates improved
large-scale supervision for the next model. An illustration of this method is shown in Figure~\ref{fig:self_improve}, while the detailed algorithm is given in Appendix~\ref{sec:pseudocode}.  In the numerical experiments, we report the
effect of one self-improvement round to show the effect of this step. The numerical results in
Section~\ref{subsec:exp_iter} show that this
self-improved model substantially improves transfer to larger product spaces
while preserving the computational efficiency of the pruning framework.

\begin{figure}[h]
    \centering
    \includegraphics[width=0.9\linewidth]{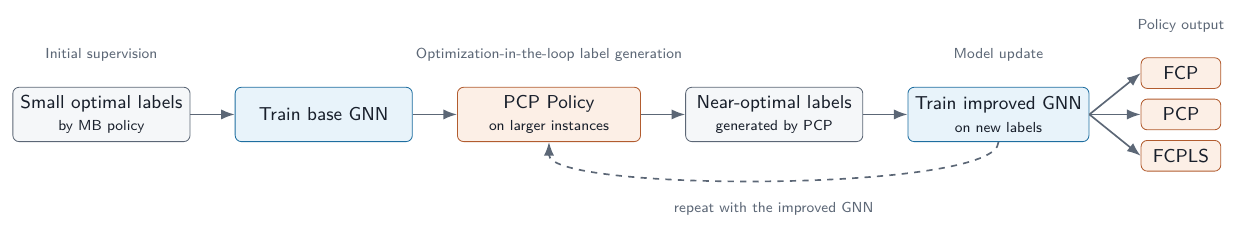}
    \caption{Iterative self-improvement}
    \label{fig:self_improve}
\end{figure}

\section{Theoretical Foundation}
\label{sec:theory_main}

In this section, we provide a theoretical justification for the edge-level
prediction target used by our GNN-guided pruning framework. Recall that the
implemented network in Section~\ref{subsec:GNN_structure} outputs a
segment-product probability matrix
\(\mathbf{P} \in(0,1)^{m\times n}\), where \(\mathbf{P}_{kj}\) is interpreted as
the probability that product \(j\) is included in the bundle selected by
segment \(k\). The inference policies in
Section~\ref{subsec:inference_policy} use this matrix to construct restricted
candidate bundle families before solving the mixed bundling MILP.

Our GNN does not attempt to represent the full mixed bundling solution,
which contains exponentially many bundle-level prices and choice variables.
Instead, we focus on the product-assignment projection of an optimal
mixed bundling solution. Let \(W^\star=(\boldsymbol{\Theta}^\star,\mathbf p^\star,
\mathbf P^{\mathrm{pay},\star},\mathbf s^\star,\mathbf Z^\star)\) be an optimal
solution of the full formulation \(\Xi(\mathfrak F)\), where
\(\mathfrak F=2^{[n]}\). Let \(\mathbf M\in\{0,1\}^{(2^n)\times n}\) denote the
bundle-product incidence matrix with
\(
(\mathbf M)_{bj}:=\mathbbm 1_{\{j\in \mathcal{A}(b)\}}.
\)
The product-assignment projection is
\[
\mathbf Q^{\star}=\boldsymbol{\Theta}^{\star} \mathbf M\in\{0,1\}^{m\times n},
\qquad
q^{\star}_{kj}
=
\sum_{b:j\in \mathcal{A}(b)}\theta^{\star}_{kb}.
\]
Thus, \(q^{\star}_{kj}=1\) if and only if product \(j\) is contained in the bundle
selected by segment \(k\) in the optimal mixed bundling solution. This
projection is naturally an edge-level object on the compact segment-product
graph since each coordinate corresponds to one segment-product pair.

In this section, for ease of theoretical analysis, we generalize the network in Section~\ref{subsec:GNN_structure} and consider a class \(\mathcal F_{\mathrm{edge}}^{W,\sigma}\) of finite-layer GNNs.
We first consider a class \(\mathcal F_{\mathrm{edge}}^{W}\) of finite-layer GNNs without the last sigmoid layer.

Let \(X=(G,H,E)\in\mathcal X_{m,n}\) denote the segment-product graph input,
where the formal input space \(\mathcal X_{m,n}\) and the permutation action
are defined in Appendix~\ref{app:preliminaries}. Let
\(\mathbf y_k^S\), \(\mathbf y_j^P\), and \(u_{kj}\) denote the raw customer,
product, and segment-product edge features defined in
Section~\ref{subsec:GraphRepresentation}. We
choose learnable continuous maps
\(
\iota_S,\iota_P,\iota_E,
\{\phi_S^{(\ell)},\phi_P^{(\ell)},\Gamma_S^{(\ell)},
\Gamma_P^{(\ell)},\Gamma_E^{(\ell)}\}_{\ell=1}^{\omega},
R_E, R_1,
\)
which are shared across customers, products, and edges of the same type.

The logit-output GNN class first
initializes
\[
\mathbf v_k^{S,(0)}=\iota_S(\mathbf y_k^S),
\qquad
\mathbf v_j^{P,(0)}=\iota_P(\mathbf y_j^P),
\qquad
\mathbf e_{jk}^{(0)}
=
\iota_E(\mathbf y_j^P,\mathbf y_k^S,u_{kj}).
\]
At each layer \(\ell=1,\ldots,\omega\), the customer and product embeddings
are updated by shared message-passing maps:
\[
\mathbf v_k^{S,(\ell)}
=
\Gamma_S^{(\ell)}
\left(
\mathbf v_k^{S,(\ell-1)},
\sum_{j=1}^n u_{kj}\,
\phi_P^{(\ell)}(\mathbf v_j^{P,(\ell-1)})
\right),
\qquad
\mathbf v_j^{P,(\ell)}
=
\Gamma_P^{(\ell)}
\left(
\mathbf v_j^{P,(\ell-1)},
\sum_{k=1}^m u_{kj}\,
\phi_S^{(\ell)}(\mathbf v_k^{S,(\ell-1)})
\right).
\]

The edge embedding is then updated using the current endpoint embeddings and
the previous edge embedding:
\[
\mathbf z_{jk}^{(\ell)}
:=
\mathbf v_j^{P,(\ell)}
\Vert
\mathbf v_k^{S,(\ell)}
\Vert
\mathbf e_{jk}^{(\ell-1)},
\qquad
\mathbf e_{jk}^{(\ell)}
=
\Gamma_E^{(\ell)}\bigl(\mathbf z_{jk}^{(\ell)}\bigr),
\]
where \(\Vert\) denotes concatenation. Let
\(\mathcal U^{(\omega)}(X)\) collect all terminal customer, product, and edge
embeddings. The logit output is defined by
\(
F(X)_{kj}
=
R_E\bigl(\mathbf e_{jk}^{(\omega)},\mathcal U^{(\omega)}(X)\bigr),
k\in[m],\ j\in[n],
\)
where the edge readout \(R_E\) is invariant in $\mathcal U^{(\omega)}(X)$. Each \(F\in\mathcal F_{\mathrm{edge}}^{W}\) maps an instance \(X\) to a
real-valued matrix
\(
F(X)\in\mathbb R^{m\times n}.
\)
The
corresponding scalar-output class is denoted by \(\mathcal F_{\mathrm{edge}}^1\),
where a scalar output has the form
\(
f(X)=R_1(\mathcal U^{(\omega)}(X))
\)
with a permutation-invariant scalar readout \(R_1\).

Then, we define the corresponding class \(\mathcal F_{\mathrm{edge}}^{W,\sigma}\) of finite layer GNNs with a sigmoid layer as 
\[
\mathcal F_{\mathrm{edge}}^{W,\sigma}
=
\{\, \sigma \circ \Phi : \Phi \in \mathcal{F}_{\mathrm{edge}}^{W} \,\},
\]
where the sigmoid function \(\sigma(\cdot)\) is applied entry-wise.

\begin{remark}[Scalar-to-edge lift]
\label{rem:scalar_edge_lift}
The scalar-valued and edge-valued classes are compatible in the following
sense:
\(
\{f\,\mathbf 1_{m\times n}: f\in\mathcal F_{\mathrm{edge}}^{1}\}
\subset
\mathcal F_{\mathrm{edge}}^{W}.
\)
Indeed, if
\(
f(X)=R_1(\mathcal U^{(\omega)}(X)),
\)
then choose the edge readout
\(
R_E(\mathbf e, \mathcal U):=R_1(\mathcal U).
\)
This gives \(F(X)_{kj}=f(X)\) for every
\((k,j)\in[m]\times[n]\).
\end{remark}

The main representation result below will be stated for a canonical
single-valued optimal assignment mapping \(\Phi_Q\), which is
constructed in Appendix~\ref{subsec:opt_assign}. Before
stating that result, we first introduce the WL refinement used to characterize
the expressive power of \(\mathcal F_{\mathrm{edge}}^{W}\).

\subsection{Segment-Product WL Refinement and Unfoldability}
\label{subsec:wl_unfoldability_main}
In this subsection, following the idea of \cite{chen2023milp}, given a finite set of instances, we establish that GNN can approximate the optimal product assignment to full mixed bundling problem $\Xi(\mathfrak F)$ under some conditions.

We now describe the color-refinement procedure used to analyze the expressive
power of \(\mathcal F_{\mathrm{edge}}^{W}\). The Weisfeiler--Lehman (WL) test \citep{weisfeiler1968reduction} iteratively assigns colors to
nodes based on their own colors and the multiset of neighbor colors. Different from the  node-output WL test in \cite{chen2023milp}, our network
maintains edge embeddings and outputs an edge-level matrix. We therefore use a
segment-product WL refinement that augments node-side colors with edge-side
colors. This modification allows us to track which segment-product edge
coordinates are distinguishable by the edge-output architecture.

\begin{algorithm}[h]
\caption{Segment-Product WL Test with Edge Colors, denoted by \(\mathrm{WL_{SP}}\)}
\label{alg:wl_cp}
\begin{algorithmic}
\Statex \textbf{Input:} A segment-product graph instance \(X=(G,H,E)\in\mathcal X_{m,n}\) and the number of WL refinement rounds \(T\).
\Statex \textbf{Initialize:}
\(
C^{0,S}_k=\mathrm{HASH}_{0,S}(\mathbf y^S_k),
C^{0,P}_j=\mathrm{HASH}_{0,P}(\mathbf y^P_j),
C^{0,E}_{kj}
=
\mathrm{HASH}_{0,E}\bigl(C^{0,S}_k,C^{0,P}_j,u_{kj}\bigr).
\)

\For{\(\ell=1,2,\ldots,T\)}

\(
C_k^{\ell,S}
=
\mathrm{HASH}_{\ell,S}\left(C_k^{\ell-1,S},\sum_{j\in[n]}u_{kj}\,\mathrm{HASH}'_{\ell,P}\bigl(C_j^{\ell-1,P}\bigr)\right),
\quad \forall k\in[m].
\)

\(
C_j^{\ell,P}
=
\mathrm{HASH}_{\ell,P}\left(C_j^{\ell-1,P},\sum_{k\in[m]}u_{kj}\,\mathrm{HASH}'_{\ell,S}\bigl(C_k^{\ell-1,S}\bigr)\right),
\quad \forall j\in[n].
\)

\(
C_{kj}^{\ell,E}
=
\mathrm{HASH}_{\ell,E}
\bigl(C_{kj}^{\ell-1,E},C_k^{\ell,S},C_j^{\ell,P}\bigr),
\qquad
\forall k\in[m],\ j\in[n].
\)
\EndFor
\Statex \textbf{Return:}
\(
\left(
\{C_k^{T,S}\}_{k=1}^{m},\{C_j^{T,P}\}_{j=1}^{n},\{C_{kj}^{T,E}\}_{k=1,j=1}^{m,n}
\right).
\)
\end{algorithmic}
\end{algorithm}

\begin{assumption}[WL hash family]
\label{ass:wl_hash}
In Algorithm~\ref{alg:wl_cp}, a hash map is an abstract color encoder used by the WL refinement: it maps each tuple of previous colors and graph features to a discrete color. We assume that all hash maps are injective. The maps
\(\mathrm{HASH}'_{\ell,S}\) and \(\mathrm{HASH}'_{\ell,P}\) use linearly
independent color codes, so the weighted sums in the WL updates are
collision-free.
\end{assumption}

We next introduce the notion of foldability. Intuitively, a foldable instance
contains two customer nodes or two product nodes that remain indistinguishable
under all rounds of WL refinement. Such nodes cannot be assigned different
representations by a message-passing GNN that respects the graph symmetry.

\begin{definition}[Foldable and unfoldable instances]
\label{def:foldable_main}
An instance \(X\in\mathcal X_{m,n}\) is called foldable under
\(\mathrm{WL}_{\mathrm{SP}}\) if there exist two distinct customer nodes
\(k\neq k'\) or two distinct product nodes \(j\neq j'\) that receive identical
node-side colors at every refinement level. That is, \(X\) is foldable if
either
\(
C_k^{\ell,S}(X)=C_{k'}^{\ell,S}(X), \forall \ell\ge0,
\)
for some \(k\neq k'\), or
\(
C_j^{\ell,P}(X)=C_{j'}^{\ell,P}(X), \forall \ell\ge0,
\)
for some \(j\neq j'\). Otherwise, \(X\) is called unfoldable. The set of
foldable instances is denoted by \(\mathcal D_{\mathrm{foldable}}\).
\end{definition}

For an unfoldable instance, the node-side WL colors become discrete after at most \(m+n\)
rounds of refinement, where all segment-side colors are pairwise distinct and all product-side
colors are pairwise distinct. Since the
edge hash in Algorithm~\ref{alg:wl_cp} is injective in the current endpoint
colors, unfoldability also implies that distinct edge coordinates are
distinguishable after finitely many rounds.

Finally, we define the cross-instance WL equivalence relation used in the
separation theorem. This relation compares two segment-product graphs up to a
simultaneous relabeling of customer and product indices.

\begin{definition}[Cross-instance WL equivalence]
\label{def:cross_instance_wl_equiv}
For two instances \(X,\widehat X\in\mathcal X_{m,n}\), we write
\(X\sim\widehat X\) if, for every refinement level \(L\ge0\), there exists
\((\pi_S^{(L)},\pi_P^{(L)})\in S_m\times S_n\) such that
\[
C_k^{L,S}(X)=C_{\pi_S^{(L)}(k)}^{L,S}(\widehat X),\quad
C_j^{L,P}(X)=C_{\pi_P^{(L)}(j)}^{L,P}(\widehat X),\quad
C_{kj}^{L,E}(X)
=
C_{\pi_S^{(L)}(k)\,\pi_P^{(L)}(j)}^{L,E}(\widehat X),
\quad
\forall k\in[m],\ j\in[n].
\]
\end{definition}

The next subsection uses this relation to connect the separation power (ability to distinguish non-isomorphic graphs) of
\(\mathcal F_{\mathrm{edge}}^1\) and \(\mathcal F_{\mathrm{edge}}^W\) to
\(\mathrm{WL}_{\mathrm{SP}}\), and then states the representation theorem for
the canonical optimal assignment mapping.

\subsection{Representation result}

We now define the target mapping represented by \(\mathcal F_{\mathrm{edge}}^{W,\sigma}\). Throughout this subsection, the target is defined with respect to the full
mixed bundling formulation \(\Xi(\mathfrak F)\), where
\(\mathfrak F=2^{[n]}\) is the full bundle family. For an instance
\(X\in\mathcal X_{m,n}\), let
\(
\Phi_{\mathrm{feas}}(X)
:=
\mathbbm 1\{\Xi(\mathfrak F;X)\text{ is feasible}\},
\mathcal D_{\mathrm{solu}}
:=
\mathcal X_{m,n}\cap \Phi_{\mathrm{feas}}^{-1}(1).
\)
Here \(\Xi(\mathfrak F;X)\) denotes the full mixed bundling formulation induced
by instance \(X\). By the boundedness argument in
Lemma~\ref{lem:bounded_formulation}, every instance in
\(\mathcal D_{\mathrm{solu}}\) admits an optimal solution.

For a feasible solution \(W=(\boldsymbol{\Theta},\mathbf p,
\mathbf P^{\mathrm{pay}},\mathbf s,\mathbf Z)\) of \(\Xi(\mathfrak F;X)\), define
its product-assignment projection by
\(
\mathbf Q(W):=\boldsymbol{\Theta}(W)\mathbf M,
\)
where \(\mathbf M\in\{0,1\}^{|\mathcal K|\times n}\) is the bundle-product incidence
matrix with entries
\(
(\mathbf M)_{bj}=\mathbbm 1_{\{j\in \mathcal{A}(b)\}},
b\in\mathcal K,\ j\in[n].
\)

Since \(\Xi(\mathfrak F;X)\) may have multiple optimal solutions, the optimal
product-assignment projection may not be unique. We therefore define a
single-valued canonical optimal assignment mapping
\(
\Phi_Q:
\mathcal D_{\mathrm{solu}}\setminus\mathcal D_{\mathrm{foldable}}
\to
\{0,1\}^{m\times n}.
\)
Formally, Appendix~\ref{subsec:opt_assign} constructs
\(\Phi_Q\) by first applying a WL-based canonical sorting map
\(\Phi_{\mathrm{sort}}\) on non-foldable instances and then selecting the
lexicographically smallest matrix among all optimal product-assignment
projections. This construction makes \(\Phi_Q\) well defined and permutation
equivariant.

We state the separation result connecting this edge-output GNN class to
the WL refinement on the segment-product graph.

\begin{theorem}
\label{thm:edge_thm42}
For any \(X,\widehat X\in\mathcal X_{m,n}\), the following are equivalent:
\begin{enumerate}
    \item[(i)] \(X\sim \widehat X\);

    \item[(ii)] \(f(X)=f(\widehat X)\) for all
    \(f\in\mathcal F_{\mathrm{edge}}^{1}\);

    \item[(iii)] for every \(F\in\mathcal F_{\mathrm{edge}}^{W}\), there
    exists \((\pi_S,\pi_P)\in S_m\times S_n\) such that
    \(
    F(\widehat X)=\pi_S F(X)\pi_P^\top.
    \)
\end{enumerate}
\end{theorem}

Theorem~\ref{thm:edge_thm42} shows that \(\mathcal F_{\mathrm{edge}}^{W}\) has the
same separation power as the WL refinement.

\begin{theorem}
\label{thm:sigmoid_assignment_recovery}
Fix \(m,n\), and let
\(
D\subset
\mathcal D_{\mathrm{solu}}\setminus\mathcal D_{\mathrm{foldable}}
\)
be a finite dataset of solvable unfoldable mixed bundling instances with
\(m\) customer segments and \(n\) products. For any
\(\varepsilon\in(0,\tfrac12)\), there exists
\(
F_\varepsilon^\sigma\in\mathcal F_{\mathrm{edge}}^{W,\sigma}
\)
such that
\[
\left|
F_\varepsilon^\sigma(X)_{kj}
-
\Phi_Q(X)_{kj}
\right|
<
\varepsilon \qquad
\forall X\in D, k\in[m], j\in[n].
\]
Moreover, thresholding at \(1/2\) exactly recovers the canonical binary
product-assignment matrix:
\[
\mathbbm 1_{\{F_\varepsilon^\sigma(X)_{kj}>1/2\}}
=
\Phi_Q(X)_{kj},
\qquad
\forall X\in D, k\in[m], j\in[n].
\]
\end{theorem}

A direct implication of Theorem~\ref{thm:sigmoid_assignment_recovery} is that
the empirical binary cross-entropy loss \(\mathcal L_D(F_\varepsilon^\sigma)\) can be made arbitrarily small on the
finite dataset \(D\):
\(
\mathcal L_D(F_\varepsilon^\sigma)
\le
\max\{w_{\mathrm{pos}},1\}[-\log(1-\varepsilon)]
\to0
\quad\text{as }\varepsilon\downarrow0.
\)

\section{Numerical Experiments}
\label{sec:experiment}
We now present comprehensive numerical experiments designed to evaluate the effectiveness and scalability of our proposed GNN-guided bundle pricing strategies. The goals of this section are twofold: (i) to demonstrate that our approaches consistently achieve high profit ratios while maintaining significant computational efficiency, and (ii) to validate their robustness across heterogeneous problem settings with varying numbers of customer segments and products. 

To thoroughly test scalability and robustness, we evaluate our strategies under varying numbers of customer segment $m$ and product $n$. For each scenario, results are averaged over 100 instances. The GNN model training was conducted with a single NVIDIA A100 GPU, while the GNN model inference and the inference policies were performed on a local machine featuring an Apple M1 Pro CPU (3.2 GHz) and 16GB RAM, utilizing up to 8 threads. The implementation uses Gurobi 12.0, Python 3.10, PyTorch 2.8, and CUDA 12.8. 

To evaluate any method, we adopt two normalized metrics: \emph{profit ratio (PR)} and \emph{time ratio (TR)}
\[
  PR_{A,B} = \tfrac{\text{Profit of policy A}}{\text{Profit of policy B}},\quad
TR_{A,B} = \tfrac{\text{Runtime of policy A}}{\text{Runtime of policy B}},
\]
where PR measures solution quality and TR captures efficiency.
These two metrics allow direct comparison of solution quality and speed 
across different algorithms.

\subsection{Main Experiments}
\label{subsec:basic_simu}
We compare our proposed approach against the two most fundamental and widely recognized policies in the literature: the exact Mixed Bundling (MB) policy and the approximation-based Bundle Size Pricing (BSP) policy:

\begin{enumerate}
    \item \textbf{Mixed Bundling (MB) policy.} For MB, we follow the MILP formulation of \citet{hanson1990optimal}. The MB formulation is a MILP that assigns a separate price to each candidate bundle. Detailed formulation is provided in Section~\ref{app:hanson}. We solve the MB policy with a MIPGap threshold of 0.1\%.
    \item \textbf{Bundle Size Pricing (BSP) policy.} The BSP baseline is proposed by \citet{chu2011bundle}. The BSP approximates MB by assigning the same price to all bundles of equal size. Detailed formulation is provided in Appendix~\ref{app:bsp}. We solve the BSP policy with an MIPGap threshold of 0.1\%.
\end{enumerate}

It is worth noting that we exclude certain heuristic algorithms and general neural-network-based MILP methods from our baselines due to their inapplicability to the general bundle pricing problem; a detailed discussion is provided in  Appendix~\ref{subsec:limitation_policies}.

The instance-generation procedure and GNN training details are
provided in Appendix~\ref{app:main_deterministic_training}. In the main
experiments, we train the base GNN on 3000 small instances with
\((m_{\mathrm{train}},n_{\mathrm{train}})=(10,10)\), where labels are obtained
by solving the exact MB formulation and projecting the optimal assignments to
segment-product labels. The dataset is split into training and validation
sets, and the best checkpoint is selected by validation loss. To mitigate random initialization effects, we train 10 GNN
models with seeds 1 to 10 (see Appendix~\ref{subsec:seed_performance} for details) and report the average performance of the resulting
inference policies. All MILP-based policies are solved with a MIPGap threshold
of \(0.1\%\).

In Table~\ref{tab:mb-n10-comparison}, we report the numerical results of our
algorithms compared with both baselines for problems with \(n=10\) and varying
numbers of customer segments. All three GNN-based policies recover more than
\(98\%\) of the optimal MB profit while requiring only a small fraction of the
MB runtime. In contrast, BSP is computationally efficient but incurs a
substantial profit loss, achieving only about \(85\%\)--\(88\%\) of the MB
benchmark. Among the proposed policies, FCP provides the fastest 
policy, PCP achieves the highest profit by retaining a richer candidate family,
and FCPLS offers an intermediate trade-off by improving upon FCP through local
search while remaining faster than PCP.

\begin{table}[h]
\centering
\caption{Performance of GNN-based inference policies compared with the MB policy (\(n=10\)).}
\label{tab:mb-n10-comparison}
\resizebox{\textwidth}{!}{
\begin{tabular}{l rrr rrr rrr rrr}
\toprule
& \multicolumn{3}{c}{\textbf{FCP}}
& \multicolumn{3}{c}{\textbf{PCP}}
& \multicolumn{3}{c}{\textbf{FCPLS}}
& \multicolumn{3}{c}{\textbf{BSP}} \\
\cmidrule(lr){2-4} \cmidrule(lr){5-7} \cmidrule(lr){8-10} \cmidrule(lr){11-13}
\textbf{Scenario}
& \(PR_{\cdot,MB}\) (std) & Time (s) & \(TR_{\cdot,MB}\)
& \(PR_{\cdot,MB}\) (std) & Time (s) & \(TR_{\cdot,MB}\)
& \(PR_{\cdot,MB}\) (std) & Time (s) & \(TR_{\cdot,MB}\)
& \(PR_{\cdot,MB}\) (std) & Time (s) & \(TR_{\cdot,MB}\) \\
\midrule
\(m_{\mathrm{test}}=10,n_{\mathrm{test}}=10\)
& 0.989 (0.007) & 0.021 & 0.004
& 0.995 (0.005) & 0.169 & 0.029
& 0.996 (0.004) & 0.134 & 0.024
& 0.879 (0.028) & 0.168 & 0.031 \\
\addlinespace

\(m_{\mathrm{test}}=20,n_{\mathrm{test}}=10\)
& 0.985 (0.007) & 0.177 & 0.007
& 0.995 (0.004) & 2.010 & 0.061
& 0.991 (0.006) & 1.280 & 0.052
& 0.864 (0.025) & 0.391 & 0.015 \\
\addlinespace

\(m_{\mathrm{test}}=30,n_{\mathrm{test}}=10\)
& 0.982 (0.006) & 1.051 & 0.007
& 0.994 (0.004) & 17.640 & 0.079
& 0.988 (0.006) & 6.798 & 0.055
& 0.854 (0.020) & 1.210 & 0.008 \\
\bottomrule
\end{tabular}}
\vspace{0.1cm}
\begin{flushleft}
\footnotesize
\textit{Note:} \(PR_{\cdot,MB}\) denotes the profit ratio relative to the
full mixed bundling policy MB, and \(TR_{\cdot,MB}\) denotes the time ratio
relative to MB. The reported \(TR_{\cdot,MB}\) is computed as the average of
instance-wise runtime ratios.
\end{flushleft}
\end{table}

For problems with more than 10 products, solving the full MB formulation becomes
computationally challenging. Therefore, we use BSP as the baseline for evaluating
larger instances. The results are reported in Table~\ref{tab:bsp-comparison}.

\begin{table}[h]
\centering
\caption{Comparison of different inference policies across various problem sizes compared with the BSP policy.}
\label{tab:bsp-comparison}
\resizebox{\textwidth}{!}{
\begin{tabular}{l rrr rrr rrr} 
\toprule
& \multicolumn{3}{c}{\textbf{FCP}} 
& \multicolumn{3}{c}{\textbf{PCP}} 
& \multicolumn{3}{c}{\textbf{FCPLS}} \\
\cmidrule(lr){2-4} \cmidrule(lr){5-7} \cmidrule(lr){8-10}
\textbf{Scenario} 
& \(PR_{\cdot,BSP}\) (std) & Time (s) & \(TR_{\cdot,BSP}\)
& \(PR_{\cdot,BSP}\) (std) & Time (s) & \(TR_{\cdot,BSP}\)
& \(PR_{\cdot,BSP}\) (std) & Time (s) & \(TR_{\cdot,BSP}\) \\
\midrule
\(m_{\mathrm{test}}=10,n_{\mathrm{test}}=20\) 
& 1.136 (0.040) & 0.021 & 0.147
& 1.162 (0.042) & 0.672 & 4.604
& 1.150 (0.041) & 0.240 & 1.696 \\
\addlinespace

\(m_{\mathrm{test}}=10,n_{\mathrm{test}}=30\) 
& 1.117 (0.039) & 0.025 & 0.083
& 1.182 (0.037) & 2.060 & 7.261
& 1.141 (0.038) & 0.333 & 1.112 \\
\addlinespace

\(m_{\mathrm{test}}=10,n_{\mathrm{test}}=40\) 
& 1.076 (0.047) & 0.028 & 0.056
& 1.196 (0.037) & 6.039 & 12.091
& 1.113 (0.045) & 0.508 & 1.010 \\
\addlinespace

\(m_{\mathrm{test}}=20,n_{\mathrm{test}}=20\) 
& 1.161 (0.032) & 0.175 & 0.165
& 1.191 (0.035) & 5.414 & 4.817
& 1.175 (0.034) & 5.717 & 5.419 \\
\addlinespace

\(m_{\mathrm{test}}=20,n_{\mathrm{test}}=30\) 
& 1.137 (0.037) & 0.207 & 0.076
& 1.207 (0.037) & 41.048 & 15.422
& 1.161 (0.037) & 16.667 & 6.367 \\
\addlinespace

\(m_{\mathrm{test}}=20,n_{\mathrm{test}}=40\) 
& 1.096 (0.045) & 0.268 & 0.064
& \multicolumn{1}{c}{--} & \multicolumn{1}{c}{--} & \multicolumn{1}{c}{--}
& 1.130 (0.040) & 36.170 & 8.661 \\
\bottomrule
\end{tabular}}
\vspace{0.1cm}
\begin{flushleft}
\footnotesize
\textit{Note:} \(PR_{\cdot,BSP}\) denotes the profit ratio relative to BSP,
and \(TR_{\cdot,BSP}\) denotes the time ratio relative to BSP. The symbol ``--''
indicates that PCP was not evaluated for that scenario due to excessive
computation time.
\end{flushleft}
\end{table}

Overall, FCP achieves strong performance with high computational efficiency,
consistently outperforming BSP while using only a fraction of its runtime.
PCP further improves the profit ratio by retaining a richer candidate
bundle family, but its runtime grows substantially; hence it is not evaluated
for \((m_{\mathrm{test}}=20,n_{\mathrm{test}}=40)\). FCPLS provides a middle ground by improving
upon FCP through GNN-guided local search while remaining more scalable than
PCP on larger instances.

\begin{figure}[h]
    \centering
    \includegraphics[width=0.9\linewidth]{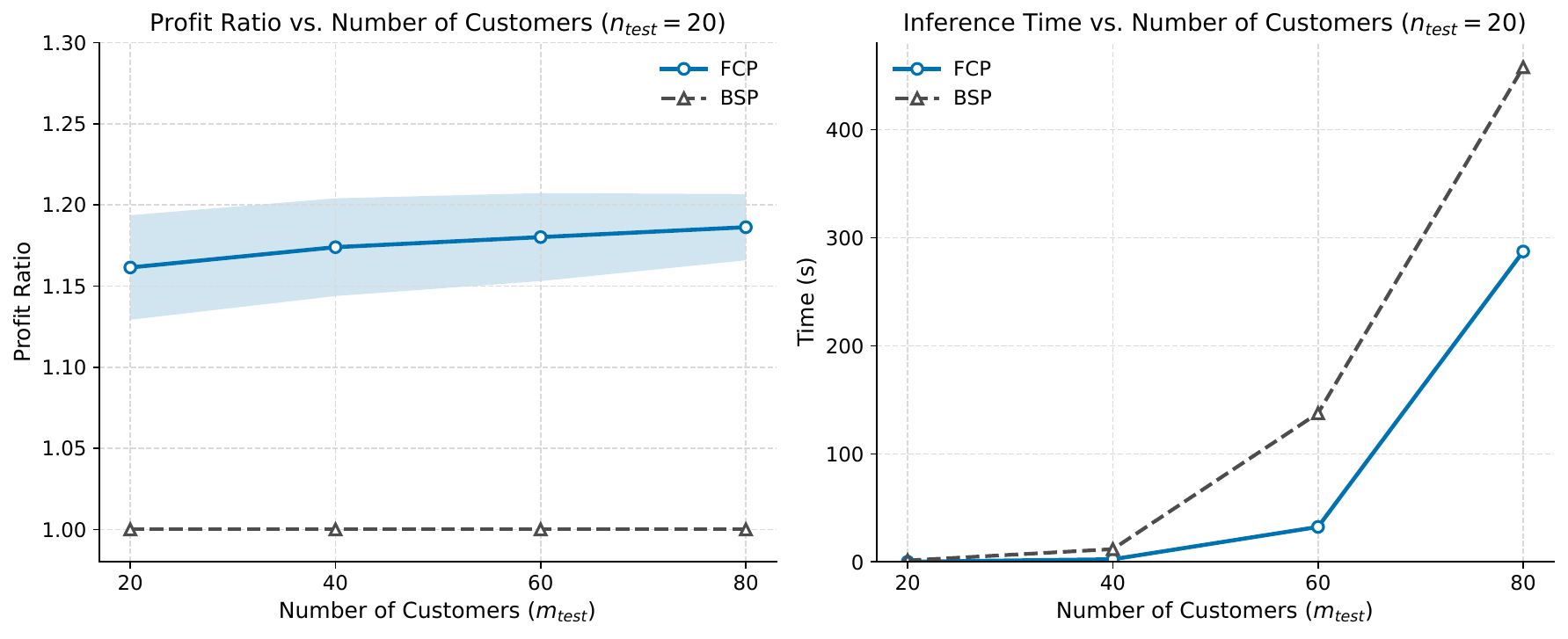}
    \caption{Model scalability across $m_{\mathrm{test}}$ under FCP policy.}
    \label{fig:scalability_m}
\end{figure}

\paragraph{Scalability across the number of customer segments \(m_{\mathrm{test}}\).}
Under a 600-second time limit, Figure~\ref{fig:scalability_m} illustrates the
scalability of the FCP policy as the number of customer segments \(m_{\mathrm{test}}\)
varies while fixing \(n_{\mathrm{test}}=20\). The FCP policy consistently outperforms
BSP in terms of profit across all tested values of \(m_{\mathrm{test}}\), achieving
profit ratios above 1.16. In terms of runtime, FCP remains substantially
faster than BSP for all sizes of \(m_{\mathrm{test}}\). Overall, these results
show that FCP scales effectively with the number of customer segments while
maintaining strong solution quality.

\begin{figure}[h]
    \centering
    \includegraphics[width=0.9\linewidth]{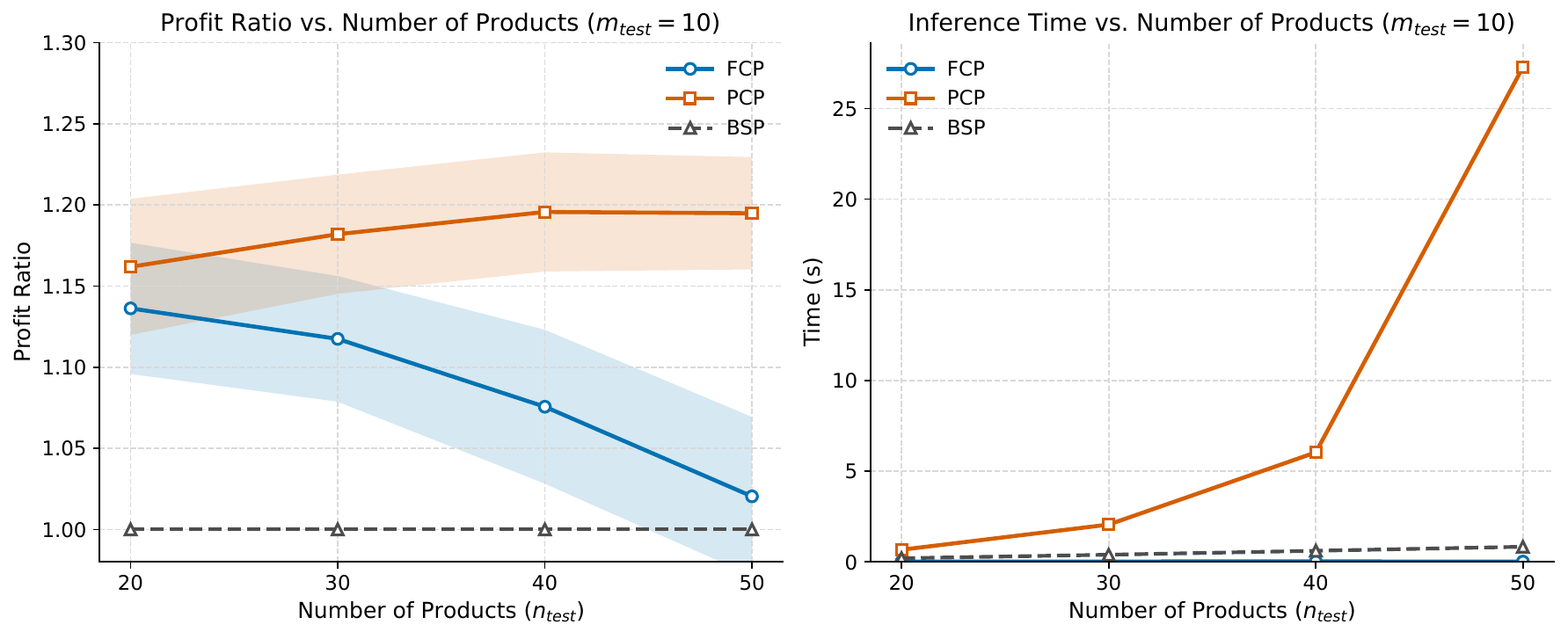}
    \caption{Comparison of model scalability across $n_{\mathrm{test}}$ under different inference policies.}
    \label{fig:scalability_n}
\end{figure}

\paragraph{Scalability across the number of products \(n_{\mathrm{test}}\).}
Under a 600-second time limit, Figure~\ref{fig:scalability_n} illustrates the
scalability of FCP and PCP as the number of products \(n_{\mathrm{test}}\) varies
while fixing \(m_{\mathrm{test}}=10\). PCP demonstrates strong solution quality,
consistently achieving profit ratios above 1.16 relative to BSP and
improving as \(n_{\mathrm{test}}\) increases. This benefit, however, comes with higher
computational cost because PCP retains a larger candidate bundle family.
By contrast, FCP remains extremely fast across all tested product sizes, but
its solution quality decreases as \(n_{\mathrm{test}}\) grows. In particular, when
\(n_{\mathrm{test}}=50\), FCP becomes comparable to BSP in terms of profit.

These results reveal a useful but incomplete form of size generalization. The
GNN trained on \(n_{\mathrm{train}}=10\) captures transferable
segment-product patterns, but a large increase in the product dimension
changes the combinatorial structure of the optimal bundle family. Rather than
treating this as a limitation of the GNN framework, we use it as an opportunity
for self-improvement: the small-scale model can still generate high-quality
large-scale solutions through PCP, and these solutions can in turn serve as
near-optimal labels for retraining at the larger scale.

\subsection{Iterative Self-Improvement}
\label{subsec:exp_iter}
To evaluate the efficacy of the iterative self-improvement strategy proposed in Section~\ref{subsec:iter_self_improve}, we conduct a specific experiment to assess cross-scale generalization.

In the numerical experiments, we first train a GNN model using instances of size $m_{\mathrm{train}}=10, n_{\mathrm{train}}=10$ following the setting in Section~\ref{subsec:basic_simu}. Then, we use the PCP policy to generate near-optimal labels for 3000 instances of size $m_{\mathrm{train}}=10, n_{\mathrm{train}}=50$, and we train a separate GNN model using these 3000 samples with $m_{\mathrm{train}}=10$ and $n_{\mathrm{train}}=50$. The corresponding inference policies using the new model are denoted by FCP-I, PCP-I, and FCPLS-I.

\begin{figure}[h]
    \centering
    \includegraphics[width=0.9\linewidth]{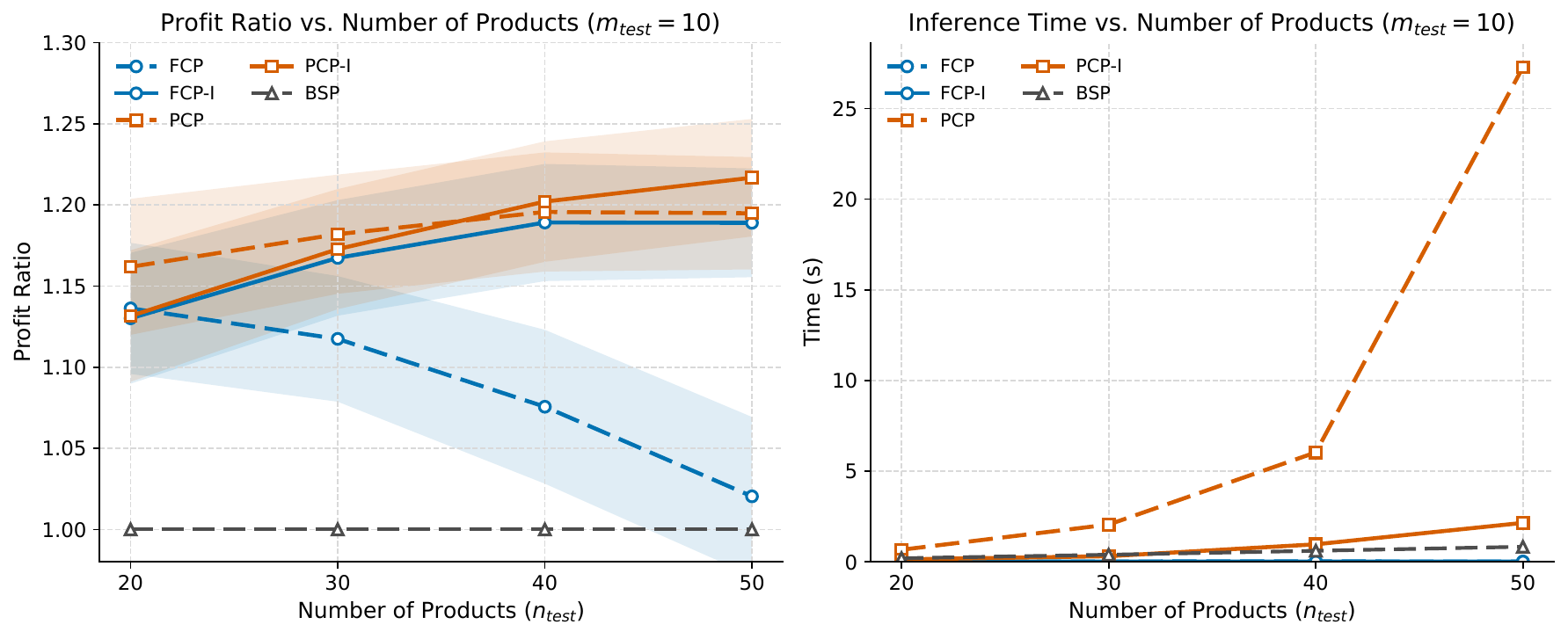}
    \caption{Comparison of model scalability under different training configurations}
    \label{fig:iterative_improvement}
\end{figure}

Figure~\ref{fig:iterative_improvement} illustrates the effect of the iterative
self-improvement strategy. The self-improved model trained on
\(n_{\mathrm{train}}=50\) improves size generalization, especially for problems with larger
product size. For FCP, FCP-I is slightly worse than the base model at
\(n_{\mathrm{test}}=20\), but outperforms it for all \(n_{\mathrm{test}}\ge30\),
with larger gains as \(n_{\mathrm{test}}\) increases. For PCP, PCP-I is worse than the base PCP model at
\(n_{\mathrm{test}}=20\), but becomes comparable at \(n_{\mathrm{test}}=30\)
and outperforms the base model for all \(n_{\mathrm{test}}\ge 40\), with the
largest gain observed at \(n_{\mathrm{test}}=50\).

Runtime remains comparable between FCP and FCP-I, while PCP-I is generally
faster than the base PCP model despite achieving stronger solution quality.
These results show that iterative self-improvement improves transfer to larger
product spaces while preserving the computational efficiency of the pruning
framework.

Table~\ref{tab:bsp-comparison-large} reports the performance of the
self-improved policies on larger instances up to \(n_{\mathrm{test}}=100\).
FCP-I is the most scalable policy: it consistently outperforms BSP in profit,
with profit ratios between about \(1.13\) and \(1.21\), while requiring only a
small fraction of BSP's runtime. FCPLS-I provides a moderate but consistent
quality improvement over FCP-I, at the cost of additional local-search time.
PCP-I achieves the highest profit ratios in the reported
\(m_{\mathrm{test}}=10\) settings, but its runtime grows quickly with
\(n_{\mathrm{test}}\), so it is not reported for the larger
\(m_{\mathrm{test}}=20\) settings.

\begin{table}[h]
\centering
\caption{Comparison of different inference policies across various problem sizes trained on PCP-generated labels of size \(m_{\mathrm{train}}=10,n_{\mathrm{train}}=50\).}
\label{tab:bsp-comparison-large}
\resizebox{\textwidth}{!}{
\begin{tabular}{l rrr rrr rrr} 
\toprule
& \multicolumn{3}{c}{\textbf{FCP-I}} 
& \multicolumn{3}{c}{\textbf{PCP-I}} 
& \multicolumn{3}{c}{\textbf{FCPLS-I}} \\
\cmidrule(lr){2-4} \cmidrule(lr){5-7} \cmidrule(lr){8-10}
\textbf{Scenario} 
& \(PR_{\cdot,BSP}\) (std) & Time (s) & \(TR_{\cdot,BSP}\)
& \(PR_{\cdot,BSP}\) (std) & Time (s) & \(TR_{\cdot,BSP}\)
& \(PR_{\cdot,BSP}\) (std) & Time (s) & \(TR_{\cdot,BSP}\) \\
\midrule
\(m_{\mathrm{test}}=10,n_{\mathrm{test}}=60\) 
& 1.187 (0.033) & 0.021 & 0.026
& 1.241 (0.034) & 3.472 & 4.053
& 1.197 (0.032) & 0.170 & 0.203 \\
\addlinespace

\(m_{\mathrm{test}}=10,n_{\mathrm{test}}=80\) 
& 1.168 (0.045) & 0.028 & 0.019
& 1.297 (0.047) & 19.466 & 13.468
& 1.184 (0.045) & 0.186 & 0.127 \\
\addlinespace

\(m_{\mathrm{test}}=10,n_{\mathrm{test}}=100\) 
& 1.131 (0.049) & 0.031 & 0.016
& 1.347 (0.050) & 100.913 & 47.902
& 1.151 (0.047) & 0.217 & 0.111 \\
\addlinespace

\(m_{\mathrm{test}}=20,n_{\mathrm{test}}=60\) 
& 1.213 (0.032) & 0.184 & 0.023
& \multicolumn{1}{c}{--} & \multicolumn{1}{c}{--} & \multicolumn{1}{c}{--}
& 1.218 (0.032) & 10.127 & 1.230 \\
\addlinespace

\(m_{\mathrm{test}}=20,n_{\mathrm{test}}=80\) 
& 1.186 (0.039) & 0.216 & 0.021
& \multicolumn{1}{c}{--} & \multicolumn{1}{c}{--} & \multicolumn{1}{c}{--}
& 1.194 (0.038) & 16.872 & 1.640 \\
\addlinespace

\(m_{\mathrm{test}}=20,n_{\mathrm{test}}=100\) 
& 1.143 (0.042) & 0.242 & 0.018
& \multicolumn{1}{c}{--} & \multicolumn{1}{c}{--} & \multicolumn{1}{c}{--}
& 1.154 (0.040) & 25.733 & 1.881 \\
\bottomrule
\end{tabular}}
\vspace{0.1cm}
\begin{flushleft}
\footnotesize
\textit{Note:} The symbol ``--''
indicates that PCP-I was not evaluated for that scenario due to excessive
computation time.
\end{flushleft}
\end{table}

\section{Extension: Additive Random Valuations}
\label{sec:external_validation_random_valuation}

The main experiments focus on the deterministic segment-level mixed bundling
setting studied in Sections~\ref{sec:problem}--\ref{sec:experiment}. In that
setting, each instance contains a fixed set of customer segments with known
bundle valuations, and the objective of \(\Xi(\mathfrak F)\) is the exact
profit for that deterministic population. In this section, we conduct an
external validation experiment under
additive random valuations. Under additive valuations, the value of
bundle \(b\) for customer \(i\) is
\(
R_{ib}
=
\sum_{j\in \mathcal{A}(b)} v_{ij}.
\)

In a random-valuation problem, customer valuations are drawn from an underlying
distribution. The seller's objective is therefore an expected profit, rather
than the profit from a fixed deterministic set of segments. A common approach
for solving such stochastic pricing problems is sample average approximation
(SAA): draw a finite sample of customers, solve the deterministic optimization
problem induced by this sample, and then evaluate the resulting policy on
fresh out-of-sample valuation draws. We adopt this SAA approach here. Thus,
the sampled customers play a role analogous to the deterministic customer segments in
the main formulation, but the evaluation criterion is different: a good policy
must perform well not only on the optimization sample but also out of sample.

This experiment serves two purposes. First, it tests whether the proposed
GNN-guided pruning framework remains effective in a random-valuation setting
when combined with an SAA formulation. Second, it evaluates whether the restricted
mixed bundling formulation solved by FCP can produce prices that generalize to
unseen random valuations. For a sampled instance, FCP constructs a candidate
family \(\mathfrak B^{\mathrm{FCP}}\) from the GNN predictions and
solves the restricted formulation \(\Xi(\mathfrak B^{\mathrm{FCP}})\). We then
compare its in-sample and out-of-sample performance with two structured
benchmarks: bundle-size pricing (BSP) and CPBSD-A, an additive approximation
of component pricing with bundle-size discounts based on \cite{chen2025component}
and described in Appendix~\ref{app:cpbsd_additive_details}.

In this section, we use \(N\) to denote the number of products,
\(K_{\mathrm{in}}\) to denote the number of in-sample customers used in the
SAA optimization problem, and \(K_{\mathrm{oos}}\) to denote the number of
fresh out-of-sample valuation draws used for evaluation. We consider two
product-scale blocks. The first uses \(N=10\) and \(K_{\mathrm{in}}=50\); the
second uses \(N=30\) and \(K_{\mathrm{in}}=50\). In both cases, the resulting
price menu is evaluated on \(K_{\mathrm{oos}}\) out-of-sample draws. For each
scale, we test three product cost settings: zero costs, independent random
costs, and costs correlated with valuation means. For FCP, we reuse the same
GNN checkpoint trained once on smaller additive random-valuation instances,
without any scale-specific or cost-specific retraining. We report in-sample
profit (InS), out-of-sample profit (OOS), and runtime. Details on valuation
generation, cost definitions, the CPBSD-A formulation, implementation settings,
and the FCP evaluation procedure under SAA are provided in
Appendix~\ref{app:cpbsd_additive_details}; the procedure is summarized in
Algorithm~\ref{alg:saa_fcp_random_valuation}.

\begin{table}[h]
\centering
\caption{Additive random-valuation problems under
\(N=10,K_{\mathrm{in}}=50\) and \(N=30,K_{\mathrm{in}}=50\), evaluated on
\(K_{\mathrm{oos}}=5000\) out-of-sample draws.}
\label{tab:fcp_cpbsd_additive_transfer}
\resizebox{\linewidth}{!}{
\begin{tabular}{llrrrrrrrrr}
\toprule
 & & \multicolumn{3}{c}{FCP} & \multicolumn{3}{c}{BSP} & \multicolumn{3}{c}{CPBSD-A} \\
\cmidrule(lr){3-5}\cmidrule(lr){6-8}\cmidrule(lr){9-11}
Scale & Cost & InS & OOS & Time (s) & InS & OOS & Time (s) & InS & OOS & Time (s) \\
\midrule
\(N=10,K_{\mathrm{in}}=50\) & \texttt{zero} & 51.940 & \textbf{50.400} & \textbf{0.084} & 51.996 & 50.081 & 0.541 & \textbf{52.009} & 50.009 & 1.828 \\
 & \texttt{random\_ind} & \textbf{19.845} & \textbf{19.040} & \textbf{1.054} & 18.111 & 16.937 & 1.528 & 15.895 & 15.287 & 5.742 \\
 & \texttt{random\_corr} & \textbf{27.698} & \textbf{25.991} & \textbf{1.191} & 27.390 & 25.533 & 1.365 & 22.020 & 21.367 & 12.899 \\
\midrule
\(N=30,K_{\mathrm{in}}=50\) & \texttt{zero} & \textbf{159.064} & 156.261 & \textbf{0.114} & 159.008 & \textbf{156.282} & 1.367 & 159.051 & 154.968 & 300.147 \\
 & \texttt{random\_ind} & \textbf{56.142} & \textbf{54.627} & \textbf{2.860} & 46.391 & 44.439 & 35.186 & 24.154 & 22.829 & 300.141 \\
 & \texttt{random\_corr} & \textbf{83.123} & \textbf{79.116} & \textbf{4.686} & 81.183 & 77.388 & 5.149 & 60.684 & 58.735 & 300.162 \\
\bottomrule
\end{tabular}
}

\vspace{0.1cm}
\begin{flushleft}
\footnotesize
\textit{Note:} InS and OOS denote in-sample and out-of-sample profits, respectively. Bold values indicate the best value within each row and metric: higher is better for InS and OOS, and lower is better for runtime.
\end{flushleft}
\end{table}

Table~\ref{tab:fcp_cpbsd_additive_transfer} shows that FCP has the best overall performance across these additive random-valuation problems. It achieves the lowest runtime in every cost setup at both product scales and obtains the highest OOS profit in five out of six settings. The only exception is the zero-cost case at \(N=30,K_\mathrm{in}=50\), where BSP is marginally higher in OOS profit, while the difference is negligible relative to the runtime gap. Under the two random-cost settings, FCP attains the highest InS and OOS profit in all four rows, indicating that its advantage appears both in the SAA objective and in OOS generalization. In the simpler zero-cost cases, all three methods have close profit values: CPBSD-A achieves the highest InS profit at \(N=10,K_{\mathrm{in}}=50\), whereas FCP achieves the highest InS profit at \(N=30,K_{\mathrm{in}}=50\). The scalability contrast is especially clear at \(N=30,K_{\mathrm{in}}=50\): CPBSD-A reaches the 300-second time limit in all three cost setups, whereas FCP remains within a few seconds. These results support the main message that learned pruning works well under additive random-valuation problems while preserving both profitability and computational efficiency.

\section{Conclusion}
This paper develops a GNN-guided pruning framework for large-scale mixed
bundle pricing. Instead of constructing and optimizing over the full
exponential bundle space, we learn segment-product inclusion probabilities on a compact
segment-product graph. These predictions are used to prune the search space and construct a much smaller
candidate bundle family, over which we solve the
mixed bundling formulation. 

Algorithmically, we propose two pruning policies, FCP and PCP, that trade off candidate-family size and solution quality. We further introduce a GNN-guided local-search procedure that improves solution quality from a given initial solution. To enhance scalability, we propose an iterative self-improvement scheme that uses GNN-generated high-quality large-scale solutions as near-optimal labels to improve subsequent GNN-based policies. Numerical experiments show that the proposed framework achieves near-optimal performance on tractable instances, scales to much larger product spaces, and outperforms structured heuristics. Additional validation under additive random valuations shows that the same learning-and-pruning idea can be embedded in an SAA pipeline and can produce prices that generalize well out of sample while maintaining a clear
computational advantage.

Theoretically, we show that the optimal product-assignment projection is the right edge-level object for learning since it avoids the
exponential bundle-level MILP graph while preserving the assignment structure
needed for pruning. We further establish, via a segment-product WL refinement
with edge colors, that a specific edge-output GNN can represent this
canonical assignment target on finite solvable and non-foldable datasets and
recover the binary assignment by thresholding.

Our framework clarifies \textit{when} and \textit{why} learning adds value in bundle pricing: the GNN does not replace the pricing optimization but concentrates it, directing scarce exact mixed-bundling computation to the few product combinations each segment is likely to select, so the firm preserves the self-selection and pricing rigor of the exact model while breaking its scalability barrier. A second insight concerns the lifecycle of the AI component: because exact labels are available only at small scales, the iterative self-improvement procedure lets the deployed model generate its own near-optimal supervision on progressively larger instances, keeping the model maintainable as the catalog grows without ever requiring exact large-scale labels. Operationally, the pruning policies let a firm match computational effort to its constraints---FCP suits fast, frequent repricing, whereas PCP and the GNN-guided local search trade additional runtime for higher profit.

% Acknowledgments here
% \ACKNOWLEDGMENT{%
% % Enter the text of acknowledgments here
% }% Leave this (end of acknowledgment)

\newpage 
\bibliographystyle{informs2014}
\bibliography{reference}
  % References here (outcomment the appropriate case)

\newpage
\begin{appendices}
\begin{center}
    \textbf{\Large Appendix}
\end{center}

\renewcommand{\theHsection}{A\arabic{section}}

\section{Formulations for Bundle Pricing}
\label{app:milp}

\subsection{Bundle Size Pricing (BSP) Formulation \citet{chu2011bundle}}
\label{app:bsp}

\noindent \textbf{Additional Variables:}
\begin{itemize}
    \item $p_s$: price for bundles of size $s \in [n]_+$;
    % \item $s_b$: cardinality of bundle $b$;
    \item $P_{ks} = p_{s}\theta_{ks}$.
\end{itemize}

\begin{align}
\max \quad
& \sum_{k=1}^m \sum_{s=0}^n \alpha_k Z_{ks}
\label{eq:bsp_obj}
\\
\text{s.t.}\quad
& \sum_{s=0}^n \theta_{ks}=1,
&& \forall k\in[m]
\label{eq:bsp_assign}
\\
& s_k \ge R_{ks}-p_s,
&& \forall k\in[m],\ s\in [n]_+
\label{eq:bsp_surplus_lb}
\\
& P_{ks}\le p_s,
&& \forall k\in[m],\ s\in [n]_+
\label{eq:bsp_price_ub}
\\
& P_{ks}\ge p_s-M(1-\theta_{ks}),
&& \forall k\in[m],\ s\in [n]_+
\label{eq:bsp_price_lb}
\\
& s_k=\sum_{s=0}^n (R_{ks}\theta_{ks}-P_{ks}),
&& \forall k\in[m]
\label{eq:bsp_surplus_def}
\\
& R_{ks}\theta_{ks}-P_{ks}\ge0,
&& \forall k\in[m],\ s\in [n]_+
\label{eq:bsp_ir}
\\
& s_k\ge \sum_{s=0}^n (R_{ks}\theta_{\ell s}-P_{\ell s}),
&& \forall k\in[m],\ \ell\in[m],\ \ell\neq k
\label{eq:bsp_ic}
\\
& Z_{ks}=P_{ks}-c_{ks}\theta_{ks},
&& \forall k\in[m],\ s\in [n]_+
\label{eq:bsp_profit_def}
\\
& p_{s_1+s_2}\le p_{s_1}+p_{s_2},
&& \forall s_1,s_2\ge0\ \text{s.t. } s_1, s_2 \in [n]_+, s_1+s_2\le n
\label{eq:bsp_subadd}
\\
& p_{s+1}\ge p_s,
&& \forall s\in [n-1]_+
\label{eq:bsp_monotonicity}
\\
& p_s,P_{ks},s_k,Z_{ks}\ge0,\quad \theta_{ks}\in\{0,1\},
&& \forall k\in[m],\ s\in [n]_+.
\label{eq:bsp_domain}
\end{align}

\subsection{LP Formulation (Fixed Assignment)}
\label{app:lp-ic}

In FCPLS, the local-search step evaluates a fixed segment-wise bundle assignment
by solving an LP. Let
\(
\mathbf b=(b_1,\dots,b_m)
\)
denote a fixed assignment, where \(b_k\in\mathcal K\) is the bundle index
assigned to segment \(k\). The corresponding active candidate bundle family is
\(
\mathfrak B^{\mathrm{LP}}(\mathbf b)
=
\{\mathcal A(b_k):k\in[m]\}\cup\{\emptyset\},
\)
with associated index set
\(
\mathcal I_{\mathfrak B^{\mathrm{LP}}}
=
\{b\in\mathcal K:\mathcal{A}(b)\in\mathfrak B^{\mathrm{LP}}(\mathbf b)\}.
\)
Thus \(b_k\in\mathcal I_{\mathfrak B^{\mathrm{LP}}}\) for every segment \(k\),
and \(0\in\mathcal I_{\mathfrak B^{\mathrm{LP}}}\).

\noindent \textbf{Decision variables:}
\begin{itemize}
    \item \(p_b\ge0\): price of bundle index
    \(b\in\mathcal I_{\mathfrak B^{\mathrm{LP}}}\);
    \item \(s_k\ge0\): surplus of segment \(k\in[m]\).
\end{itemize}

\noindent \textbf{Objective:}
\begin{equation}
\max_{p,s} \quad
\sum_{k=1}^m \alpha_k\bigl(p_{b_k}-c_{k b_k}\bigr),
\tag{L.1}
\end{equation}
where \(b_k\) is the fixed bundle index assigned to segment \(k\).

\noindent \textbf{Constraints:}
\begin{align}
& s_k \ge R_{kb}-p_b,
&& \forall k\in[m],\ \forall b\in\mathcal I_{\mathfrak B^{\mathrm{LP}}}
\qquad \text{(surplus lower bound)}
\tag{L.2}\\
& s_k \le R_{k b_k}-p_{b_k},
&& \forall k\in[m]
\qquad \text{(assignment binding)}
\tag{L.3}\\
& p_b \le \sum_{c\in\mathcal C}p_c,
&& \forall b\in\mathcal I_{\mathfrak B^{\mathrm{LP}}},\ 
   \forall \mathcal C\subseteq
   \mathcal I_{\mathfrak B^{\mathrm{LP}}}
\notag\\
&&& \text{with }
\mathcal{A}(b)\subseteq \bigcup_{c\in\mathcal C}\mathcal A(c)
\qquad \text{(cover subadditivity)}
\tag{L.4}\\
& p_0=0
&& \qquad \text{(outside option price)}
\tag{L.5}
\end{align}

\noindent \textit{Remarks.}
\begin{itemize}
    \item The bundle assignments are fixed externally: segment \(k\) is assigned
    to bundle index \(b_k\), and no binary assignment variables are used in this
    LP.

    \item Constraints (L.2) and (L.3) together enforce incentive compatibility
    within the active candidate family:
    \(
    s_k
    =
    R_{k b_k}-p_{b_k}
    =
    \max_{b\in\mathcal I_{\mathfrak B^{\mathrm{LP}}}}
    \{R_{kb}-p_b\}.
    \)

    \item Constraint (L.4) is the cover-based price-subadditivity condition used
    in the FCPLS implementation. It differs from the two-way partition used in the standard MB.

    \item In implementation, it is sufficient to add only minimal cover
    inequalities, since non-minimal covers are redundant.

    \item The LP is solved only over the active index set
    \(\mathcal I_{\mathfrak B^{\mathrm{LP}}}\), which contains the bundles indices
    currently assigned to at least one segment and the outside option.

    \item The optimal value of this fixed-assignment LP is a lower bound for the
    restricted mixed bundling formulation
    \(\Xi(\mathfrak B^{\mathrm{LP}}(\mathbf b))\), because the MILP optimizes
    over both prices and assignments while this LP keeps the assignment fixed.
\end{itemize}

\subsection{Modified Price Subadditivity Constraints}
\label{app:subadd}

Our mixed bundling formulation follows the formulation of \citet{hanson1990optimal}
in terms of decision variables, objective function, and most constraints,
including consumer surplus, single-price schedule, self-selection, and
individual rationality constraints. The only difference lies in how we write
the price-subadditivity constraints for a general candidate bundle family
\(\mathfrak B\subseteq\mathfrak F\).

In the original full mixed bundling formulation of
\citet{hanson1990optimal}, all product subsets are included, i.e.,
\(\mathfrak B=\mathfrak F=2^{[n]}\), or equivalently
\(\mathcal I_{\mathfrak B}=\mathcal K\) at the bundle-index level. In that
case, price subadditivity can be written as
\[
p_b
\le
\sum_{c\in\mathcal C}p_c,
\qquad
\forall b\in\mathcal K,\quad
\forall \mathcal C\subseteq\mathcal K
\text{ with }
\mathcal{A}(b)=\bigcup_{c\in\mathcal C}\mathcal A(c).
\]
Since the full formulation contains every product subset, this condition can
be simplified to two-way partition subadditivity:
\[
p_b\le p_{b_1}+p_{b_2},
\qquad
\forall b,b_1,b_2\in\mathcal K
\text{ with }
\mathcal A(b_1)\cap \mathcal A(b_2)=\emptyset
\text{ and }
\mathcal A(b_1)\cup \mathcal A(b_2)=\mathcal{A}(b).
\]
The equivalence is proved in
Appendix~\ref{proof:subadd}.

For a restricted candidate family \(\mathfrak B\subset\mathfrak F\), however,
the two-way partition simplification is generally not valid, because
\(\mathfrak B\) is not necessarily closed under partitions. Therefore, in the
restricted formulation \(\Xi(\mathfrak B)\), we impose price subadditivity in
the cover form:
\[
p_b
\le
\sum_{c\in\mathcal C}p_c,
\qquad
\forall b\in\mathcal I_{\mathfrak B},\quad
\forall \mathcal C\subseteq\mathcal I_{\mathfrak B}
\text{ with }
\mathcal{A}(b)\subseteq\bigcup_{c\in\mathcal C}\mathcal A(c).
\]
This form applies both to the full bundle universe and to restricted candidate
families generated by our pruning policies. In implementation, it is sufficient
to add only minimal cover inequalities, since non-minimal covers are redundant
under nonnegative prices.

\begin{definition}[K-way cover subadditivity \(S_{C,K}\)]
For any bundle index \(b\) and any collection
\(\mathcal C=\{c_1,\dots,c_K\}\) of \(K\ge2\) bundle indices such that
\(
\mathcal{A}(b)\subseteq \bigcup_{r=1}^K \mathcal A(c_r),
\)
the following holds:
\(
p_b \le \sum_{r=1}^{K} p_{c_r}.
\)
\end{definition}

\begin{definition}[K-way partition subadditivity \(S_{P,K}\)]
For any bundle index \(b\) and any \(K\ge2\) bundle indices
\(b_1,\dots,b_K\) such that \(\mathcal A(b_1),\dots,\mathcal A(b_K)\) form a pairwise disjoint
partition of \(\mathcal{A}(b)\), the following holds:
\(
p_b \le \sum_{r=1}^{K}p_{b_r}.
\)
\end{definition}

\begin{definition}[Two-way partition subadditivity \(S_{P,2}\)]
For any bundle indices \(b,b_1,b_2\in\mathcal K\) such that
\(\mathcal A(b_1)\cap \mathcal A(b_2)=\emptyset\) and
\(\mathcal A(b_1)\cup \mathcal A(b_2)=\mathcal{A}(b)\), the following holds:
\(
p_b\le p_{b_1}+p_{b_2}.
\)
\end{definition}

\subsubsection{Non-additivity and model size}
\label{app:nonadditive_scaling}
It is crucial to emphasize that our work addresses the challenging non-additive valuation setting, where consumer valuations and costs for bundles exhibit sub-additivity or super-additivity. While the mixed bundling MILP formulation appears syntactically identical regardless of additivity, the underlying computational complexity diverges fundamentally. As established by \citet{hanson1990optimal}, strict additivity allows for a powerful problem decomposition: because a bundle's valuation and cost are simply the sum of its individual components, the model can implicitly evaluate unlisted bundles as ``composite bundles'' dynamically assembled from individual items. This structural property collapses the exponential search space, allowing the problem to be formulated using only $O(mn)$ item-level decision variables. In contrast, the non-additive setting breaks this independence, enforcing a tightly coupled combinatorial structure where a bundle's valuation cannot be derived by merely summing its constituents. Consequently, instead of evaluating just $n$ items per segment, one must explicitly account for the valuations of all $m \cdot 2^n$ possible bundles in the MILP. This combinatorial explosion expands both the input size and the search space from linear to exponential, rendering the exact optimization problem computationally intractable.

\subsubsection{Restricted bundle families versus full bundle families}
\label{app:restricted_menu_two_stage}
Bundle pricing can be viewed as a two-stage problem: first choosing which
bundles to offer, and then optimizing their prices. The full mixed bundling
formulation \(\Xi(\mathfrak F)\) combines these two stages by retaining all
possible bundles and optimizing prices subject to self-selection and
subadditivity constraints. However, solving \(\Xi(\mathfrak F)\) and then
keeping only the bundles purchased in its optimal solution is generally not
equivalent to solving a restricted pricing problem over that selected menu.

The reason is that removing unoffered bundles also removes the corresponding
self-selection and cover-subadditivity constraints. Hence, if
\(\mathfrak S^\star\subseteq\mathfrak F\) denotes the family of product sets
purchased in an optimal solution of \(\Xi(\mathfrak F)\), together with the
empty bundle, then \(\Xi(\mathfrak S^\star)\ge \Xi(\mathfrak F)\), and the
inequality can be strict.

For example, identify the three-product set with \(\{A,B,C\}\), assume zero
costs, and consider three customer segments with weights
\(\alpha_1=1\), \(\alpha_2=1\), and \(\alpha_3=2\). The valuations are additive
with utility vectors \(\mathbf u_1=(5,3,5)\), \(\mathbf u_2=(0,9,1)\), and
\(\mathbf u_3=(9,10,6)\). For readability in this example, for any product set
\(S\subseteq\{A,B,C\}\), let \(b(S):=\mathcal A^{-1}(S)\) be its bundle index and write
\(p_S:=p_{b(S)}\).

In the full formulation \(\Xi(\mathfrak F)\), one optimal solution sells
\(\{C\}\) to segment 1, \(\{B\}\) to segment 2, and \(\{A,B,C\}\) to segment
3, with prices \(p_{\{C\}}=5\), \(p_{\{B\}}=9\), and
\(p_{\{A,B,C\}}=23\), yielding
\(\Xi(\mathfrak F)=1\cdot5+1\cdot9+2\cdot23=60\). If we restrict the menu to the
selected family
\(\mathfrak S^\star=\{\emptyset,\{B\},\{C\},\{A,B,C\}\}\), then the restricted
model can charge \(p_{\{C\}}=5\), \(p_{\{B\}}=9\), and
\(p_{\{A,B,C\}}=24\), with the same assignments, yielding
\(\Xi(\mathfrak S^\star)=1\cdot5+1\cdot9+2\cdot24=62>60\).

This gap arises because the full menu contains additional bundles, such as
\(\{B,C\}\), whose self-selection and subadditivity constraints limit the price
of \(\{A,B,C\}\). Once these bundles are not offered in the restricted menu,
the corresponding constraints disappear, allowing a higher optimal price and a
larger restricted-menu objective.

\section{Supplementary Algorithm Details}
\label{sec:pseudocode}

\begin{breakablealgorithm}
\caption{Fixed Cutoff Pruning (FCP) Policy}
\label{alg:fcp}
\begin{algorithmic}
\State \textbf{Input:} Predicted probability matrix $\mathbf{P} \in [0,1]^{m \times n}$, parameters $(\mathbf{U}, \mathbf c^{\mathit u},\mathbf c^{\mathit s}, \boldsymbol{\alpha})$, cutoff $\tau$(default $0.5$)
\State \textbf{Output:} Optimal prices $\mathbf{p}^{\star}$, assignment $\boldsymbol{\Theta}^{\star}$, and candidate bundle family $\mathfrak B^{\mathrm{FCP}}$
\State Initialize candidate bundle family $\mathfrak B^{\mathrm{FCP}} \gets \{\emptyset\}$
\For{$k = 1$ to $m$}
    \State $S_k \gets \{ j \in [n] \mid \mathbf{P}_{kj} \ge \tau \}$ \Comment{Identify high-probability products}
    \If{$S_k = \emptyset$}
        \Comment{Avoid empty predicted bundle by selecting max probability product}
        \State $j^{\star} \gets \argmax_{j \in [n]} \mathbf{P}_{kj}$
        \State $B_k^{\mathrm{FCP}} \gets \{j^{\star}\}$
    \Else
        \State $B_k^{\mathrm{FCP}} \gets S_k$
    \EndIf
    \State $\mathfrak B^{\mathrm{FCP}} \gets \mathfrak B^{\mathrm{FCP}} \cup \{ B_k^{\mathrm{FCP}} \}$
\EndFor
\State Solve $\Xi(\mathfrak B^{\mathrm{FCP}})$ to obtain optimal solution
\State \Return $\mathbf{p}^{\star}, \boldsymbol{\Theta}^{\star}, \mathfrak B^{\mathrm{FCP}}$
\end{algorithmic}
\end{breakablealgorithm}

\begin{breakablealgorithm}
\caption{Progressive Cutoff Pruning (PCP) Policy}
\label{alg:pcp_cp}
\begin{algorithmic}
\State \textbf{Input:} Predicted probability matrix $\mathbf{P} \in [0,1]^{m \times n}$, parameters $(\mathbf{U}, \mathbf c^{\mathit u},\mathbf c^{\mathit s}, \boldsymbol{\alpha})$, cutoff $\tau$ (default $0.5$)
\State \textbf{Output:} Optimal prices $\mathbf{p}^{\star}$, assignment $\boldsymbol{\Theta}^{\star}$, and candidate bundle family $\mathfrak B^{\mathrm{PCP}}$
\State Initialize candidate bundle family $\mathfrak B^{\mathrm{PCP}} \gets \{\emptyset\}$
\For{$k = 1$ to $m$}
    \State $U_k \gets \{ j \in [n] \mid \mathbf{P}_{kj} \ge \tau \}$ \Comment{Filter products}
    \State Sort $U_k$ by descending $\mathbf{P}_{kj}$ to obtain sequence $(j_{(1)}, j_{(2)}, \dots, j_{(|U_k|)})$
    \State Initialize prefix chain $B_{k,0}^{\text{PCP}} \gets \emptyset$
    \For{$i = 1$ to $|U_k|$}
        \State $B_{k,i}^{\mathrm{PCP}} \gets B_{k,i-1}^{\text{PCP}} \cup \{j_{(i)}\}$ \Comment{Progressive construction}
        \State $\mathfrak B^{\mathrm{PCP}} \gets \mathfrak B^{\mathrm{PCP}} \cup \{ B_{k,i}^{\mathrm{PCP}} \}$
    \EndFor
\EndFor
\State Solve $\Xi(\mathfrak B^{\mathrm{PCP}})$ with cutting-plane subadditivity separation to obtain optimal solution
\State \Return $\mathbf{p}^{\star}, \boldsymbol{\Theta}^{\star}, \mathfrak B^{\mathrm{PCP}}$
\end{algorithmic}
\end{breakablealgorithm}

\begin{breakablealgorithm}
\caption{Adaptive Global Local Search (Global Top-K)}
\label{alg:global_ls}
\begin{algorithmic}
\State \textbf{Input:} Initial product-assignment matrix $\mathbf Q^{(0)} \in \{0,1\}^{m \times n}$, probability matrix $\mathbf P$, MaxIter
\State \textbf{Output:} Improved product-assignment matrix $\widehat{\mathbf Q}$
\State \textbf{Initialize:} $\widehat{\mathbf Q} \gets \mathbf Q^{(0)}$, $rev^{\star} \gets \text{LP-Eval}(\widehat{\mathbf Q})$
\State \textbf{Parameter:} $K \gets \lceil \sqrt{m} \rceil$ \Comment{Adaptive sublinear neighborhood}

\State $iter \gets 0$
\While{$iter <$ MaxIter}
    \State $iter \gets iter + 1$, $improve \gets \text{FALSE}$
    
    \State \textbf{Step 1: Identify Candidates Globally}
    \State Let $\mathcal{U} = \{(k, j) : \widehat{\mathbf Q}_{kj} = 0\}$ be the set of unselected items
    \State Let $\mathcal{S} = \{(k, j) : \widehat{\mathbf Q}_{kj} = 1\}$ be the set of selected items
    
    \State $\mathcal{C}_{\text{add}} \gets \operatorname*{arg\,top}_{K} \{ \mathbf P_{kj} \mid (k,j) \in \mathcal{U} \}$ \Comment{Highest probability adds}
    \State $\mathcal{C}_{\text{drop}} \gets \operatorname*{arg\,top}_{K} \{ 1 - \mathbf P_{kj} \mid (k,j) \in \mathcal{S} \}$ \Comment{Lowest probability drops}
    
    \State \textbf{Step 2: Mix and Sort}
    \State $\mathcal{C} \gets \text{SortDescending}(\mathcal{C}_{\text{add}} \cup \mathcal{C}_{\text{drop}})$ based on scores
    
    \State \textbf{Step 3: Sequential Evaluation}
    \For{move $\mu \in \mathcal{C}$}
        \State $\mathbf Q' \gets \text{ApplyMove}(\widehat{\mathbf Q}, \mu)$
        \State $(feas, rev) \gets \text{LP-Eval}(\mathbf Q')$
        
        \If{$feas$ \textbf{and} $rev > rev^{\star} + \epsilon$}
            \State $\widehat{\mathbf Q} \gets \mathbf Q'$, $rev^{\star} \gets rev$
            \State $improve \gets \text{TRUE}$
            \State \textbf{break} \Comment{Greedy accept \& restart iteration}
        \EndIf
    \EndFor
    
    \If{\textbf{not} $improve$}
        \State \textbf{break} \Comment{Converged}
    \EndIf
\EndWhile
\State \Return $\widehat{\mathbf Q}$
\end{algorithmic}
\end{breakablealgorithm}

We denote by $\operatorname*{arg\,top}_{K}(\mathcal X)$  the operator that returns the set of K elements from $\mathcal X$ with the highest values. Formally, given a set of candidate moves and their associated scores, this operator selects the subset of size K that maximizes the scores, breaking ties arbitrarily.

\begin{breakablealgorithm}
\caption{Iterative self-improvement strategy}
\label{alg:iterative_self_improvement}
\begin{algorithmic}
% Redefine Require/Ensure to Input/Output
\renewcommand{\algorithmicrequire}{\textbf{Input:}}
\renewcommand{\algorithmicensure}{\textbf{Output:}}

\Require Small-scale dimensions $(m_{small}, n_{small})$; Large-scale dimensions $(m_{large}, n_{large})$; Mixed Bundling Policy $\pi_{\text{MB}}$; Inference Policy $\pi_{\text{PCP}}$.
\Ensure Trained GNN model $\zeta_{new}$ for large-scale instances.

\Statex \textbf{Phase 1: Base Model Training (Small Scale)}
\State Generate $N_{small}$ historical instances $\mathcal{I}_{small}$ with dimensions $(m_{small}, n_{small})$.
\State Initialize dataset $\mathcal{D}_{base} \leftarrow \emptyset$.
\For{each instance $\ell \in \mathcal{I}_{small}$}
    \State Obtain optimal bundle assignment $\mathbf{Q}^{\star}_{(\ell)} \leftarrow \pi_{\text{MB}}(\text{instance } \ell)$. \Comment{Exact labels via MB}
    \State $\mathcal{D}_{base} \leftarrow \mathcal{D}_{base} \cup \{(\text{instance } \ell, \mathbf{Q}^{\star}_{(\ell)})\}$.
\EndFor
\State Split $\mathcal{D}_{base}$ into training/validation sets ($80\%/20\%$).
\State Initialize GNN parameters $\zeta_{base}$.
\State $\zeta_{base} \leftarrow \text{TrainGNN}(\zeta_{base}, \mathcal{D}_{base})$. \Comment{Train base model on exact labels}

\Statex
\Statex \textbf{Phase 2: Self-improvement via Bootstrapping (Large Scale)}
\State Generate $N_{large}$ instances $\mathcal{I}_{large}$ with dimensions $(m_{large}, n_{large})$.
\State Initialize dataset $\mathcal{D}_{new} \leftarrow \emptyset$.
\For{each instance $\ell' \in \mathcal{I}_{large}$}
    \State Predict selection probabilities $\mathbf{P}_{(\ell')} \leftarrow \text{GNN}(\text{instance } \ell'; \zeta_{base})$.
    \State Generate solution $\hat{\mathbf{Q}}^{\star}_{(\ell')} \leftarrow \pi_{\text{PCP}}(\mathbf{P}_{(\ell')})$. \Comment{Near-optimal labels via PCP}
    \State $\mathcal{D}_{new} \leftarrow \mathcal{D}_{new} \cup \{(\text{instance } \ell', \hat{\mathbf{Q}}^{\star}_{(\ell')})\}$.
\EndFor
\State Split $\mathcal{D}_{new}$ into training/validation sets ($80\%/20\%$).
\State Initialize GNN parameters $\zeta_{new}$.
\State $\zeta_{new} \leftarrow \text{TrainGNN}(\zeta_{new}, \mathcal{D}_{new})$. \Comment{Train new model on near-optimal labels}

\State \Return $\zeta_{new}$
\end{algorithmic}
\end{breakablealgorithm}

\section{Supplementary Proofs}
\label{app:theory}

\subsection{Preliminaries}
\label{app:preliminaries}
We use the segment-product graph representation defined in
Section~\ref{subsec:GraphRepresentation}. For fixed numbers of customer
segments \(m\) and products \(n\), let \(\mathcal G_{m,n}\) denote the
collection of bipartite graphs with \(m\) customer nodes and \(n\) product
nodes. Let \(\mathcal H^S\) and \(\mathcal H^P\) be the segment-side and
product-side node-feature spaces, respectively, and let \(\mathscr E\) be the
edge-feature space. We write
\[
\mathcal H_m^S:=(\mathcal H^S)^m,\qquad
\mathcal H_n^P:=(\mathcal H^P)^n,\qquad
\mathscr E_{m,n}:=\mathscr E^{m\times n}.
\]
The full input space is
\(
\mathcal X_{m,n}
:=
\mathcal G_{m,n}\times \mathcal H_m^S\times \mathcal H_n^P\times
\mathscr E_{m,n}.
\)
An instance is denoted by
\(
X=(G,H,E)\in\mathcal X_{m,n},
\)
where \(G\in\mathcal G_{m,n}\) is the bipartite graph, \(H\in
\mathcal H_m^S\times\mathcal H_n^P\) stacks all customer and product node
features, and \(E=(u_{kj})_{k\in[m],j\in[n]}\in\mathscr E_{m,n}\) collects all
segment-product edge features. The primitive features are those defined in
Section~\ref{subsec:GraphRepresentation}; we do not repeat them here.

Let
\(
\Gamma_{m,n}:=S_m\times S_n
\)
be the product of the customer segment and product permutation groups. For
\(g=(\pi_S,\pi_P)\in\Gamma_{m,n}\), we write \(g\star X\) for the instance
obtained by relabeling customer segments by \(\pi_S\) and products by
\(\pi_P\). In coordinates, the relabeled edge features satisfy
\(
(g\star E)_{kj}
=
E_{\pi_S^{-1}(k),\,\pi_P^{-1}(j)}.
\)
The node features are relabeled analogously.

A scalar mapping \(f:\mathcal X_{m,n}\to\mathbb R\) is permutation invariant if
\[
f(g\star X)=f(X),
\qquad
\forall g\in\Gamma_{m,n},\ X\in\mathcal X_{m,n}.
\]
An edge-output mapping \(F:\mathcal X_{m,n}\to\mathbb R^{m\times n}\) is
permutation equivariant if
\[
F(g\star X)
=
\pi_S F(X)\pi_P^\top,
\qquad
\forall g=(\pi_S,\pi_P)\in\Gamma_{m,n},\ X\in\mathcal X_{m,n}.
\]
Equivalently, for every \(k\in[m]\) and \(j\in[n]\),
\(
F(g\star X)_{kj}
=
F(X)_{\pi_S^{-1}(k),\,\pi_P^{-1}(j)}.
\)

\subsection{The Weisfeiler-Lehman (WL) Test for mixed bundling and Problem Unfoldability}

\begin{lemma}[Terminal discreteness on unfoldable instances]
\label{lem:unfoldable_terminal_discreteness}
Let \(X\in\mathcal X_{m,n}\setminus\mathcal D_{\mathrm{foldable}}\).
Under the WL hashing of Assumption~\ref{ass:wl_hash}, the node-side
WL refinement becomes discrete after at most \(m+n\) rounds. That is, for
\(T_\star:=m+n\),
\(
C_k^{T_\star,S}(X)\neq C_{k'}^{T_\star,S}(X),
 \forall k\neq k',
\)
and
\(
C_j^{T_\star,P}(X)\neq C_{j'}^{T_\star,P}(X),
 \forall j\neq j'.
\)
Consequently, all edge coordinates are also distinguished:
\(
C_{kj}^{T_\star,E}(X)\neq C_{k'j'}^{T_\star,E}(X),
 \forall (k,j)\neq(k',j').
\)
\end{lemma}

\proof{Proof.}
Consider the color partitions induced by the WL colors on the disjoint
union of customer and product nodes. Since the color update at level \(l\)
includes the previous color at level \(l-1\), the sequence of color
partitions is monotone: once two nodes receive different colors, they can
never be merged again at a later iteration.

There are only \(m+n\) nodes in total. Hence the number of color classes can
strictly increase at most \(m+n-1\) times. Therefore the refinement process
stabilizes after at most \(m+n\) iterations.

Suppose, for contradiction, that two distinct customer nodes \(k\neq k'\)
satisfy
\(
C_k^{m+n,S}(X)=C_{k'}^{m+n,S}(X).
\)
Since the refinement has stabilized by level \(m+n\), these two customer
nodes will continue to receive the same color at all later levels. By
monotonicity, this implies
\(
C_k^{l,S}(X)=C_{k'}^{l,S}(X),
\quad \forall l\ge 0,
\)
which means that \(X\) is foldable. This contradicts
\(X\notin\mathcal D_{\mathrm{foldable}}\). Hence all customer-side colors
are pairwise distinct at level \(m+n\). The product-side argument is
identical.
\qed
\endproof

\subsection{The Sorting Mapping $\Phi_{sort}$}
\label{subsec:sorting_map}

Similar to the order-refinement construction for MILP solution
mappings in \citet{chen2023milp}, we fix a deterministic sorting map
\[
\Phi_{\mathrm{sort}}:
\mathcal X_{m,n}\setminus \mathcal D_{\mathrm{foldable}}
\to S_m\times S_n,
\qquad
\Phi_{\mathrm{sort}}(X)=(\sigma_S(X),\sigma_P(X)).
\]
The construction refines the WL descriptors by lexicographic comparisons
until the segment-side and product-side orders become strict. Since \(X\) is
unfoldable, the final WL colors are pairwise distinct on both sides, so
\(\Phi_{\mathrm{sort}}\) is well defined. Moreover, it is permutation equivariant:
for every \((\pi_S,\pi_P)\in S_m\times S_n\),
\[
\Phi_{\mathrm{sort}}\big((\pi_S,\pi_P)\star X\big)
=
(\pi_S\circ\sigma_S(X),\pi_P\circ\sigma_P(X)).
\]

\subsection{The Optimal Assignment Mapping $\Phi_Q$}
\label{subsec:opt_assign}

\begin{lemma}
\label{lem:bounded_formulation}
For every \(X\in\mathcal D_{\mathrm{solu}}\), the full mixed bundling
formulation \(\Xi(\mathfrak F;X)\) admits an equivalent bounded formulation.
In particular, the following bounds are implied by the existing constraints and may be added without loss:
\[
0\le p_b\le R_{\max},\qquad
0\le P_{kb}\le R_{\max},\qquad
0\le s_k\le R_{\max},\qquad
-C_{\max}\le Z_{kb}\le R_{\max},
\]
for all \(k\in[m]\) and \(b\in\mathcal K\), where
\(
R_{\max}:=\max_{k,b}R_{kb},
C_{\max}:=\max_{k,b}c_{kb}.
\)
Consequently, every feasible instance in \(\mathcal D_{\mathrm{solu}}\)
admits an optimal solution.
\end{lemma}

\proof{Proof.}
For any \(k,b\), the constraints \(P_{kb}\ge0\) and
\(R_{kb}\theta_{kb}-P_{kb}\ge0\) imply
\(
0\le P_{kb}\le R_{kb}\theta_{kb}\le R_{\max}.
\)
If \(\theta_{kb}=0\), then \(P_{kb}=0\), and the single-price constraint
\(p_b-R_{\max}(1-\theta_{kb})\le P_{kb}\) gives \(p_b\le R_{\max}\).
If \(\theta_{kb}=1\), the constraints \(p_b\le P_{kb}\le p_b\) imply
\(p_b=P_{kb}\le R_{\max}\). Hence \(0\le p_b\le R_{\max}\) for all \(b\).

Moreover, since each segment selects exactly one bundle and unselected bundles
have \(P_{kb}=0\),
\(
s_k=\sum_{b\in\mathcal K}(R_{kb}\theta_{kb}-P_{kb})\in[0,R_{\max}].
\)
Finally, \(Z_{kb}=P_{kb}-c_{kb}\theta_{kb}\), so
\(Z_{kb}\in[-C_{\max},R_{\max}]\). Thus all continuous variables are bounded,
and the binary variables take values in a finite set. The feasible region is
closed and compact, so the linear objective attains its maximum.
\qed
\endproof

For \(X\in\mathcal D_{\mathrm{solu}}\), let
\(\mathcal W_{\mathrm{feas}}(X)\) be the feasible region of
\(\Xi(\mathfrak F;X)\), and let \(z(W;X)\) be its objective value. By
Lemma~\ref{lem:bounded_formulation}, the optimal solution set
\(
\mathcal W_{\mathrm{opt}}(X)
:=
\operatorname*{arg\,max}_{W\in\mathcal W_{\mathrm{feas}}(X)} z(W;X)
\)
is nonempty. Define
\[
\mathcal Q_{\mathrm{opt}}(X)
:=
\{\mathbf Q(W): W\in\mathcal W_{\mathrm{opt}}(X)\}.
\]
Since \(\mathbf Q(W)\in\{0,1\}^{m\times n}\), the set
\(\mathcal Q_{\mathrm{opt}}(X)\) is finite and nonempty.

\begin{definition}[Canonical optimal assignment mapping]
\label{def:optimal_assignment_mapping}
For \(X\in\mathcal D_{\mathrm{solu}}\setminus\mathcal D_{\mathrm{foldable}}\),
let \(\Phi_{\mathrm{sort}}(X)=(\sigma_S,\sigma_P)\). For
\(\mathbf Q\in\mathbb R^{m\times n}\), let
\(
\operatorname{vec}_{\sigma_S,\sigma_P}(\mathbf Q)
:=
\bigl(Q_{\sigma_S(1),\sigma_P(1)},\ldots,
Q_{\sigma_S(1),\sigma_P(n)},\ldots,
Q_{\sigma_S(m),\sigma_P(n)}\bigr).
\)
Define \(\Phi_Q(X)\) as the matrix in \(\mathcal Q_{\mathrm{opt}}(X)\)
whose \(\operatorname{vec}_{\sigma_S,\sigma_P}\)-vector is lexicographically
smallest. Since \(\mathcal Q_{\mathrm{opt}}(X)\) is finite and nonempty, this
matrix exists and is unique.
\end{definition}

\begin{lemma}[Equivariance of \(\Phi_Q\)]
\label{lem:phiq_equiv}
For any
\(X\in\mathcal D_{\mathrm{solu}}\setminus\mathcal D_{\mathrm{foldable}}\)
and any \((\pi_S,\pi_P)\in S_m\times S_n\),
\(
\Phi_Q\bigl((\pi_S,\pi_P)\star X\bigr)
=
\pi_S\Phi_Q(X)\pi_P^\top .
\)
\end{lemma}

\proof{Proof.}
Since \(\mathfrak F=2^{[n]}\), every product permutation \(\pi_P\) induces a
bijection \(\tau_{\pi_P}\) on bundle indices, defined by
\(\mathcal A(\tau_{\pi_P}(b))=\pi_P(\mathcal{A}(b))\). Reindexing customers by \(\pi_S\) and
bundles by \(\tau_{\pi_P}\) gives an objective-preserving bijection between
the feasible regions of \(\Xi(\mathfrak F;X)\) and
\(\Xi(\mathfrak F;(\pi_S,\pi_P)\star X)\). Hence
\(
\mathcal Q_{\mathrm{opt}}\bigl((\pi_S,\pi_P)\star X\bigr)
=
\{\pi_S \mathbf Q\pi_P^\top: \mathbf Q\in\mathcal Q_{\mathrm{opt}}(X)\}.
\)

Let \(\Phi_{\mathrm{sort}}(X)=(\sigma_S,\sigma_P)\). By equivariance of the
sorting map,
\(
\Phi_{\mathrm{sort}}\bigl((\pi_S,\pi_P)\star X\bigr)
=
(\pi_S\circ\sigma_S,\pi_P\circ\sigma_P).
\)
The lexicographic order is transported by the same relabeling: for any
\(\mathbf Q,\mathbf Q'\),
\(
\operatorname{vec}_{\pi_S\circ\sigma_S,\pi_P\circ\sigma_P}
(\pi_S \mathbf Q\pi_P^\top)
=
\operatorname{vec}_{\sigma_S,\sigma_P}(\mathbf Q).
\)
Therefore the lexicographically smallest element of
\(\mathcal Q_{\mathrm{opt}}((\pi_S,\pi_P)\star X)\) is exactly
\(\pi_S\Phi_Q(X)\pi_P^\top\). This proves the claim.
\qed
\endproof

\subsection{Vectorization of Edge Outputs}
\label{subsec:R_mn_equiv}
Let \(\mathcal Y_{m,n}:=\mathbb R^{m\times n}\), endowed with entrywise
addition and Hadamard multiplication. Thus \(\mathcal Y_{m,n}\) is a
finite-dimensional commutative algebra. Let
\(\Gamma_{m,n}:=S_m\times S_n\) act on \(\mathcal Y_{m,n}\) by
\(
(\pi_S,\pi_P)\star \mathbf Q:=\pi_S \mathbf Q\pi_P^\top .
\)
We identify \(\mathbb R^{m\times n}\) with \(\mathbb R^{mn}\) through the
row-major vectorization map \(\operatorname{vec}\). Under this identification,
the action of \((\pi_S,\pi_P)\in S_m\times S_n\) is represented by the linear
map \(\rho_{m,n}(\pi_S,\pi_P)\) defined by
\[
\rho_{m,n}(\pi_S,\pi_P)\operatorname{vec}(\mathbf Q)
:=
\operatorname{vec}(\pi_S \mathbf Q\pi_P^\top),
\qquad \mathbf Q\in\mathbb R^{m\times n}.
\]
Thus edge-level permutation equivariance,
\(
\Phi((\pi_S,\pi_P)\star X)=\pi_S\Phi(X)\pi_P^\top,
\)
is exactly \(\rho_{m,n}\)-equivariance of
\(\operatorname{vec}\circ\Phi\).

\subsection{Edge-Side WL Refinement and Indistinguishability}

Recall that for each instance $X=(G,H,E)\in \mathcal X_{m,n}$, the
node-side WL procedure produces colors $C_k^{l,S}(X)$ for customer nodes
and $C_j^{l,P}(X)$ for product nodes at every iteration $l\ge 0$.

The next definition describes indistinguishability of edge coordinates
within a fixed instance.

\begin{definition}[Within-instance edge indistinguishability]
\label{def:within_instance_edge_indist}
Let $X=(G,H,E)\in \mathcal X_{m,n}$. For two edge indices
$(k,j),(k',j')\in [m]\times [n]$, we write
$(k,j)\equiv_x^E (k',j')$ if
$C_{kj}^{l,E}(X)=C_{k'j'}^{l,E}(X)$ for every $l\ge 0$.
\end{definition}

In words, $(k,j)\equiv_x^E (k',j')$ means that the two edge coordinates
cannot be distinguished by the edge-side WL refinement on the instance
$X$.

\begin{definition}[Cross-instance edge-coordinate WL equivalence]
\label{def:cross_instance_edge_equiv}
Let \(X=(G,H,E)\in\mathcal X_{m,n}\) and
\(\widehat X=(\widehat G,\widehat H,\widehat E)\in\mathcal X_{m,n}\).
We write
\(
X\sim_E \widehat X
\)
if the edge-side WL colors agree coordinatewise at every refinement level:
\[
C_{kj}^{l,E}(X)=C_{kj}^{l,E}(\widehat X),
\qquad
\forall (k,j)\in[m]\times[n],\ \forall l\ge 0.
\]
\end{definition}

\begin{remark}
\label{rem:edge_wl_two_roles}
The relation \(\sim\) compares two instances up to simultaneous relabeling. The relation \(\sim_E\) compares
two instances with edge coordinates fixed and characterizes exact equality
of all edge-output functions. By contrast, \(\equiv_x^E\) is defined within
a fixed instance and is used to describe when two edge coordinates cannot be
separated by the edge-output function class.
\end{remark}

\begin{lemma}
\label{lem:layerwise_alignment_edge}
Let \(X,\widehat X\in\mathcal X_{m,n}\), and fix an \(L\)-layer logit-output GNN. Suppose that there exists
\((\pi_S,\pi_P)\in S_m\times S_n\) such that the level-\(L\) WL colors are
aligned as
\[
C_k^{L,S}(X)=C_{\pi_S(k)}^{L,S}(\widehat X),
\quad
C_j^{L,P}(X)=C_{\pi_P(j)}^{L,P}(\widehat X),
\quad
C_{kj}^{L,E}(X)
=
C_{\pi_S(k)\pi_P(j)}^{L,E}(\widehat X),
\qquad
\forall k\in[m],\ j\in[n].
\]

Then, for every \(\ell=0,1,\dots,L\),
\[
\mathbf v_k^{S,(\ell)}(X)
=
\mathbf v_{\pi_S(k)}^{S,(\ell)}(\widehat X),
\qquad
\forall k\in[m],
\]
\[
\mathbf v_j^{P,(\ell)}(X)
=
\mathbf v_{\pi_P(j)}^{P,(\ell)}(\widehat X),
\qquad
\forall j\in[n],
\]
\[
\mathbf e_{jk}^{(\ell)}(X)
=
\mathbf e_{\pi_P(j)\pi_S(k)}^{(\ell)}(\widehat X),
\qquad
\forall j\in[n],\ k\in[m].
\]
\end{lemma}

\proof{Proof.}
Since each WL update contains the previous color, level-\(L\) alignment
implies the same alignment at all levels \(\ell\le L\). We prove the claim
by induction on \(\ell\). For \(\ell=0\), the result follows from the
injective initial encodings of customer features, product features, and raw
edge features.

Assume the claim holds at level \(\ell-1\). For each customer node, the
aggregated message is a sum over all product nodes using shared message maps.
By the induction hypothesis, the raw edge-feature alignment, and the change of
variables \(j'=\pi_P(j)\), the segment-side message of node \(k\) in \(X\)
equals the segment-side message of node \(\pi_S(k)\) in \(\widehat X\).
The shared customer update map therefore gives
\(\mathbf v^{S,(\ell)}_k(X)=\mathbf v^{S,(\ell)}_{\pi_S(k)}(\widehat X)\). The product-side
argument is identical, using the change of variables \(k'=\pi_S(k)\).

Finally, the edge-update input is a shared function of the current endpoint
embeddings and the previous edge embedding. All these components are aligned
by the induction hypothesis and the node-alignment just proved. Applying the
shared edge update map yields
\(
\mathbf e_{jk}^{(\ell)}(X)
=
\mathbf e_{\pi_P(j)\,\pi_S(k)}^{(\ell)}(\widehat X),
\qquad
\forall j\in[n],\ k\in[m].
\)
This completes the induction.
\qed
\endproof

\begin{lemma}
\label{lem:forward_edge_indistinguishability}
Let \(X,\widehat X\in\mathcal X_{m,n}\). If \(X\sim\widehat X\), then the
following hold:
\begin{enumerate}
    \item[(i)] \(f(X)=f(\widehat X)\) for all
    \(f\in\mathcal F_{\mathrm{edge}}^{1}\).

    \item[(ii)] For every \(F\in\mathcal F_{\mathrm{edge}}^{W}\), there
    exists \((\pi_S,\pi_P)\in S_m\times S_n\) such that
    \(
    F(\widehat X)=\pi_S F(X)\pi_P^\top.
    \)
\end{enumerate}
\end{lemma}

\proof{Proof.}
Let \(F\in\mathcal F_{\mathrm{edge}}^{W}\) be realized by a logit-output GNN of
depth \(\omega\). Since \(X\sim\widehat X\), the level-\(\omega\) WL color
configurations of \(X\) and \(\widehat X\) are aligned by some
\((\pi_S,\pi_P)\). By Lemma~\ref{lem:layerwise_alignment_edge}, the terminal
hidden states are aligned:
\[
\mathcal U^{(\omega)}(\widehat X)
=
(\pi_S,\pi_P)\star \mathcal U^{(\omega)}(X),
\qquad
\mathbf e_{\pi_P(j)\pi_S(k)}^{(\omega)}(\widehat X)
=
\mathbf e_{jk}^{(\omega)}(X).
\]
Using the edge-readout
\(
F(X)_{kj}
=
R_E\!\left(
\mathbf e_{jk}^{(\omega)}(X),
\mathcal U^{(\omega)}(X)
\right),
\)
and the invariance of \(R_E\), we obtain
\[
F(\widehat X)_{\pi_S(k),\pi_P(j)}
=
R_E\!\left(
\mathbf e_{jk}^{(\omega)}(X),
(\pi_S,\pi_P)\star\mathcal U^{(\omega)}(X)
\right)
=
F(X)_{kj}.
\]
Hence \(F(\widehat X)=\pi_SF(X)\pi_P^\top\), proving (ii).

The scalar-output case is identical. If
\(f(X)=R_1(\mathcal U^{(\omega)}(X))\) with invariant readout \(R_1\), then
\[
f(\widehat X)
=
R_1\!\left((\pi_S,\pi_P)\star\mathcal U^{(\omega)}(X)\right)
=
R_1\!\left(\mathcal U^{(\omega)}(X)\right)
=
f(X).
\]
This proves (i).
\qed
\endproof

\begin{lemma}
\label{lem:finite_color_realization}
Fix \(X,\widehat X\in\mathcal X_{m,n}\) and \(L\in\mathbb N\). Then there
exists an \(L\)-layer logit-output GNN such that, for each
\(\ell=0,1,\dots,L\), there exist injective maps
\[
\eta_\ell^{S}:\mathcal C_\ell^{S}\to \mathbb R^{d_\ell^{S}},
\qquad
\eta_\ell^{P}:\mathcal C_\ell^{P}\to \mathbb R^{d_\ell^{P}},
\qquad
\eta_\ell^{E}:\mathcal C_\ell^{E}\to \mathbb R^{d_\ell^{E}},
\]
where \(\mathcal C_\ell^{S}\), \(\mathcal C_\ell^{P}\), and
\(\mathcal C_\ell^{E}\) denote the sets of customer, product, and edge WL
colors appearing at level \(\ell\) on \(X\) and \(\widehat X\), such that for
every \(X'\in\{X,\widehat X\}\), \(k\in[m]\), and \(j\in[n]\),
\[
\mathbf v_k^{S,(\ell)}(X')
=
\eta_\ell^{S}\bigl(C_k^{\ell,S}(X')\bigr),
\qquad
\mathbf v_j^{P,(\ell)}(X')
=
\eta_\ell^{P}\bigl(C_j^{\ell,P}(X')\bigr),
\qquad
\mathbf e_{jk}^{(\ell)}(X')
=
\eta_\ell^{E}\bigl(C_{kj}^{\ell,E}(X')\bigr).
\]

In particular, if no \((\pi_S,\pi_P)\in S_m\times S_n\) aligns the level-\(L\)
WL color configurations of \(X\) and \(\widehat X\), then no such permutation
pair aligns the corresponding level-\(L\) hidden-state configurations
\(\mathcal U^{(L)}(X)\) and \(\mathcal U^{(L)}(\widehat X)\).
\end{lemma}

\proof{Proof.}
Since only
the two fixed instances \(X,\widehat X\) and finitely many levels
\(\ell\le L\) are involved, only finitely many WL colors and refinement
descriptors appear. Hence the shared continuous maps in the logit-output GNN can be prescribed arbitrarily on these finite sets.

We proceed by induction on \(\ell\). For \(\ell=0\), choose injective code
maps \(\eta_0^S,\eta_0^P,\eta_0^E\) for the initial customer, product, and
edge colors. Since the initial WL colors are injective encodings of the
corresponding raw features, choose the initial embedding maps
\(\iota_S,\iota_P,\iota_E\) so that
\(
\mathbf v_k^{S,(0)}(X')
=
\eta_0^{S}\bigl(C_k^{0,S}(X')\bigr),
\)
\(
\mathbf v_j^{P,(0)}(X')
=
\eta_0^{P}\bigl(C_j^{0,P}(X')\bigr),
\)
and
\(
\mathbf e_{jk}^{(0)}(X')
=
\eta_0^{E}\bigl(C_{kj}^{0,E}(X')\bigr)
\)
for every \(X'\in\{X,\widehat X\}\), \(k\in[m]\), and \(j\in[n]\).

Assume the claim holds up to level \(\ell-1\). Choose the shared message maps
\(\phi_P^{(\ell)}\) and \(\phi_S^{(\ell)}\) so that, on the finite sets of
previous-level color codes,
\(
\phi_P^{(\ell)}
\bigl(\eta_{\ell-1}^{P}(c)\bigr)
=
\mathrm{HASH}'_{\ell,P}(c),
c\in\mathcal C_{\ell-1}^{P},
\)
and
\(
\phi_S^{(\ell)}
\bigl(\eta_{\ell-1}^{S}(c)\bigr)
=
\mathrm{HASH}'_{\ell,S}(c),
c\in\mathcal C_{\ell-1}^{S}.
\)
Then, by Assumption~\ref{ass:wl_hash}, the GNN aggregated messages coincide exactly with the collision-free weighted WL descriptors
\[
\begin{aligned}
\mathbf a_k^{S,(\ell)}(X')
&= \sum_{j=1}^n u_{kj}(X')\,
\mathrm{HASH}'_{\ell,P}\bigl(C_j^{\ell-1,P}(X')\bigr),\\[-1mm]
\mathbf a_j^{P,(\ell)}(X')
&= \sum_{k=1}^m u_{kj}(X')\,
\mathrm{HASH}'_{\ell,S}\bigl(C_k^{\ell-1,S}(X')\bigr).
\end{aligned}
\]
Since the outer WL hashes are injective, the previous color together with the
weighted descriptor determines the new node color. Thus the shared update maps
\(\Gamma_S^{(\ell)}\) and \(\Gamma_P^{(\ell)}\) can be prescribed on the
finite set of encountered update inputs so that
\(
\mathbf v_k^{S,(\ell)}(X')
=
\eta_\ell^{S}\bigl(C_k^{\ell,S}(X')\bigr),
\mathbf v_j^{P,(\ell)}(X')
=
\eta_\ell^{P}\bigl(C_j^{\ell,P}(X')\bigr).
\)

For edges, the new WL color is an injective hash of
\(
\bigl(C_{kj}^{\ell-1,E}(X'),C_k^{\ell,S}(X'),C_j^{\ell,P}(X')\bigr).
\)
By the induction hypothesis and the node-color realization just proved, the
logit-output GNN update input
\(
\bigl(
\mathbf v_j^{P,(\ell)}(X'),\,
\mathbf v_k^{S,(\ell)}(X'),\,
\mathbf e_{jk}^{(\ell-1)}(X')
\bigr)
\)
determines this tuple consistently on the finite set of encountered inputs.
Therefore \(\Gamma_E^{(\ell)}\) can be prescribed so that
\(
\mathbf e_{jk}^{(\ell)}(X')
=
\eta_\ell^{E}\bigl(C_{kj}^{\ell,E}(X')\bigr).
\)
This completes the induction.

For the final statement, if the level-\(L\) hidden-state configurations could
be aligned by some \((\pi_S,\pi_P)\), then the injectivity of
\(\eta_L^S,\eta_L^P,\eta_L^E\) would imply that the level-\(L\) WL color
configurations are aligned by the same permutation pair, a contradiction.
\qed
\endproof

\begin{lemma}
\label{lem:invariant_orbit_separator}
Let \(\Gamma:=S_m\times S_n\) act linearly on a finite-dimensional Euclidean
space \(U\). For any \(u,\widehat u\in U\), suppose that there does not exist
\(g\in \Gamma\) such that
\(
\widehat u=g\star u.
\)
Then there exists a continuous \(\Gamma\)-invariant function
\(\rho:U\to\mathbb R\) such that
\(
\rho(u)\neq \rho(\widehat u).
\)
\end{lemma}

\proof{Proof.}
Since \(\Gamma\) is finite, the two sets
\(
\{\,g\star u : g\in \Gamma\,\}
\)
and
\(
\{\,g\star \widehat u : g\in \Gamma\,\}
\)
are finite. By assumption, they are disjoint. Hence there exists a
continuous function \(\psi:U\to\mathbb R\) such that
\[
\psi\equiv 1 \text{ on } \{\,g\star u : g\in \Gamma\,\},
\qquad
\psi\equiv 0 \text{ on } \{\,g\star \widehat u : g\in \Gamma\,\}.
\]
Define
\(
\rho(v):=\frac{1}{|\Gamma|}\sum_{g\in\Gamma}\psi(g\star v),
\forall v\in U.
\)
Then \(\rho\) is continuous and \(\Gamma\)-invariant. Moreover,
\(
\rho(u)=1, \rho(\widehat u)=0.
\)
Therefore \(\rho(u)\neq \rho(\widehat u)\).
\qed
\endproof

\begin{lemma}
\label{lem:reverse_scalar_edge_separation}
For any \(X,\widehat X\in\mathcal X_{m,n}\), if
\(
f(X)=f(\widehat X),
\forall f\in\mathcal F_{\mathrm{edge}}^{1},
\)
then
\(
X\sim \widehat X.
\)
\end{lemma}

\proof{Proof.}
We prove the statement by contraposition. Assume that
\(
X\not\sim \widehat X.
\)
By Definition~\ref{def:cross_instance_wl_equiv}, there exists
\(L\in\mathbb N\) such that the level-\(L\) customer, product, and edge color
configurations of \(X\) and \(\widehat X\) cannot be matched by any
simultaneous relabeling of customer and product indices.

Apply Lemma~\ref{lem:finite_color_realization} to \(X\), \(\widehat X\), and
\(L\). By the final statement of Lemma~\ref{lem:finite_color_realization},
the constructed level-\(L\) hidden-state configurations
\(
\mathcal U^{(L)}(X)
\)
and \(
\mathcal U^{(L)}(\widehat X)
\)
cannot be matched by any simultaneous relabeling of customer and product
indices.

Let \(\Gamma:=S_m\times S_n\). Applying
Lemma~\ref{lem:invariant_orbit_separator}, there exists a continuous
\(\Gamma\)-invariant function \(\rho\) such that
\(
\rho\bigl(\mathcal U^{(L)}(X)\bigr)
\neq
\rho\bigl(\mathcal U^{(L)}(\widehat X)\bigr).
\)
Define
\(f(X'):=\rho\bigl(\mathcal U^{(L)}(X')\bigr),
\quad \forall X'\in\mathcal X_{m,n}.
\)
Since \(\rho\) is continuous and \(\Gamma\)-invariant, \(f\) is a continuous
permutation-invariant scalar-valued function realizable by the logit-output GNN. Hence \(f\in\mathcal F_{\mathrm{edge}}^{1}\).
Moreover,
\(
f(X)=\rho\bigl(\mathcal U^{(L)}(X)\bigr)
\neq
\rho\bigl(\mathcal U^{(L)}(\widehat X)\bigr)
=f(\widehat X).
\)
Thus, if \(X\not\sim\widehat X\), there exists
\(f\in\mathcal F_{\mathrm{edge}}^{1}\) such that
\(f(X)\neq f(\widehat X)\). The contrapositive proves the lemma.
\qed
\endproof

\proof{Proof of Theorem~\ref{thm:edge_thm42}.}
The implication \((i)\Rightarrow(ii)\) follows from
Lemma~\ref{lem:forward_edge_indistinguishability}(i), and
\((i)\Rightarrow(iii)\) follows from
Lemma~\ref{lem:forward_edge_indistinguishability}(ii).

The implication \((ii)\Rightarrow(i)\) follows from
Lemma~\ref{lem:reverse_scalar_edge_separation}.

It remains to prove \((iii)\Rightarrow(ii)\). Let
\(f\in\mathcal F_{\mathrm{edge}}^{1}\) be arbitrary. By Remark~\ref{rem:scalar_edge_lift},
\(
F_f(X'):=f(X')\mathbf 1_{m\times n},
\forall X'\in\mathcal X_{m,n},
\)
belongs to \(\mathcal F_{\mathrm{edge}}^{W}\). Applying (iii) to \(F_f\),
there exists \((\pi_S,\pi_P)\in S_m\times S_n\) such that
\(
F_f(\widehat X)=\pi_S F_f(X)\pi_P^\top.
\)
Since permutation matrices leave \(\mathbf 1_{m\times n}\) invariant, we have
\[
f(\widehat X)\mathbf 1_{m\times n}
=
\pi_S\bigl(f(X)\mathbf 1_{m\times n}\bigr)\pi_P^\top
=
f(X)\mathbf 1_{m\times n}.
\]
Therefore \(f(\widehat X)=f(X)\). Since
\(f\in\mathcal F_{\mathrm{edge}}^{1}\) was arbitrary, (ii) follows.

Thus (i), (ii), and (iii) are equivalent.
\qed
\endproof

\begin{lemma}
\label{lem:edge_exact_output_indistinguishability}
For any \(X,\widehat X\in\mathcal X_{m,n}\), the following are equivalent:
\begin{enumerate}
    \item[(i)] \(X\sim_E \widehat X\);

    \item[(ii)] \(F(X)=F(\widehat X)\) for every
    \(F\in\mathcal F_{\mathrm{edge}}^{W}\).
\end{enumerate}
\end{lemma}

\proof{Proof.}
We first prove \((i)\Rightarrow(ii)\). Suppose \(X\sim_E\widehat X\). Then
the edge-side WL colors agree coordinatewise at every refinement level:
\[
C_{kj}^{\ell,E}(X)=C_{kj}^{\ell,E}(\widehat X),
\qquad
\forall (k,j)\in[m]\times[n],\ \forall \ell\ge0.
\]
Since \(C_{kj}^{0,E}\) is an injective hash of
\((C_k^{0,S},C_j^{0,P},u_{kj})\), coordinatewise equality of level-zero edge
colors implies coordinatewise equality of the initial customer colors,
product colors, and raw edge features. For \(\ell\ge1\), the edge-color update
is injective in \((C_{kj}^{\ell-1,E},C_k^{\ell,S},C_j^{\ell,P})\), so
coordinatewise edge-color equality also aligns the corresponding node-side
colors at every refinement level.
Thus the WL colors of \(X\) and \(\widehat X\) are aligned under the identity
permutations on \([m]\) and \([n]\).

Let \(F\in\mathcal F_{\mathrm{edge}}^{W}\) be realized by a logit-output GNN of depth \(\omega\). Applying
Lemma~\ref{lem:layerwise_alignment_edge} with the identity permutations gives
\[
\mathcal U^{(\omega)}(X)=\mathcal U^{(\omega)}(\widehat X),
\qquad
\mathbf e_{jk}^{(\omega)}(X)
=
\mathbf e_{jk}^{(\omega)}(\widehat X),
\quad \forall j\in[n],\ k\in[m].
\]
Therefore, using the shared edge readout,
\[
F(X)_{kj}
=
R_E\!\left(\mathbf e_{jk}^{(\omega)}(X),
\mathcal U^{(\omega)}(X)\right)
=
R_E\!\left(\mathbf e_{jk}^{(\omega)}(\widehat X),
\mathcal U^{(\omega)}(\widehat X)\right)
=
F(\widehat X)_{kj}.
\]
Hence \(F(X)=F(\widehat X)\).

We prove \((ii)\Rightarrow(i)\) by contraposition. If
\(X\not\sim_E\widehat X\), then for some refinement level \(L\ge0\) and some
edge coordinate \((k,j)\),
\(
C_{kj}^{L,E}(X)\neq C_{kj}^{L,E}(\widehat X).
\)
By Lemma~\ref{lem:finite_color_realization}, there exists an \(L\)-layer
logit-output GNN whose terminal edge embeddings injectively encode
level-\(L\) edge colors. Hence
\(
\mathbf e_{jk}^{(L)}(X)\neq \mathbf e_{jk}^{(L)}(\widehat X).
\)
Let
\(
\mathbf v:=\mathbf e_{jk}^{(L)}(X)-\mathbf e_{jk}^{(L)}(\widehat X)\neq0,
\)
and choose the continuous edge readout
\(
R_E(\mathbf e, \mathcal U):=\langle \mathbf e,\mathbf v\rangle,
\)
which ignores the global-state argument \(\mathcal U\). This readout is admissible and
defines some \(F\in\mathcal F_{\mathrm{edge}}^{W}\). At coordinate \((k,j)\),
\[
F(X)_{kj}-F(\widehat X)_{kj}
=
\left\langle
\mathbf e_{jk}^{(L)}(X)-\mathbf e_{jk}^{(L)}(\widehat X),\mathbf v
\right\rangle
=
\|\mathbf v\|^2>0.
\]
Thus \(F(X)\neq F(\widehat X)\), proving the contrapositive.
\qed
\endproof

For any \(\varepsilon\in(0,\tfrac12)\), define the auxiliary logit target
\[
\Psi_Q^{(\varepsilon)}(X)_{kj}
=
\begin{cases}
1+\log(1/\varepsilon-1), & \text{if } \Phi_Q(X)_{kj}=1,\\
-1-\log(1/\varepsilon-1), & \text{if } \Phi_Q(X)_{kj}=0.
\end{cases}
\]
This is the target approximated by the logit-output class
\(\mathcal F_{\mathrm{edge}}^W\). The sigmoid-output class
\(\mathcal F_{\mathrm{edge}}^{W,\sigma}\) is then obtained by composing the
approximating logit-output GNN with the sigmoid function.

\begin{lemma}
\label{lem:target_consistency_psi}
Let \(X,\widehat X\in\mathcal D_{\mathrm{solu}}\setminus
\mathcal D_{\mathrm{foldable}}\). If \(X\sim_E\widehat X\), then, for every
\(\varepsilon\in(0,\tfrac12)\),
\(
\Psi_Q^{(\varepsilon)}(X)=\Psi_Q^{(\varepsilon)}(\widehat X).
\)
\end{lemma}

\proof{Proof.}
Since \(X\sim_E\widehat X\), the level-zero edge colors agree coordinatewise:
\(C_{kj}^{0,E}(X)=C_{kj}^{0,E}(\widehat X)\) for all \(k\in[m]\) and
\(j\in[n]\). By the injectivity of
\(C_{kj}^{0,E}=\mathrm{HASH}_{0,E}(C_k^{0,S},C_j^{0,P},u_{kj})\), the initial
customer colors, product colors, and raw edge features coincide
coordinatewise. Since \(\mathrm{HASH}_{0,S}\) and \(\mathrm{HASH}_{0,P}\) are
injective, the raw customer and product features also coincide coordinatewise.
Hence the two segment-product graph instances are identical, i.e.,
\(X=\widehat X\). Therefore \(\Phi_Q(X)=\Phi_Q(\widehat X)\), and the
definition of \(\Psi_Q^{(\varepsilon)}\) gives
\(\Psi_Q^{(\varepsilon)}(X)=\Psi_Q^{(\varepsilon)}(\widehat X)\).
\qed
\endproof

Let \(C_{\mathrm{eq}}(\mathcal X_{m,n},\mathbb R^{m\times n})\) denote
the set of continuous edge-equivariant maps.

\begin{lemma}[Subalgebra property of \(\mathcal F_{\mathrm{edge}}^{W}\)]
\label{lem:subalgebra_edgeW}
The class \(\mathcal F_{\mathrm{edge}}^{W}\) is a subalgebra of
\(C_{\mathrm{eq}}(\mathcal X_{m,n},\mathbb R^{m\times n})\). In particular:

\begin{enumerate}
    \item[(i)] if \(F,\widehat F\in \mathcal F_{\mathrm{edge}}^{W}\), then
    \(F+\widehat F\in \mathcal F_{\mathrm{edge}}^{W}\);

    \item[(ii)] if \(F,\widehat F\in \mathcal F_{\mathrm{edge}}^{W}\), then
    \(F\odot \widehat F\in \mathcal F_{\mathrm{edge}}^{W}\), where
    \((F\odot \widehat F)(X):=F(X)\odot \widehat F(X)\) denotes the
    entrywise product;

    \item[(iii)] for every \(c\in\mathbb R\), the constant matrix-valued map
    \(X\mapsto c\,\mathbf 1_{m\times n}\) belongs to
    \(\mathcal F_{\mathrm{edge}}^{W}\).
\end{enumerate}
\end{lemma}

\proof{Proof.}
The proof follows the parallel-network construction used to prove
the subalgebra property of GNN function classes in
\citet{chen2023lp}. We only highlight the minor adaptation needed for our
edge-output architecture.

First, every map in \(\mathcal F_{\mathrm{edge}}^{W}\) is continuous and
edge-equivariant by construction, since all node updates, edge updates, and
edge readouts are shared across customer, product, and edge coordinates. Hence
\(
\mathcal F_{\mathrm{edge}}^{W}
\subset
C_{\mathrm{eq}}(\mathcal X_{m,n},\mathbb R^{m\times n}).
\)

Let \(F,\widehat F\in\mathcal F_{\mathrm{edge}}^{W}\). As in the parallel
construction of \citet{chen2023lp}, we may pad the shallower network with
identity layers and assume that the two networks have the same depth. Run the
two networks in parallel by concatenating their segment-node, product-node,
and edge hidden states at every layer:
\[
\widetilde{\mathbf v}^{S,(\ell)}_k
=
\mathbf v^{S,(\ell)}_k\Vert \widehat{\mathbf v}^{S,(\ell)}_k,\qquad
\widetilde{\mathbf v}^{P,(\ell)}_j
=
\mathbf v^{P,(\ell)}_j\Vert \widehat{\mathbf v}^{P,(\ell)}_j,\qquad
\widetilde{\mathbf e}^{(\ell)}_{jk}
=
\mathbf e^{(\ell)}_{jk}\Vert \widehat{\mathbf e}^{(\ell)}_{jk}.
\]
Because the original update maps are shared and continuous, the componentwise
parallel updates again define a valid logit-output GNN.

It remains only to choose the final shared edge readout. If \(R_E\) and
\(\widehat R_E\) are the readouts realizing \(F\) and \(\widehat F\), define
\(
\widetilde R_+(\mathbf e,\widehat{\mathbf e},\mathcal U,\widehat{\mathcal U})
=
R_E(\mathbf e, \mathcal U)+\widehat R_E(\widehat{\mathbf e},\widehat{\mathcal U}).
\)
The resulting parallel network realizes \(F+\widehat F\). Similarly, the
readout
\(
\widetilde R_\odot(\mathbf e,\widehat{\mathbf e},\mathcal{U},\widehat{\mathcal{U}})
=
R_E(\mathbf e, \mathcal{U})\widehat R_E(\widehat{\mathbf e},\widehat{\mathcal{U}})
\)
realizes the entrywise product \(F\odot\widehat F\). Finally, the constant map
\(X\mapsto c\,\mathbf 1_{m\times n}\) is realized by a trivial logit-output GNN with
constant edge readout \(R_E\equiv c\). Therefore
\(\mathcal F_{\mathrm{edge}}^{W}\) is closed under addition, entrywise
multiplication, and constants, and hence is a subalgebra of
\(C_{\mathrm{eq}}(\mathcal X_{m,n},\mathbb R^{m\times n})\).
\qed
\endproof

\begin{lemma}
\label{lem:edge_coordinate_separation_unfoldable}
Let
\(
X\in \mathcal X_{m,n}\setminus \mathcal D_{\mathrm{foldable}}.
\)
Then for any two distinct edge coordinates
\(
(k,j)\neq (k',j')\in [m]\times [n],
\)
one has
\(
(k,j)\not\equiv_x^E (k',j').
\)
Consequently, there exists
\(
F\in \mathcal F_{\mathrm{edge}}^{W}
\)
such that
\(
F(X)_{kj}\neq F(X)_{k'j'}.
\)
\end{lemma}

\proof{Proof.}
Let \(T_\star:=m+n\). By
Lemma~\ref{lem:unfoldable_terminal_discreteness}, the segment-side and
product-side WL colors are pairwise distinct at level \(T_\star\):
\[
C_k^{T_\star,S}(X)\neq C_{k'}^{T_\star,S}(X)\quad(k\neq k'),
\qquad
C_j^{T_\star,P}(X)\neq C_{j'}^{T_\star,P}(X)\quad(j\neq j').
\]
Hence for any two distinct edge coordinates \((k,j)\neq(k',j')\), at least
one endpoint color differs at level \(T_\star\). Since the edge hash at level
\(T_\star\) is injective in
\(
\bigl(C_{kj}^{T_\star-1,E},C_k^{T_\star,S},C_j^{T_\star,P}\bigr),
\)
it follows that
\(
C_{kj}^{T_\star,E}(X)\neq C_{k'j'}^{T_\star,E}(X).
\)
Therefore \((k,j)\not\equiv_x^E(k',j')\).

It remains to realize this coordinate separation by an edge-output GNN. Apply
Lemma~\ref{lem:finite_color_realization} with the pair \((X,X)\) and
\(L=T_\star\). Then there exists a \(T_\star\)-layer logit-output GNN
whose terminal edge embeddings injectively encode the level-\(T_\star\) edge
colors. Thus
\(
\mathbf e_{jk}^{(T_\star)}(X)
\neq
\mathbf e_{j'k'}^{(T_\star)}(X).
\)
Let
\(
\mathbf v:=
\mathbf e_{jk}^{(T_\star)}(X)
-
\mathbf e_{j'k'}^{(T_\star)}(X)
\neq0,
\)
and choose
\(
R_E(\mathbf e, \mathcal{U}):=\langle \mathbf e,\mathbf v\rangle.
\)
This admissible readout defines some \(F\in\mathcal F_{\mathrm{edge}}^{W}\),
and
\[
F(X)_{kj}-F(X)_{k'j'}
=
\left\langle
\mathbf e_{jk}^{(T_\star)}(X)
-
\mathbf e_{j'k'}^{(T_\star)}(X),\mathbf v
\right\rangle
=
\|\mathbf v\|^2>0.
\]
Hence \(F(X)_{kj}\neq F(X)_{k'j'}\).
\qed
\endproof

The approximation step relies on the generalized Stone--Weierstrass theorem.
Let \(G\) be a finite group acting continuously on a compact topological space
\(\Omega\) and on \(\mathbb R^r\). Define
\[
C_E(\Omega,\mathbb R^r)
:=
\{F\in C(\Omega,\mathbb R^r): F(g\star x)=g\star F(x),\
\forall x\in\Omega,\ \forall g\in G\}.
\]

\begin{theorem}[Generalized Stone--Weierstrass theorem {\normalfont\cite[Theorem~E.2]{chen2023lp}}]
\label{thm:generalized_stone_weierstrass}
Let \(\mathcal F\subset C_E(\Omega,\mathbb R^r)\) and let
\(\Phi\in C_E(\Omega,\mathbb R^r)\). Suppose that:
\begin{enumerate}
    \item[(i)] \(\mathcal F\) is a subalgebra of \(C(\Omega,\mathbb R^r)\), and
    \(\mathbf 1\in\mathcal F\);

    \item[(ii)] for any \(x,x'\in\Omega\), if
    \(f(x)=f(x')\) for every \(f\in C(\Omega,\mathbb R)\) such that
    \(f\mathbf 1\in\mathcal F\), then for every \(F\in\mathcal F\), there
    exists \(g\in G\) such that
    \(
    F(x)=g\star F(x');
    \)

    \item[(iii)] for any \(x,x'\in\Omega\), if
    \(F(x)=F(x')\) for every \(F\in\mathcal F\), then
    \(
    \Phi(x)=\Phi(x');
    \)

    \item[(iv)] for any \(x\in\Omega\), it holds that
    \(\Phi(x)_j=\Phi(x)_{j'}\) for all \((j,j')\in J(x)\), where
    \[
    J(x):=
    \{(j,j')\in\{1,\dots,r\}^2:
    F(x)_j=F(x)_{j'}\text{ for every }F\in\mathcal F\}.
    \]
\end{enumerate}
Then for any \(\varepsilon>0\), there exists
\(F_\varepsilon\in\mathcal F\) such that
\[
\sup_{x\in\Omega}\|\Phi(x)-F_\varepsilon(x)\|<\varepsilon.
\]
\end{theorem}

\proof{Proof of Theorem~\ref{thm:sigmoid_assignment_recovery}.}
Let
\(
\Gamma_{m,n}:=S_m\times S_n
\)
denote the segment-product permutation group. If \(D\) is not
\(\Gamma_{m,n}\)-invariant, replace it by its finite orbit closure
\(
\mathcal X
:=
\{\gamma \star X:\ X\in D,\ \gamma\in\Gamma_{m,n}\}.
\)
It suffices to prove the result on \(\mathcal X\), since \(D\subseteq\mathcal X\).
Moreover, by construction, \(\mathcal X\) is finite and
\(\Gamma_{m,n}\)-invariant. Since solvability and unfoldability are invariant
under segment-product permutations, we have
\(
\mathcal X
\subset
\mathcal D_{\mathrm{solu}}\setminus\mathcal D_{\mathrm{foldable}}.
\)
Under the subspace topology inherited from the graph-input space,
\(\mathcal X\) is finite and hence compact, and every function on
\(\mathcal X\) is continuous.

Fix \(\varepsilon\in(0,\tfrac12)\). Recall the auxiliary logit target
\(\Psi_Q^{(\varepsilon)}\) defined above. Since \(\Phi_Q\) is permutation
equivariant, \(\Psi_Q^{(\varepsilon)}\) is also permutation equivariant.

We now apply Theorem~\ref{thm:generalized_stone_weierstrass} under the
matrix-vector identification of \(\mathbb R^{m\times n}\) with
\(\mathbb R^{mn}\). Let
\(
\mathcal A
:=
\{F|_{\mathcal X}:\ F\in\mathcal F_{\mathrm{edge}}^W\}
\)
be the restriction of the logit-output GNN class to \(\mathcal X\). We
take the target function to be
\(
\Psi_Q^{(\varepsilon)}:\mathcal X\to\mathbb R^{m\times n}.
\)

We verify the conditions of Theorem~\ref{thm:generalized_stone_weierstrass}.
First, \(\mathcal A\) is a subalgebra containing constants by
Lemma~\ref{lem:subalgebra_edgeW}; restriction to the finite set
\(\mathcal X\) preserves these operations.

Second, suppose \(X,\widehat X\in\mathcal X\) satisfy
\(h(X)=h(\widehat X)\) for every scalar continuous function \(h\) on
\(\mathcal X\) such that \(h\mathbf 1_{m\times n}\in\mathcal A\).
For any \(f\in\mathcal F_{\mathrm{edge}}^1\), Remark~\ref{rem:scalar_edge_lift}
implies \(f\mathbf 1_{m\times n}\in\mathcal F_{\mathrm{edge}}^W\). Hence
\((f|_{\mathcal X})\mathbf 1_{m\times n}\in\mathcal A\), and therefore
\(f(X)=f(\widehat X)\) for every \(f\in\mathcal F_{\mathrm{edge}}^1\).
By Theorem~\ref{thm:edge_thm42}, \(X\sim \widehat X\). Applying
Theorem~\ref{thm:edge_thm42} again, for every
\(F\in\mathcal F_{\mathrm{edge}}^W\), there exists a permutation pair
\((\pi_S^F,\pi_P^F)\in S_m\times S_n\), possibly depending on \(F\), such that
\(
F(\widehat X)=\pi_S^F F(X)(\pi_P^F)^\top .
\)
Equivalently, using the inverse permutation pair, for every
\(F_D=F|_{\mathcal X}\in\mathcal A\), there exists
\(g_F\in\Gamma_{m,n}\) such that
\[
F_D(X)=g_F\star F_D(\widehat X).
\]

Third, suppose that
\(F_D(X)=F_D(\widehat X)\) for every \(F_D\in\mathcal A\). Then
\(F(X)=F(\widehat X)\) for every \(F\in\mathcal F_{\mathrm{edge}}^W\). By
Lemma~\ref{lem:edge_exact_output_indistinguishability}, this implies \(X\sim_E\widehat X\). Since
\(X,\widehat X\in\mathcal D_{\mathrm{solu}}\setminus
\mathcal D_{\mathrm{foldable}}\), Lemma~\ref{lem:target_consistency_psi}
gives
\(\Psi_Q^{(\varepsilon)}(X)=\Psi_Q^{(\varepsilon)}(\widehat X)\), which
verifies the target-consistency condition.

Fourth, for every \(X\in\mathcal X\), Lemma~\ref{lem:edge_coordinate_separation_unfoldable}
implies that any two distinct product-segment edge coordinates can be
separated by some \(F\in\mathcal F_{\mathrm{edge}}^W\), and hence by an
element of \(\mathcal A\). Therefore the coordinate-indistinguishability
relation contains only identical coordinate pairs, so the required coordinate
consistency of \(\Psi_Q^{(\varepsilon)}(X)\) is automatic.

All conditions of Theorem~\ref{thm:generalized_stone_weierstrass} are
satisfied for the logit target
\(\Psi_Q^{(\varepsilon)}\). We apply the theorem under the matrix-vector identification of
\(\mathbb R^{m\times n}\) with \(\mathbb R^{mn}\), equipped with the
infinity norm. Applying the theorem with approximation tolerance
\(1\), there exists \(F_\varepsilon\in\mathcal F_{\mathrm{edge}}^W\) such that
\[
\sup_{X\in\mathcal X}
\max_{k\in[m],\,j\in[n]}
\left|
F_\varepsilon(X)_{kj}
-
\Psi_Q^{(\varepsilon)}(X)_{kj}
\right|
<1.
\]
Equivalently,
\[
\left|
F_\varepsilon(X)_{kj}
-
\Psi_Q^{(\varepsilon)}(X)_{kj}
\right|
<1,
\qquad
\forall X\in\mathcal X,\ k\in[m],\ j\in[n].
\]
We have
\(
F_\varepsilon^\sigma=\sigma\circ F_\varepsilon\in\mathcal F_{\mathrm{edge}}^{W,\sigma}.
\)
Now fix \(X\in\mathcal X\), \(k\in[m]\), and \(j\in[n]\). If \(\Phi_Q(X)_{kj}=1\), then
\[
\Psi_Q^{(\varepsilon)}(X)_{kj}
=
1+\log(1/\varepsilon-1).
\]
Using the coordinatewise approximation bound,
\[
F_\varepsilon(X)_{kj}
>
1+\log(1/\varepsilon-1)-1
=
\log(1/\varepsilon-1).
\]
Therefore,
\[
F_\varepsilon^\sigma(X)_{kj}
=
\sigma(F_\varepsilon(X)_{kj})
>
\sigma(\log(1/\varepsilon-1))
=
1-\varepsilon.
\]
If \(\Phi_Q(X)_{kj}=0\), then
\[
\Psi_Q^{(\varepsilon)}(X)_{kj}
=
-1-\log(1/\varepsilon-1).
\]
Using the coordinatewise approximation bound,
\[
F_\varepsilon(X)_{kj}
<
-1-\log(1/\varepsilon-1)+1
=
-\log(1/\varepsilon-1).
\]
Therefore,
\[
F_\varepsilon^\sigma(X)_{kj}
=
\sigma(F_\varepsilon(X)_{kj})
<
\sigma(-\log(1/\varepsilon-1))
=
\varepsilon.
\]

Thus, for every \(X\in\mathcal X\), \(k\in[m]\), and \(j\in[n]\),
\[
F_\varepsilon^\sigma(X)_{kj}>1-\varepsilon
\quad
\text{if }
\Phi_Q(X)_{kj}=1,
\]
and
\[
F_\varepsilon^\sigma(X)_{kj}<\varepsilon
\quad
\text{if }
\Phi_Q(X)_{kj}=0.
\]
Since \(F_\varepsilon^\sigma(X)_{kj}\in(0,1)\), these inequalities imply
\[
\left|
F_\varepsilon^\sigma(X)_{kj}
-
\Phi_Q(X)_{kj}
\right|
<
\varepsilon,
\qquad
\forall X\in\mathcal X,\ k\in[m],\ j\in[n].
\]
Because \(\varepsilon<1/2\), each coordinate of
\(F_\varepsilon^\sigma(X)\) lies on the same side of the threshold \(1/2\) as
the corresponding binary coordinate of \(\Phi_Q(X)\). Hence,
\[
\mathbbm 1_{\{F_\varepsilon^\sigma(X)_{kj}>1/2\}}
=
\Phi_Q(X)_{kj},
\qquad
\forall X\in D, k \in [m], j \in [n].
\]
Since \(D\subseteq\mathcal X\), the same conclusions hold for all
\(X\in D\). This proves the theorem.
\qed
\endproof

\section{Proof of Equivalence Between Price Subadditivity Constraints}
\label{proof:subadd}

\begin{proposition}
\label{prop:sp2_spk_eq}
\(S_{P,2}\) and \(S_{P,K}\) are equivalent on the full
bundle universe \(\mathfrak F=2^{[n]}\), or equivalently on the full bundle
index set \(\mathcal K\).
\end{proposition}

\proof{Proof.}
We prove the proposition in both directions.

\textit{Direction 1:} \(S_{P,K}\implies S_{P,2}\).
This follows directly because \(S_{P,2}\) is the special case of \(S_{P,K}\)
with \(K=2\).

\textit{Direction 2:} \(S_{P,2}\implies S_{P,K}\).
We use induction on the number of partition blocks. The base case \(K=2\) is
exactly \(S_{P,2}\). Suppose the claim holds for \(K=k\). We prove it
for \(K=k+1\). Let \(b,b_1,\ldots,b_{k+1}\in\mathcal K\) be such that
\(\mathcal A(b_1),\ldots,\mathcal A(b_{k+1})\) form a pairwise disjoint partition of \(\mathcal{A}(b)\);
equivalently,
\[
\mathcal{A}(b)=\bigcup_{r=1}^{k+1}\mathcal A(b_r),
\qquad
\mathcal A(b_r)\cap \mathcal A(b_s)=\emptyset,\quad \forall r\ne s.
\]
Let
\(
\bar b
:=
\mathcal A^{-1}\!\left(\bigcup_{r=1}^{k}\mathcal A(b_r)\right).
\)
Because \(\mathfrak F=2^{[n]}\), this bundle index \(\bar b\in\mathcal K\)
exists. Moreover, \(\mathcal A(\bar b)\) and \(\mathcal A(b_{k+1})\) form a pairwise disjoint
partition of \(\mathcal{A}(b)\). By \(S_{P,2}\),
\(
p_b \le p_{\bar b}+p_{b_{k+1}}.
\)
By the induction hypothesis applied to \(\mathcal A(\bar b)\), we have
\(
p_{\bar b}\le \sum_{r=1}^{k}p_{b_r}.
\)
Substituting this inequality into the preceding one yields
\(
p_b \le \sum_{r=1}^{k+1}p_{b_r}.
\)
Thus, \(S_{P,2}\implies S_{P,K}\).
\qed
\endproof

\begin{proposition}
\label{prop:subadd_eq}
Under Assumption~\ref{ass:price_monotonicity}, the family of partition
subadditivity \(S_{P,K}\) and the family of cover subadditivity \(S_{C,K}\) are equivalent on the full bundle universe
\(\mathfrak F=2^{[n]}\).
\end{proposition}

\proof{Proof.}
We prove the proposition in both directions under Assumption~\ref{ass:price_monotonicity}.

\textit{Direction 1:} \(S_{C,K}\implies S_{P,K}\).
A pairwise disjoint partition is a special case of a cover. Therefore, if the
cover inequality holds for every \(K\)-way cover, then it also holds for every
\(K\)-way partition.

\textit{Direction 2:} \(S_{P,K}\implies S_{C,K}\).
By Proposition~\ref{prop:sp2_spk_eq}, \(S_{P,K}\) is equivalent to \(S_{P,2}\) on the full bundle
universe, so it suffices to use \(S_{P,2}\) together with price monotonicity.

We first prove the binary cover case. Suppose
\(b,b_1,b_2\in\mathcal K\) satisfy
\(
\mathcal{A}(b)\subseteq \mathcal A(b_1)\cup \mathcal A(b_2).
\)
Let
\(
u:=\mathcal A^{-1}\!\left(\mathcal A(b_1)\cup \mathcal A(b_2)\right).
\)
Since \(\mathcal{A}(b)\subseteq \mathcal A(u)\), price monotonicity gives
\(
p_b\le p_u.
\)
Next define
\(
d:=\mathcal A^{-1}\!\left(\mathcal A(b_1)\setminus \mathcal A(b_2)\right).
\)
Then \(\mathcal A(d)\) and \(\mathcal A(b_2)\) form a pairwise disjoint partition of \(\mathcal A(u)\).
Therefore, by \(S_{P,2}\),
\(
p_u \le p_d+p_{b_2}.
\)
Because \(\mathcal A(d)\subseteq \mathcal A(b_1)\), price monotonicity implies
\(
p_d\le p_{b_1}.
\)
Combining the three inequalities gives
\(
p_b\le p_u\le p_d+p_{b_2}\le p_{b_1}+p_{b_2}.
\)
Thus the binary cover inequality holds.

The \(K\)-way cover case follows by induction on \(K\). For
\(K=2\), the claim is the binary cover case above. Suppose the claim holds for
\(K=k\), and consider a \((k+1)\)-way cover
\(
\mathcal{A}(b)\subseteq \bigcup_{r=1}^{k+1}\mathcal A(c_r).
\)
Let
\(
\bar c:=\mathcal A^{-1}\!\left(\bigcup_{r=1}^{k}\mathcal A(c_r)\right).
\)
Then \(\mathcal{A}(b)\subseteq \mathcal A(\bar c)\cup \mathcal A(c_{k+1})\), so the binary cover case gives
\(
p_b\le p_{\bar c}+p_{c_{k+1}}.
\)
By the induction hypothesis applied to the cover of \(\mathcal A(\bar c)\) by
\(\mathcal A(c_1),\ldots,\mathcal A(c_k)\),
\(
p_{\bar c}\le \sum_{r=1}^{k}p_{c_r}.
\)
Substitution yields
\(
p_b\le \sum_{r=1}^{k+1}p_{c_r}.
\)
Hence \(S_{C,K}\) holds for all \(K\ge2\).
\qed
\endproof

\begin{theorem}
Under Assumption~\ref{ass:price_monotonicity}, the two-way partition
subadditivity \(S_{P,2}\) is equivalent to the \(K\)-way cover
subadditivity \(S_{C,K}\) on the full bundle universe
\(\mathfrak F=2^{[n]}\).
\end{theorem}

\proof{Proof.}
By Proposition~\ref{prop:sp2_spk_eq}, \(S_{P,2}\iff S_{P,K}\) on the full bundle universe
\(\mathfrak F\). By Proposition~\ref{prop:subadd_eq}, under Assumption~\ref{ass:price_monotonicity},
\(S_{P,K}\iff S_{C,K}\). Therefore,
\(
S_{P,2}\iff S_{C,K}.
\)
\qed
\endproof

\section{Details of Numerical Experiments}
\subsection{Limitation of Other Policies}
\label{subsec:limitation_policies}
First, existing approximation algorithms such as CPBSD \citep{chen2025component} and PBDC \citep{ma2021reaping} are designed primarily under the strict assumption of additive valuations. Their formulations express the valuation of a bundle as a product-wise summation of individual item values. However, in a non-additive setting where substitution and complementarity effects (sub-additivity and super-additivity) are present, the valuation of a bundle is not equal to the simple sum of its constituent products. As a result, these additive formulations cannot be directly applied to the more complex non-additive bundle valuation structure considered in our setting.

Second, recent neural-network-based MILP approaches, such as Neural Branching \citep{gasse2019exact} or Neural Diving \citep{nair2020solving} algorithms, are computationally infeasible for this problem. These methods typically rely on encoding the MILP into a variable-constraint bipartite graph. However, since the standard formulation of mixed bundling involves an exponential number of decision variables ($2^n$ bundles), constructing and processing such a graph is almost impossible.

\subsection{Data Generation and GNN Training}
\label{app:main_deterministic_training}
\paragraph{Solution Sample Generation.} Let $\mathcal U[a, b]$ denote the uniform distribution over the interval $[a, b]$, and let $\mathcal U_D[N]$ denote the discrete uniform distribution over the set $[N]$. In the numerical experiments, given the number of products $n$, the number of customer types $m$, we generate each problem instance as follows:
\begin{itemize}
    \item \textbf{Customer Proportions.} We first draw $\beta_k$ from $\mathcal U[0,1]$ independently and normalize to obtain the customer proportions $\boldsymbol \alpha=(\alpha_1,\cdots,\alpha_m)$, where $\alpha_k=\beta_k / \sum_{k'=1}^m\beta_{k'}$.
    \item \textbf{Product Utilities and Costs.} For each product $j \in [n]$ and customer segment $k \in [m]$,  we draw utilities $u_{kj}$ from $\mathcal U[0,1]$, unit costs $c_j^u$ from $\mathcal U[0,0.2]$ and shipping costs $c_k^s$ from $\mathcal U[0,0.2]$. Then, the bundle costs and valuations can be computed by $c_{kb}= \left(\sum_{j \in \mathcal A(b)} c_j^u\right) + c_k^s$ and $R_{kb} = \sqrt{\sum_{j\in \mathcal{A}(b)} u_{kj}}$.
    \item \textbf{Label Generation.} Finally, for each generated instance, we compute the bundle assignment $\theta^{\star}_{kb}$'s using MB policy and convert them into the ground truth labels $q^{\star}_{kj}$'s. The resulting sample is characterized by $(\mathbf{c}^u, \mathbf{c}^s, \mathbf{U}, \boldsymbol{\alpha}, \sqrt{\cdot},\mathbf Q^{\star})$.
\end{itemize}

The dataset is then randomly divided into a training set (80\%) and a validation set (20\%). We use hidden dimension $d_{hid}=128$ and $\omega=4$ GNN layers. The model is optimized using AdamW \citep{loshchilov2018decoupled} with
learning rate \(10^{-3}\), weight decay \(10^{-4}\), dropout rate \(0.2\), and
batch size \(256\). We use a ReduceLROnPlateau learning-rate scheduler, which
reduces the learning rate when the validation loss stops improving, and apply
gradient clipping with maximum norm \(1.0\) to stabilize training. Training is capped at 200 epochs per trial, incorporating an early-stopping criterion (patience=50) that halts training if the reduction in validation loss remains less than $10^{-12}$ for 50 consecutive epochs. The total training time is approximately $264$ seconds.

\section{Implementation of Cover-Form Price Subadditivity}
\label{sec:cover_subadditivity_implementation}

For a restricted candidate bundle family \(\mathfrak B\subseteq\mathfrak F\),
we impose price subadditivity in the cover form. Let
\(
\mathcal I_{\mathfrak B}
:=
\{b\in\mathcal K:\mathcal A(b)\in\mathfrak B\}.
\)
For a target bundle index \(b\in\mathcal I_{\mathfrak B}\), a collection
\(\mathcal C\subseteq\mathcal I_{\mathfrak B}\setminus\{b\}\) is called a
cover of \(b\) if
\(
\mathcal A(b)\subseteq\bigcup_{c\in\mathcal C}\mathcal A(c).
\)
The corresponding cover-subadditivity inequality is
\(
p_b\le\sum_{c\in\mathcal C}p_c.
\)
Since prices are nonnegative, it is sufficient to add minimal covers: if
\(\mathcal C'\subsetneq\mathcal C\) also covers \(b\), then the inequality
generated by \(\mathcal C\) is dominated by that generated by \(\mathcal C'\).

\paragraph{FCP.}
Under FCP, the candidate family contains at most one predicted bundle for each
segment, plus the empty bundle. Therefore the number of retained bundles is at
most \(m+1\), and all minimal cover-subadditivity inequalities can be
pre-enumerated. For each target bundle index, we first add all singleton cover
inequalities
\[
p_b\le p_c,
\qquad
\forall b,c\in\mathcal I_{\mathfrak B}
\text{ with }
\mathcal A(b)\subseteq\mathcal A(c).
\]
We then enumerate multi-bundle minimal covers by a depth-first search over
candidate bundles that have nonempty intersection with \(\mathcal{A}(b)\). During the
search, we keep the currently covered product set and branch on an uncovered
product that has the fewest remaining candidate bundles covering it. When a
cover is found, we add it only if it is minimal, i.e., no selected bundle is
redundant in the cover.
\begin{breakablealgorithm}
\caption{Pre-enumeration of Minimal Cover Cuts for FCP}
\label{alg:fcp_cover_pre_enum}
\begin{algorithmic}
\Statex \textbf{Input:} Candidate index set \(\mathcal I_{\mathfrak B}\) generated by FCP.
\For{\(b\in\mathcal I_{\mathfrak B}\setminus\{0\}\)}
    \For{\(c\in\mathcal I_{\mathfrak B}\setminus\{0,b\}\)}
        \If{\(\mathcal{A}(b)\subseteq \mathcal A(c)\)}
            \State add \(p_b\le p_c\).
        \EndIf
    \EndFor
    \State \(\mathcal C_b\leftarrow
\{c\in\mathcal I_{\mathfrak B}\setminus\{b\}:
\mathcal A(c)\cap\mathcal A(b)\neq\emptyset,\
\mathcal A(b)\nsubseteq\mathcal A(c)\}\).
\State Use DFS over \(\mathcal C_b\) to enumerate minimal collections
\(\mathcal C\) satisfying
\(\mathcal A(b)\subseteq\bigcup_{c\in\mathcal C}\mathcal A(c)\).
\For{each minimal cover \(\mathcal C\)}
    \State add \(p_b\le\sum_{c\in\mathcal C}p_c\).
\EndFor
\EndFor
\end{algorithmic}
\end{breakablealgorithm}

\paragraph{PCP.}
Under PCP, each segment \(k\) generates a nested prefix chain
\(
B^{\mathrm{PCP}}_{k,1}
\subseteq
B^{\mathrm{PCP}}_{k,2}
\subseteq
\cdots
\subseteq
B^{\mathrm{PCP}}_{k,|U_k|}.
\)
We add the within-chain monotonicity constraints upfront:
\[
p_{b^{\mathrm{PCP}}_{k,i}}
\le
p_{b^{\mathrm{PCP}}_{k,i+1}},
\qquad
\forall k\in[m],\ i=1,\ldots,|U_k|-1.
\]
We also add singleton cover inequalities upfront:
\[
p_b\le p_c,
\qquad
\forall b,c\in\mathcal I_{\mathfrak B}
\text{ with }
\mathcal A(b)\subseteq\mathcal A(c).
\]
The remaining multi-bundle cover inequalities are generated by a lazy-cut
separation procedure. Here, ``separation'' means checking whether the current
incumbent price vector \(\bar{\mathbf p}\) violates any cover-subadditivity constraint,
and, if so, returning one violated inequality as a lazy cut.

At an incumbent solution \(\bar{\mathbf p}\), for each target bundle index
\(b\), we solve the cheapest-cover separation problem
\[
\zeta_b(\bar{\mathbf p})
:=
\min_{\mathcal C\subseteq\mathcal I_{\mathfrak B}}
\left\{
\sum_{c\in\mathcal C}\bar p_c:
\mathcal A(b)\subseteq\bigcup_{c\in\mathcal C}\mathcal A(c),\
|\mathcal C|\ge2
\right\}.
\]
If \(\zeta_b(\bar{\mathbf p})<\bar p_b-\varepsilon\), then the cover
\(\mathcal C_b^\star\) attaining \(\zeta_b(\bar{\mathbf p})\) yields the
violated inequality
\(
p_b\le\sum_{c\in\mathcal C_b^\star}p_c,
\)
which is added as a lazy constraint.

The nested prefix-chain structure allows a more efficient separation routine.
Because prices are nondecreasing along each prefix chain, an optimal cheapest
cover never needs to use two bundles from the same chain: if two selected
bundles come from the same chain, the larger one covers everything the smaller
one covers, and dropping the smaller one weakly decreases the cover cost.
Therefore, the separation problem can be solved by a sparse dynamic program
over segment chains.

For each segment
\(r\), let \(\mathcal H_r\) denote its prefix index chain. Define the set of
usable options
\[
\mathcal O_r(b)
:=
\{(\mathcal A(c)\cap \mathcal A(b),c): c\in\mathcal H_r\setminus\{b\},\
\mathcal A(c)\cap \mathcal A(b)\neq\emptyset\}.
\]
The dynamic program stores the minimum cover cost for each covered subset
\(M\subseteq \mathcal A(b)\). Initialize \(D_0(\emptyset)=0\) and
\(D_0(M)=+\infty\) for \(M\neq\emptyset\). For each chain \(r\), update
\[
D_r(M)
=
\min\left\{
D_{r-1}(M),\
\min_{\substack{M'\subseteq \mathcal A(b),\ (o,c)\in\mathcal O_r(b):\\
M=M'\cup o}}
D_{r-1}(M')+\bar p_c
\right\}.
\]
After all chains are processed, \(D_R(\mathcal A(b))\) is the cheapest multi-chain cover
cost. If \(D_R(\mathcal A(b))<\bar p_b-\varepsilon\), the corresponding cover is added as
a lazy cut.

\begin{breakablealgorithm}
\caption{Lazy Separation of Cover-Subadditivity Cuts for PCP}
\label{alg:pcp_lazy_cover}
\begin{algorithmic}
\Statex \textbf{Input:} Candidate bundle index set
\(\mathcal I_{\mathfrak B}\), prefix index chains
\(\{\mathcal H_k\}_{k\in[m]}\), incumbent prices
\(\bar{\mathbf p}\), tolerance \(\varepsilon\).
\For{each target bundle \(b\in\mathcal I_{\mathfrak B}\setminus\{0\}\), in descending order of \(\bar p_b\)}
    \State Solve the cheapest-cover problem
    \[
    \zeta_b(\bar{\mathbf p})=
\min\left\{
\sum_{c\in\mathcal C}\bar p_c:
\mathcal{A}(b)\subseteq\bigcup_{c\in\mathcal C}\mathcal A(c),\
|\mathcal C|\ge2
\right\}.
    \]
    by the segment-chain dynamic program.
    \If{\(\zeta_b(\bar{\mathbf p})<\bar p_b-\varepsilon\)}
        \State Let \(\mathcal C_b^\star\) be the cheapest violated cover.
        \State Add the lazy cut \(p_b\le\sum_{c\in \mathcal C_b^\star}p_c\).
        \State \textbf{break} \Comment{one violated lazy cut is enough to reject the incumbent}
    \EndIf
\EndFor
\end{algorithmic}
\end{breakablealgorithm}

\section{Additional Experimental Details}
\subsection{Seed Performance}
\label{subsec:seed_performance}
In this subsection, we present the results under 10 independent trials as the random seed ranging from 1 to 10.
\begin{table}[h]
  \centering
  \caption{Performance metrics across different random seeds for
  \(m_{\mathrm{test}}=10, n_{\mathrm{test}}=10\).}
  \label{tab:seed_performance}
  \begin{tabular}{cccccc}
    \toprule
    Seed & Training Loss & Validation Loss
    & \(PR_{\mathrm{FCP},\mathrm{MB}}\) (std)
    & \(TR_{\mathrm{FCP},\mathrm{MB}}\) & Time (s) \\
    \midrule
    1  & 0.0779 & 0.0938 & 0.989 (0.008) & 0.0038 & 0.0210 \\
    2  & 0.0803 & 0.0968 & 0.988 (0.008) & 0.0037 & 0.0203 \\
    3  & 0.0786 & 0.0974 & 0.988 (0.009) & 0.0037 & 0.0206 \\
    4  & 0.0747 & 0.0910 & 0.989 (0.009) & 0.0037 & 0.0205 \\
    5  & 0.0760 & 0.0919 & 0.989 (0.009) & 0.0037 & 0.0202 \\
    6  & 0.0730 & 0.0917 & 0.989 (0.009) & 0.0036 & 0.0203 \\
    7  & 0.0682 & 0.0887 & 0.989 (0.008) & 0.0038 & 0.0209 \\
    8  & 0.0755 & 0.0895 & 0.989 (0.008) & 0.0037 & 0.0207 \\
    9  & 0.0815 & 0.0994 & 0.988 (0.010) & 0.0037 & 0.0205 \\
    10 & 0.0800 & 0.0976 & 0.989 (0.008) & 0.0038 & 0.0211 \\
    \bottomrule
  \end{tabular}
\end{table}

\subsection{Experiments under Random Valuation}
\label{app:cpbsd_additive_details}

\subsubsection{Experimental setup and baselines.}
\label{app:cpbsd_additive_setup_baselines}

In this section, we evaluate the proposed framework under additive random
valuations using two product scales. We first consider instances with
\(N=10\) products and \(K_\mathrm{in}=50\) sampled customers, and then repeat the
experiment with \(N=30\) products and \(K_\mathrm{in}=50\) sampled customers. For each
scale, we test three cost settings: zero costs, independent random costs
(\texttt{random\_ind}), and costs correlated with valuation means
(\texttt{random\_corr}). For each SAA instance, the solver observes one
realization of \(K_\mathrm{in}\) customer valuation vectors. The resulting price vector is
then evaluated on \(5000\) independent out-of-sample valuation draws.

We compare two baselines. The first is bundle-size pricing (BSP), following
\citet{chu2011bundle} and the SAA approach. The second is CPBSD-A, a structured additive
approximation of component pricing with bundle-size discounts
\citep{chen2025component}. CPBSD-A retains component prices and bundle-size
discounts, while using preset segment-specific product rankings to reduce the
optimization size.

The GNN used in this section is trained separately on additive random-valuation
instances. In particular, it is trained once on small instances with
\(N=5\), \(K_\mathrm{in}=50\), normally distributed valuations, correlation parameter
\(\rho=0\), full customer heterogeneity, and the \texttt{random\_ind} cost
setting. The trained GNN model is then used for all \(N=10\) and
\(N=30\) evaluation instances, without retraining. All three policies are solved with MIP gap \(10^{-3}\) and a common time limit of 300 seconds.

The FCP evaluation procedure under the SAA setting is summarized in Algorithm~\ref{alg:saa_fcp_random_valuation}.

\begin{breakablealgorithm}
\caption{FCP for Additive Random-Valuation Bundle Pricing}
\renewcommand{\algorithmicrequire}{\textbf{Input:}}
\renewcommand{\algorithmicensure}{\textbf{Output:}}
\label{alg:saa_fcp_random_valuation}
\begin{algorithmic}
\Require Valuation distribution \(\mathcal V\), product number \(N\), in-sample size \(K_{\mathrm{in}}\), out-of-sample size \(K_{\mathrm{oos}}\), product costs, trained GNN.
\Ensure In-sample SAA objective, out-of-sample profit, and runtime.

\State Draw in-sample valuations
\(\{\mathbf v_i^{\mathrm{in}}\}_{i=1}^{K_{\mathrm{in}}}\sim \mathcal V\).
\State Construct the SAA instance \(\widehat X_{\mathrm{in}}\) by treating each
sampled customer \(i\) as a segment with weight \(1/K_{\mathrm{in}}\).
\State Apply FCP with the trained GNN to \(\widehat X_{\mathrm{in}}\), and
obtain a restricted candidate family \(\mathfrak B^{\mathrm{FCP}}\).
\State Define the
associated index set
\(
\mathcal I_{\mathfrak B^{\mathrm{FCP}}}
:=
\{b\in\mathcal K:\mathcal A(b)\in\mathfrak B^{\mathrm{FCP}}\}.
\)
\State For every \(b\in\mathcal I_{\mathfrak B^{\mathrm{FCP}}}\), define
\(c_b:=\sum_{j\in\mathcal A(b)}c_j\), with \(c_0=0\).
\State Solve \(\Xi(\mathfrak B^{\mathrm{FCP}})\) for \(\widehat X_{\mathrm{in}}\), and let
\(\widehat{\mathbf p}=(\widehat p_b)_{b\in\mathcal I_{\mathfrak B^{\mathrm{FCP}}}}\)
be the resulting price vector.
\State Record the in-sample objective value \(\widehat z_{\mathrm{in}}\).

\State Draw independent out-of-sample valuations
\(\{\mathbf v_t^{\mathrm{oos}}\}_{t=1}^{K_{\mathrm{oos}}}\sim \mathcal V\).
\For{\(t=1,\ldots,K_{\mathrm{oos}}\)}
    \State Determine the chosen bundle index
    \(
    \widehat b_t
    \in
    \arg\max_{b\in\mathcal I_{\mathfrak B^{\mathrm{FCP}}}}
    \left\{
    \sum_{j\in\mathcal A(b)}v_{tj}^{\mathrm{oos}}-\widehat p_b
    \right\}.
    \)
\EndFor
\State Compute
\(
z_{\mathrm{oos}}(\widehat{\mathbf p})
:=
\frac{1}{K_{\mathrm{oos}}}
\sum_{t=1}^{K_{\mathrm{oos}}}
\left(\widehat p_{\widehat b_t}-c_{\widehat b_t}\right).
\)
\State \Return \(\widehat z_{\mathrm{in}}\), \(z_{\mathrm{oos}}(\widehat{\mathbf p})\), and runtime.
\end{algorithmic}
\end{breakablealgorithm}

\subsubsection{Additive CPBSD formulation.}
\label{app:cpbsd_additive_formulation}

We report technical details for the CPBSD algorithm in Section~\ref{sec:external_validation_random_valuation}. Let \(\mathcal{N}=[N]\) denote the product set. For customer \(k\), product-level valuations are denoted by \(v_{kn}\), and the valuation of a bundle \(S\subseteq \mathcal{N}\) is additive:
\(
V_k(S)=\sum_{n\in S} v_{kn}.
\)

Following the CPBSD pricing structure \citep{chen2025component}, a policy assigns component prices \(p_n\) and bundle-size discounts \(d_s\) for \(s\in[N]\). If the customer buys bundle \(S\) with \(|S|=s\), the payment is
\(
P(S)=\sum_{n\in S}p_n - s d_s,
\)
and the realized profit from this customer is
\(
\Pi(S)=\sum_{n\in S}(p_n-c_n)-s d_s.
\)
The customer chooses a bundle maximizing additive utility net of the CPBSD payment, with the outside option normalized to zero:
\[
S_k \in \arg\max_{S\subseteq \mathcal{N}}\left\{\sum_{n\in S}v_{kn}-\sum_{n\in S}p_n+|S|d_{|S|},\,0\right\}.
\]

\subsubsection{CPBSD-A approximation formulation.}
\label{app:cpbsd_a_formulation}

CPBSD-A approximates the sample CPBSD problem by fixing a segment-specific product ranking. In our implementation, customer \(k\) ranks products by descending potential surplus \(v_{kn}-c_n\). Let \(\pi_k=(\pi_{k1},\ldots,\pi_{kN})\) denote this ranking. If customer \(k\) buys a bundle of size \(s\), CPBSD-A restricts the bundle composition to the top-\(s\) products under this ranking. Define the corresponding prefix valuation and cost as
\(
v_{ks}=\sum_{j=1}^{s}v_{k\pi_{kj}},
\quad
c_{ks}=\sum_{j=1}^{s}c_{\pi_{kj}}.
\)

The MILP uses component prices \(p_n\), size discounts \(d_s\), binary size-choice variables \(y_{ks}\), induced payments \(p_{ks}\), realized payments \(q_{ks}\), selected surplus variables \(w_{ks}\), and customer surplus variables \(w_k\). With a big-\(M\) constant large enough to upper-bound feasible bundle payments, the implemented CPBSD-A model is
\begin{align}
\max \quad
& \frac{1}{K_\mathrm{in}}
\sum_{k\in[K_\mathrm{in}]}\sum_{s\in[N]}
\left(q_{ks}-c_{ks}y_{ks}\right)
\label{eq:cpbsd_a_obj}
\\
\text{s.t.}\quad
& p_{ks}=\sum_{j=1}^{s}p_{\pi_{kj}}-s d_s,
&& \forall k\in[K_\mathrm{in}],\ s\in[N]
\label{eq:cpbsd_a_price}
\\
& v_{k\pi_{kj}}-p_{\pi_{kj}}
\ge
v_{k\pi_{k,j+1}}-p_{\pi_{k,j+1}},
&& \forall k\in[K_\mathrm{in}],\ j\in[N-1]
\label{eq:cpbsd_a_rank}
\\
& w_k \ge v_{ks}-p_{ks},
&& \forall k\in[K_\mathrm{in}],\ s\in[N]
\label{eq:cpbsd_a_surplus_ub}
\\
& \sum_{s\in[N]}y_{ks}\le 1,
&& \forall k\in[K_\mathrm{in}]
\label{eq:cpbsd_a_one_size}
\\
& q_{ks}\ge p_{ks}-M(1-y_{ks}),
&& \forall k\in[K_\mathrm{in}],\ s\in[N]
\label{eq:cpbsd_a_q_lb}
\\
& q_{ks}\le p_{ks},
&& \forall k\in[K_\mathrm{in}],\ s\in[N]
\label{eq:cpbsd_a_q_ub}
\\
& w_{ks}=v_{ks}y_{ks}-q_{ks},
&& \forall k\in[K_\mathrm{in}],\ s\in[N]
\label{eq:cpbsd_a_wks}
\\
& w_k=\sum_{s\in[N]}w_{ks},
&& \forall k\in[K_\mathrm{in}]
\label{eq:cpbsd_a_wk}
\\
& s d_s \ge s_1d_{s_1}+(s-s_1)d_{s-s_1},
&& \forall s\in[N],\ s_1\in[s-1]
\label{eq:cpbsd_a_subadd}
\\
& d_1=0,\quad p_n,p_{ks},q_{ks},w_{ks},w_k,d_s\ge0,\quad y_{ks}\in\{0,1\}.
\end{align}

Constraint~\eqref{eq:cpbsd_a_rank} preserves the preset product order after component prices are chosen, so that the top-\(s\) prefix remains consistent with the within-size choice implied by CPBSD. Thus, CPBSD-A is a structured additive heuristic baseline rather than an exact mixed bundling formulation.

\subsubsection{Data generation and cost setups.}
\label{app:cpbsd_additive_data_generation}

Both product-scale blocks use \(K_\mathrm{in}=50\), the normal valuation family, independent product-level valuation shocks, and full product heterogeneity. With the implementation's zero-based product index \(i=0,\ldots,N-1\), valuation means are
\(
\mu_i = 1 + \frac{9i}{N-1}.
\)
The Gaussian copula uses a Toeplitz correlation matrix
\(
\mathrm{Corr}(i,j)=\rho^{|i-j|}.
\)
Since \(\rho=0.0\) in the reported experiments, product-level valuation shocks are independent in the copula layer. For the normal valuation family, if \(u_{ki}\) is the copula draw and \(\Phi^{-1}(\cdot)\) is the standard normal inverse CDF, the valuation sample is generated as
\(
v_{ki}=\max\left\{\mu_i+0.5\Phi^{-1}(u_{ki}),0\right\}.
\)
The three production-cost setups are
\begin{align*}
\texttt{zero}:\quad & c_i = 0, \quad i=0,\ldots,N-1, \\
\texttt{random\_ind}:\quad & c_i \sim \mathrm{Uniform}(0,10) \quad \text{independently across products}, \\
\texttt{random\_corr}:\quad & c_i = \max\{\mu_i r_i + \epsilon_i,0\}, \quad r_i\sim \mathrm{Uniform}(0,1),\quad \epsilon_i\sim \mathcal{N}(0,0.5^2).
\end{align*}
The label \texttt{random\_corr} refers to production costs that are noisy functions of valuation means; it does not denote correlated customer valuation shocks.

\subsubsection{GNN Training under Additive Random Valuations}
\label{subsubsec:random_valuation_gnn_training}

The GNN used in the additive random-valuation experiments is trained
separately from the GNN used in the main deterministic experiments. Each
training instance is generated by first drawing a finite sample of customer
valuations from the underlying valuation distribution. Specifically, for
training instance \(\ell\), we draw
\(
\{\mathbf v_i^{(\ell)}\}_{i=1}^{K_{\mathrm{train}}}\sim \mathcal V,
\)
and treat the sampled customers as a deterministic SAA instance with equal
weights \(1/K_{\mathrm{train}}\).

We then solve the full mixed bundling formulation \(\Xi(\mathfrak F)\) on this
finite SAA instance and obtain an optimal assignment matrix
\(\boldsymbol{\Theta}^{\star,(\ell)}\). Let \(\mathbf M_{\mathfrak F}\in\{0,1\}^{|\mathfrak F|\times N}\)
denote the bundle-product incidence matrix, with
\(
(\mathbf M_{\mathfrak F})_{bj}=\mathbbm{1}\{j\in \mathcal{A}(b)\}.
\)
As in the deterministic setting, the edge-level label is the
product-assignment projection
\[
\mathbf Q^{\star,(\ell)}
=
\boldsymbol{\Theta}^{\star,(\ell)}\mathbf M_{\mathfrak F},
\qquad
q_{ij}^{\star,(\ell)}
=
\sum_{b:j\in \mathcal{A}(b)}\theta_{ib}^{\star,(\ell)}.
\]

The graph construction follows the same product-node and edge-feature design
as in Section~\ref{subsec:GraphRepresentation}. Since segment sizes and
shipping costs do not exist in this setting, the two segment-feature slots are
populated with the sample size \(K_{\mathrm{train}}\) and the customer's
positive-margin rate
$\rho_i
=
\frac{1}{N}\sum_{j=1}^{N}\mathbbm{1}\{v_{ij}>c_j\}.$
Thus, for sampled customer \(i\), the customer feature vector is
$
\mathbf y_i^S
=
\bigl[0,\;0,\;K_{\mathrm{train}},\;\rho_i\bigr].
$
Apart from this substitution in the segment-feature slots, the supervised
training procedure follows Section~\ref{subsec:GNN_training}: the model is
trained to predict the edge-level labels \(\mathbf Q^{\star,(\ell)}\) using weighted binary
cross-entropy, with the same training-validation split, optimizer, model-selection
rule, and early-stopping criterion.

\section{Cutoff Sensitivity Analysis}
\label{app:cutoff_sensitivity}

We examine the sensitivity of the cutoff-dependent pruning policies to the threshold
used to filter the GNN-predicted segment-product probabilities. For a given cutoff
\(\tau\), FCP constructs one segment-specific candidate bundle
$B^{\mathrm{FCP}}_k(\tau)=\{j\in[n]:\mathbf{P}_{kj}\ge \tau\}$,
with the same fallback rule as Algorithm~\ref{alg:fcp} when the thresholded set is empty.
The global candidate family is then obtained by taking the union of these segment-level
bundles. Hence, for FCP, the cutoff mainly controls the sparsity and composition of the
candidate bundles, while the candidate-family size remains at most \(m+1\), including
the outside option. By contrast, PCP first filters products by the same cutoff and then
constructs a ranked prefix chain from the retained products. Therefore, the cutoff also
directly affects the number of prefix bundles generated by PCP.

We vary \(\tau\in\{0.1,0.2,\ldots,0.9\}\) and evaluate FCP and PCP under identical
experimental settings. We report the profit ratio, time ratio, product-level prediction
accuracy, and product-level recall averaged over multiple instances. The prediction
diagnostics are computed by comparing the thresholded product-selection mask
$\widehat q_{kj}(\tau)=\mathbbm{1}\{\mathbf{P}_{kj}\ge \tau\}$
with the optimal product-assignment label \(q^\star_{kj}\) defined in the training process.
For PCP, these accuracy and recall statistics evaluate the cutoff filter before the
prefix-bundle expansion, rather than the final MILP assignment.

\begin{figure}[h]
\centering
\includegraphics[width=0.75\linewidth]{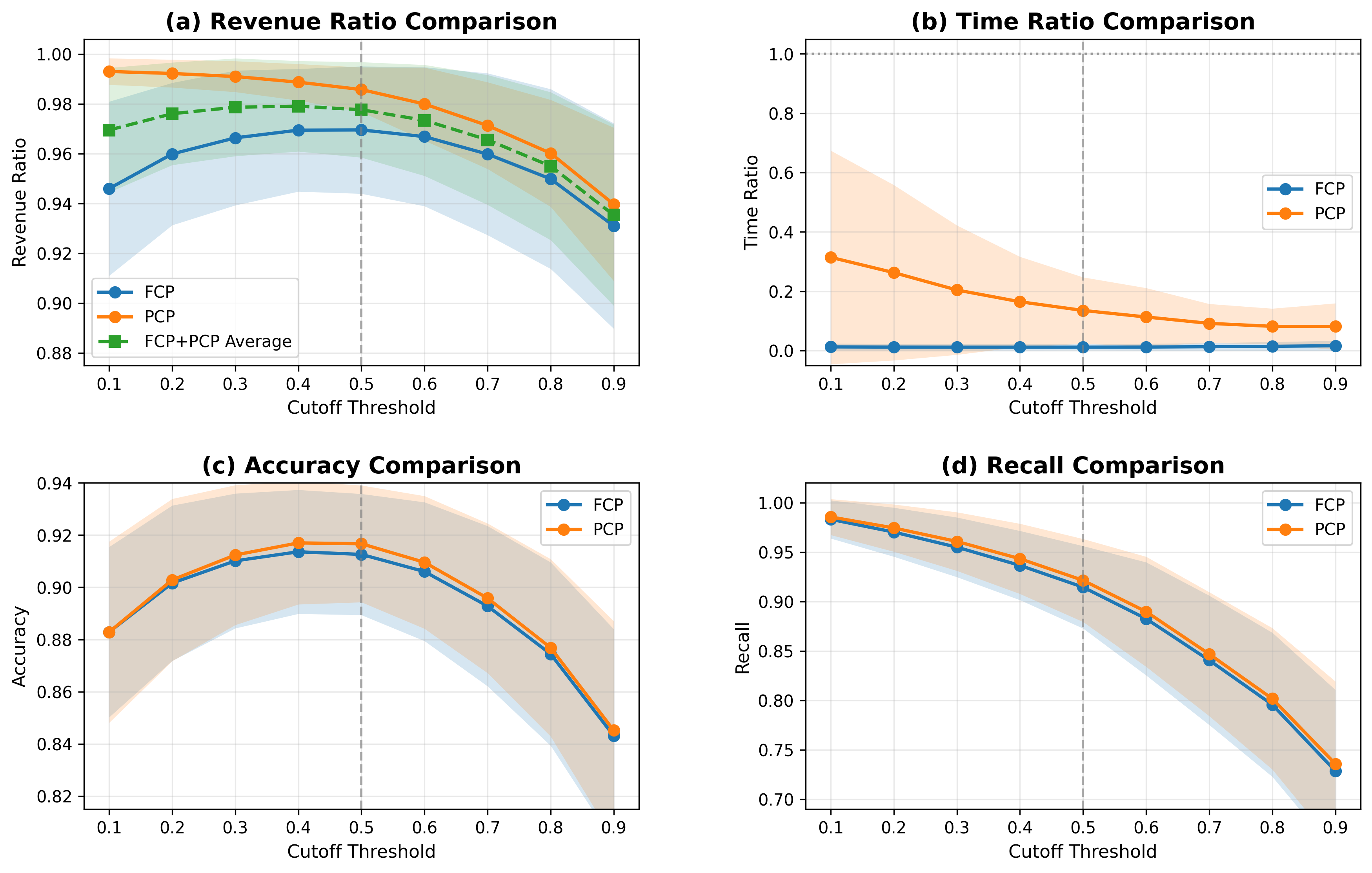}
\caption{Sensitivity of profit ratio, time ratio, product-level prediction accuracy, and product-level recall to the cutoff threshold for FCP and PCP.}
\label{fig:cutoff_sensitivity}
\end{figure}

\paragraph{Profit--Time Trade-off.}
As shown in Figure~\ref{fig:cutoff_sensitivity}, FCP achieves its strongest profit
ratio at moderate cutoff levels. Very low cutoffs may include many low-confidence products
and produce overly large or noisy segment-specific bundles, while overly high cutoffs may
remove products that appear in the optimal purchased bundles. PCP remains relatively
stable over a broad range of moderate thresholds because the prefix-chain construction
retains a richer candidate family and can absorb moderate prediction noise. However,
its performance also deteriorates when the cutoff becomes too aggressive. The runtime
patterns are consistent with the pruning mechanisms: FCP remains very fast and only
weakly sensitive to the cutoff because its candidate-family size remains \(O(m)\), whereas
PCP becomes faster as the cutoff increases because fewer products survive the filter
and fewer prefix bundles are generated.

\paragraph{Accuracy--Recall Trade-off.}
The cutoff threshold also induces a product-level accuracy--recall trade-off. As the cutoff
increases from low to moderate values, prediction accuracy improves because more
low-confidence false positives are removed. Beyond a certain point, however, false negatives
become dominant and accuracy begins to decline. Recall decreases steadily with the cutoff,
reflecting the increasing risk of excluding products that belong to the optimal purchased
bundle. Since false negatives are particularly harmful in bundle pricing, \(\tau=0.5\) provides
a conservative default: it lies close to the peak-accuracy region while preserving a reasonable
level of recall and candidate-bundle coverage.

\section{Sensitivity Analysis of Candidate Size \(K\)}
\label{app:k_justification}

To justify the default candidate budget \(K=\lceil\sqrt m\rceil\) in the Global Top-\(K\)
local-search policy, we conduct a controlled sensitivity analysis over the number of local
moves evaluated at each FCPLS iteration. In Algorithm~\ref{alg:global_ls}, \(K\) is a
per-direction candidate budget: the algorithm independently selects the top \(K\) Add
moves and the top \(K\) Drop moves according to the GNN-guided confidence score.
Thus, each iteration evaluates at most \(2K\) candidate moves after pooling and sorting.

For this diagnostic experiment, we compare the following candidate-budget rules:
\begin{itemize}
    \item \textbf{Full segment-wise neighborhood \((2m)\):} evaluate the best Add move and the best Drop move for each customer segment. This is an exhaustive one-step segment-wise benchmark, not an enumeration of the full bundle space.
    \item \textbf{Constant-\(K\):} use a fixed per-direction candidate budget \(K\in\{5,10\}\), independent of the number of customer segments.
    \item \textbf{\(K=m\):} use a linear per-direction candidate budget, evaluating at most \(2m\) globally ranked Add/Drop moves per iteration.
    \item \textbf{Sqrt-\(m\) (Proposed):} use the adaptive sublinear budget \(K=\lceil\sqrt m\rceil\).
    \item \textbf{\(K=2\sqrt m\):} use a moderately larger sublinear budget \(K=\lceil2\sqrt m\rceil\).
    \item \textbf{Sqrt-\(mn\):} use a joint segment-product scaling rule \(K=\lceil\sqrt{mn}\rceil\).
\end{itemize}

\begin{figure}[h]
    \centering
    \includegraphics[width=0.75\linewidth]{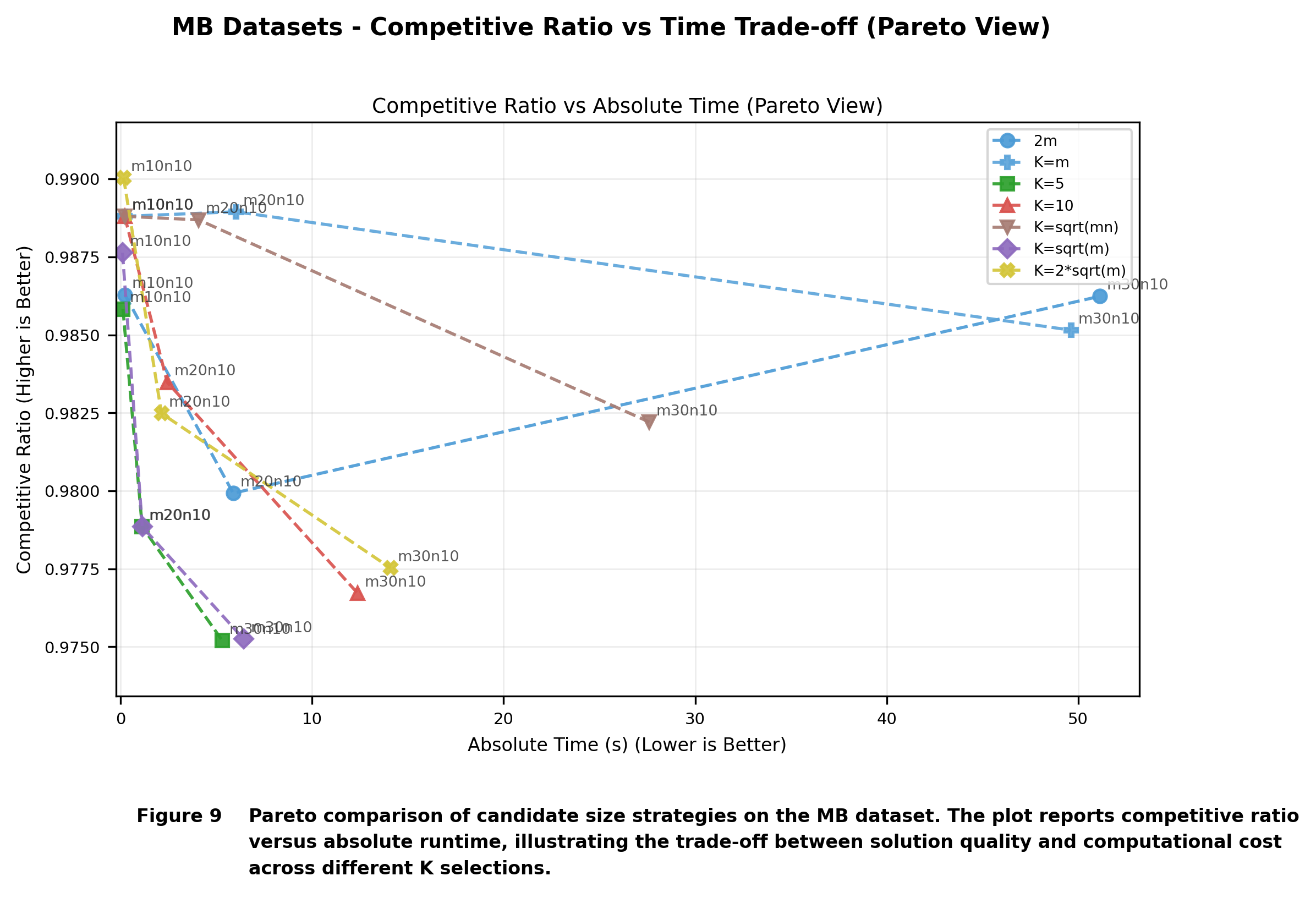}
    \caption{
    Pareto comparison of candidate-size rules on the MB dataset. The plot reports profit ratio versus absolute runtime, illustrating the trade-off between solution quality and computational cost across different \(K\) selections.
    }
    \label{fig:k_justify_relative}
\end{figure}

Figure~\ref{fig:k_justify_relative} reports the Pareto comparison between profit
ratio and absolute runtime across these candidate-size rules on the MB benchmark
instances with \(m\in\{10,20,30\}\). The Sqrt-\(m\) rule lies on or near the empirical
Pareto-efficient region across the tested settings. These results indicate that
\(K=\lceil\sqrt m\rceil\) provides a favorable balance between solution quality and
computational cost. Constant-\(K\) rules can be too restrictive when the number of
segments grows, while linear or larger scaling rules evaluate more neighbors and may
increase runtime without commensurate profit gains. The proposed Sqrt-\(m\) rule expands
the local-search neighborhood in a controlled sublinear manner, enabling richer exploration
as \(m\) increases while keeping the LP-evaluation workload computationally manageable.

\section{Consistency Analysis of LP and MILP Improvements}
\label{app:lp_consistency_analysis}

We further examine whether the fixed-assignment LP used in FCPLS provides reliable
guidance for improving the corresponding restricted mixed bundling MILP. The LP
subproblem used in local search is not a standard continuous relaxation of the full
mixed bundling MILP. Instead, as defined in Appendix~\ref{app:lp-ic}, it fixes the
segment-wise bundle assignment and optimizes only the feasible prices and surpluses over
the active candidate family. Therefore, its optimal value is a feasible-assignment lower
bound for the restricted mixed bundling formulation over the same active bundle family.

For a product-assignment matrix \(\mathbf Q\in\{0,1\}^{m\times n}\), let
$\mathfrak B^{\mathrm{LP}}(\mathbf Q)
=
\bigl\{\{j\in[n]:q_{kj}=1\}:k\in[m]\bigr\}\cup\{\emptyset\}$
be the active candidate bundle family induced by \(\mathbf Q\). We denote by \(\Xi_{\mathrm{LP}}(\mathbf Q)\)
the optimal value of the fixed-assignment LP evaluation used by FCPLS, and by
$\Xi_{\mathrm{MILP}}(\mathbf Q)
=
\Xi(\mathfrak B^{\mathrm{LP}}(\mathbf Q))$
the exact restricted MILP objective obtained by solving the mixed bundling formulation
over the same active candidate family. The purpose of this appendix is to empirically test
whether local moves that improve \(\Xi_{\mathrm{LP}}\) also tend to improve
\(\Xi_{\mathrm{MILP}}\).

\subsection{Experimental Setup}

We selected 60 instances across varying problem sizes, with the number of customer
segments ranging from \(m=10\) to \(m=30\) and with varying numbers of products \(n\).
For each instance, we executed the proposed FCPLS local-search algorithm. During the
tracking experiment, whenever a local move \(Y\) from the current solution \(X\) produced
a positive LP improvement,
$\Delta \Xi_{\mathrm{LP}}
=
\Xi_{\mathrm{LP}}(Y)-\Xi_{\mathrm{LP}}(X)>\epsilon,
\quad
\epsilon=10^{-6},$
we additionally solved the corresponding restricted MILP over
\(\mathfrak B^{\mathrm{LP}}(Y)\) to compute
$\Delta \Xi_{\mathrm{MILP}}
=
\Xi_{\mathrm{MILP}}(Y)-\Xi_{\mathrm{MILP}}(X).$
This tracking step is diagnostic: it is used to verify the quality of LP-guided local moves
and does not change the FCPLS search rule.

\subsection{Quantitative Analysis of Improvement Translation}

We define an \textbf{``Effective Translation''} as an LP-accepted local move that also
yields a strict improvement in the corresponding restricted MILP objective:
$\Delta \Xi_{\mathrm{LP}}>\epsilon
\quad\text{and}\quad
\Delta \Xi_{\mathrm{MILP}}>0.$
Across all tracked LP-accepted moves, 489 out of 522 led to immediate restricted-MILP
improvements, yielding a Global Translation Rate of \(93.68\%\). This indicates that the
fixed-assignment LP lower bound is a reliable, though not exact, empirical proxy for
ranking local-search moves in the tested neighborhood structure. The result also clarifies
the limitation of the LP guide: a positive LP improvement is not a formal guarantee of a
positive MILP improvement, but it translates successfully in the large majority of observed
moves.

\subsection{Visual Trajectory}

\begin{figure}[h]
    \centering
    \includegraphics[width=0.75\linewidth]{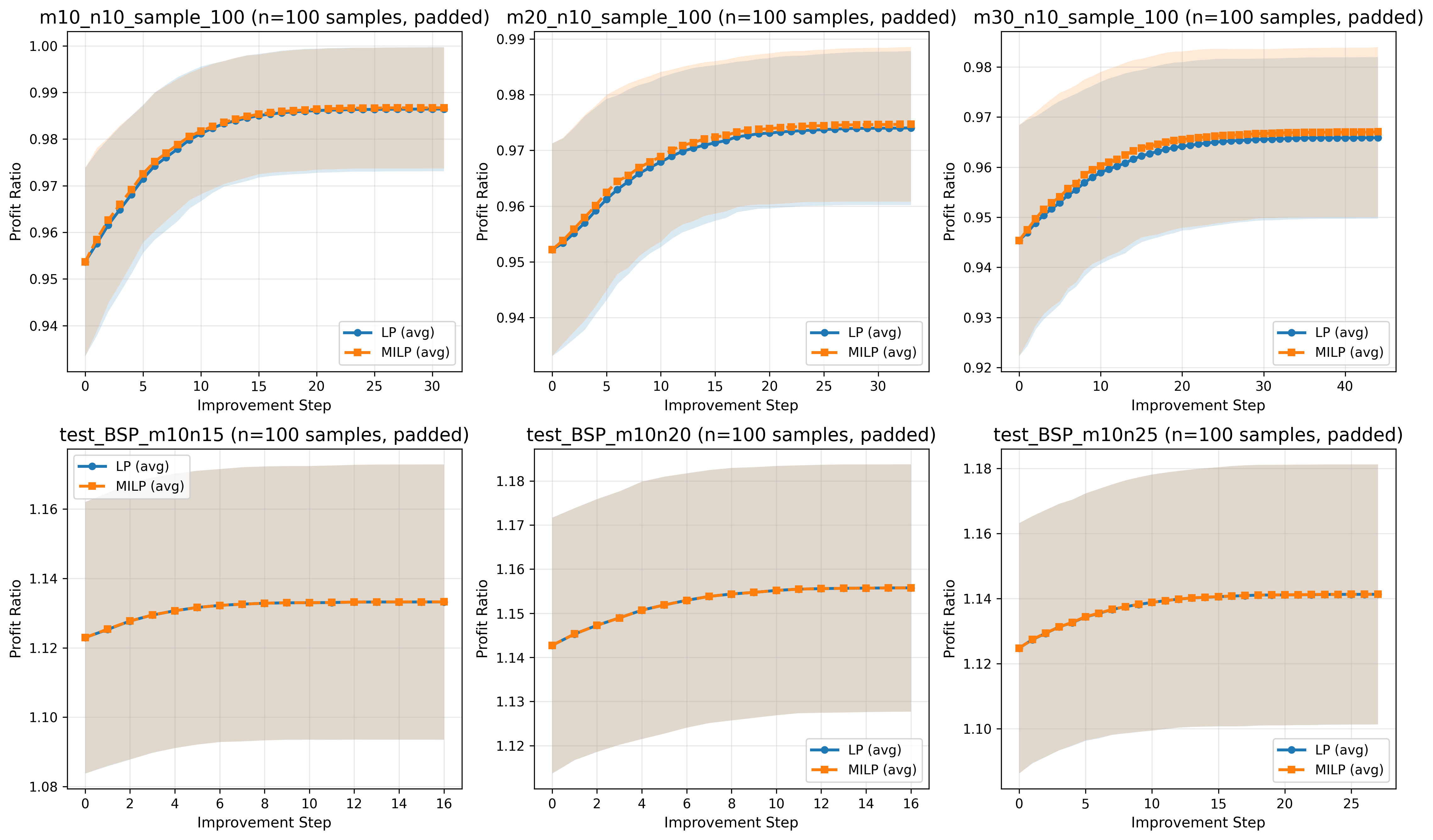}
    \caption{Trajectory comparison of normalized LP and restricted-MILP profit ratios during FCPLS local-search iterations. The synchronized upward trends indicate that the fixed-assignment LP guide is an effective empirical proxy for identifying high-quality local moves.}
    \label{fig:lp_milp_trajectory}
\end{figure}

Figure~\ref{fig:lp_milp_trajectory} reports averaged optimization trajectories over
representative problem settings. The plotted values are normalized profit ratios along
the local-search path. The LP and restricted-MILP trajectories exhibit broadly synchronized
upward trends, suggesting that the fixed-assignment LP guide tends to move the search
toward higher-quality restricted MILP solutions in practice.

\section{Out of Distribution Performance}

We test the robustness of FCP under two types of out-of-distribution shifts:
changes in the valuation function \(f(\cdot)\) and changes in the base-utility
distribution of \(u_{kj}\). The first shift changes how product-level utilities
are aggregated into bundle valuations, thereby testing whether the learned
pruning rule remains effective under different degrees of diminishing marginal
utility. The second shift changes the distribution of segment-product
preferences, testing robustness to different preference heterogeneity patterns.
The GNN is evaluated directly on these shifted instances without retraining.
Table~\ref{tab:ood-performance} shows that FCP
remains robust: it achieves a profit ratio of \(0.950\) under the more
aggressive cubic-root utility and above \(0.980\) under the other shifts, while
using about \(1\%\) or less of the MB runtime.

\begin{table}[H]
\centering
\caption{Out-of-distribution performance of FCP
\((m_{\mathrm{test}}=10,n_{\mathrm{test}}=10)\).}
\label{tab:ood-performance}
\renewcommand{\arraystretch}{1.15}
\begin{tabular}{llccc}
\toprule
\textbf{Shift Type}
& \textbf{Scenario}
& \textbf{\(PR_{\cdot,\mathrm{MB}}\) (std)}
& \textbf{Time (s)}
& \textbf{\(TR_{\cdot,\mathrm{MB}}\)} \\
\midrule
Utility function
& \(f(x)=\log(1+x)\)
& 0.980 (0.010) & 0.057 & 0.010 \\
Utility function
& \(f(x)=x^{1/3}\)
& 0.950 (0.019) & 0.060 & 0.010 \\
Distribution
& \(u_{kj}\sim \mathrm{Beta}(5,5)\)
& 0.987 (0.008) & 0.029 & 0.004 \\
Distribution
& \(u_{kj}\sim \mathrm{Beta}(0.5,0.5)\)
& 0.991 (0.008) & 0.021 & 0.003 \\
\bottomrule
\end{tabular}
\end{table}

\end{appendices}

% Acknowledgments here
% \ACKNOWLEDGMENT{}	
		% CASE 1: BiBTeX used to constantly update the references
		%   (while the paper is being written).
		%\bibliographystyle{ormsv080} % outcomment this and next line in Case 1
		%\bibliography{<your bib file(s)>} % if more than one, comma separated
		
		% CASE 2: BiBTeX used to generate mypaper.bbl (to be further fine tuned)
		%\input{mypaper.bbl} % outcomment this line in Case 2
		
		%If you don't use BiBTex, you can manually itemize references as shown below.

%\newpage

		%----------------------------------------------------------------------------------------
		
	\end{document}